\documentclass{article} 
\usepackage{iclr2027_conference,times}

\usepackage{amsmath,amsfonts,bm}

\def\eqref#1{equation~\ref{#1}}

\def\1{\bm{1}}

\DeclareMathAlphabet{\mathsfit}{\encodingdefault}{\sfdefault}{m}{sl}
\SetMathAlphabet{\mathsfit}{bold}{\encodingdefault}{\sfdefault}{bx}{n}

\newcommand{\R}{\mathbb{R}}

\usepackage{amsmath,amssymb,amsthm}
\usepackage{booktabs}
\usepackage{multirow}
\usepackage{graphicx}
\usepackage{subcaption}
\usepackage{xcolor}
\usepackage{algorithm}
\usepackage{algpseudocode}
\usepackage{tikz}
\usetikzlibrary{positioning,calc,arrows,shapes.misc}
\usepackage{hyperref}
\usepackage{url}
\usepackage{fontawesome5}
\usepackage[capitalize,nameinlink]{cleveref}
\newcommand{\method}{ParaTTT}
\newcommand{\methodshort}{PTTT}

\DeclareRobustCommand{\reviewedit}[1]{#1}

\newcommand{\bW}{\mathbf{W}}
\newcommand{\bS}{\mathbf{S}}
\newcommand{\bq}{\mathbf{q}}
\newcommand{\bk}{\mathbf{k}}
\newcommand{\bv}{\mathbf{v}}
\newcommand{\bx}{\mathbf{x}}
\newcommand{\bz}{\mathbf{z}}
\newcommand{\bo}{\mathbf{o}}
\newcommand{\bphi}{\boldsymbol{\phi}}
\newcommand{\ba}{\mathbf{a}}
\newcommand{\bb}{\mathbf{b}}
\newcommand{\bu}{\mathbf{u}}

\newcommand{\be}{\mathbf{e}}
\newcommand{\bJ}{\mathbf{J}}
\newcommand{\bI}{\mathbf{I}}
\newcommand{\bN}{\mathbf{N}}
\newcommand{\bD}{\mathbf{D}}
\newcommand{\bh}{\mathbf{h}}
\newcommand{\bg}{\mathbf{g}}

\newcommand{\dz}{\delta\bz}
\newcommand{\dzh}{\widehat{\delta\bz}}

\newtheorem{theorem}{Theorem}
\newtheorem*{theorem*}{Theorem}
\newtheorem{proposition}{Proposition}
\newtheorem{lemma}{Lemma}
\newtheorem{corollary}{Corollary}
\theoremstyle{definition}

\theoremstyle{remark}
\newtheorem{remark}{Remark}

\colorlet{fwd}{blue!60!cyan}\colorlet{rdo}{orange}\colorlet{bwd}{green!55!black}
\newcommand{\stepfwd}[1]{\textcolor{fwd!80!black}{\textbf{\textit{#1}}}}
\newcommand{\steprdo}[1]{\textcolor{rdo!85!black}{\textbf{\textit{#1}}}}
\newcommand{\stepbwd}[1]{\textcolor{bwd!70!black}{\textbf{\textit{#1}}}}
\newcommand{\kgrad}{\textsc{DeltaRuleGrad}}   
\newcommand{\by}{\mathbf{y}}
\newcommand{\br}{\mathbf{r}}
\newcommand{\bPhi}{\boldsymbol{\Phi}}
\newcommand{\bV}{\mathbf{V}}
\newcommand{\bO}{\mathbf{O}}
\newcommand{\bU}{\mathbf{U}}
\newcommand{\bT}{\mathbf{T}}
\newcommand{\bL}{\mathbf{L}}
\newcommand{\bE}{\mathbf{E}}
\newcommand{\bGamma}{\boldsymbol{\Gamma}}
\newcommand{\bbeta}{\boldsymbol{\beta}}
\newcommand{\bQ}{\mathbf{Q}}
\newcommand{\bK}{\mathbf{K}}

\DeclareRobustCommand{\rev}[1]{#1}
\DeclareRobustCommand{\revs}[1]{#1}
\DeclareRobustCommand{\revminor}[1]{#1}

\newenvironment{revblocks}{\par}{\par}
\title{\centering ParaTTT: Parallelizing Test-Time Training\\Layers with Deep Memories}

\author{%
\parbox[t]{\dimexpr\textwidth-2\tabcolsep\relax}{\centering
Noel Loo\textsuperscript{\normalfont *\,1\,2}\quad
Neehal Tumma\textsuperscript{\normalfont *\,1\,2}\quad
T. Konstantin Rusch\textsuperscript{\normalfont 2\,3}\\[0.4em]
Mathias Lechner\textsuperscript{\normalfont 2}\quad
Alexander Amini\textsuperscript{\normalfont 2}\quad
Daniela Rus\textsuperscript{\normalfont 1\,2}\\[0.5em]
\normalfont\small
\textsuperscript{1}MIT CSAIL\quad
\textsuperscript{2}Liquid AI\quad
\textsuperscript{3}Max Planck Institute for Intelligent Systems
}}

\iclrfinalcopy

\definecolor{mydarkblue}{rgb}{0,0.08,0.45}
\hypersetup{
  colorlinks=true,
  citecolor=mydarkblue,
  urlcolor=mydarkblue,
  linkcolor=mydarkblue
}

\begin{document}

\maketitle
\begingroup
\renewcommand{\thefootnote}{\fnsymbol{footnote}}
\footnotetext[1]{Equal contribution. Correspondence: \href{mailto:ntumma@mit.edu}{ntumma@mit.edu}.}
\endgroup
\fancyhead{}
\fancyhead[L]{\footnotesize\normalfont ParaTTT: Parallelizing Test-Time Training Layers with Deep Memories}
\renewcommand{\headrulewidth}{0.4pt}

\begin{abstract}
Test-time training (TTT) layers replace the linear fast-weight memory of modern recurrent architectures with a small neural network trained online by gradient descent. This nonlinear memory is more expressive than the linear memories of Gated DeltaNet and Mamba-2, but is difficult to parallelize: existing implementations such as TTT-MLP rely on sequential, block-wise weight updates that run several times slower than chunkwise-parallel linear recurrences. Moreover, their block-wise execution only approximates the token-by-token online recurrence, whereas linear recurrences admit an exact chunkwise-parallel evaluation. We introduce \method{} (\methodshort{}), an algorithm for parallelizing arbitrarily deep nonlinear memories by iteratively refining estimates of the inner-loop gradient trajectory. Starting from a guess of the hidden-layer gradients, each refinement consists of three chunkwise-parallel passes: a forward pass that evaluates the resulting feature trajectory, an exact gated-delta-rule pass that computes the readout trajectory and its feature gradients, and a backward pass that propagates these gradients into a new estimate. \rev{The exact online gradient descent recurrence is a fixed point of the iteration}. \revs{Training and prefill approximate this recurrence with a finite number of iterations, while decoding applies the exact token-by-token update without refinement.} We prove geometric convergence for \rev{a two-layer variant under sufficient contraction conditions}. With two refinement steps at 4K context, \methodshort{} evaluates the inner-loop trajectory \(32\times\) faster than TTT-MLP with the block approximation at matched error, while the full 340M model trains \(4.9\times\) faster. Models trained at 340M and 1.3B parameters match or exceed Gated DeltaNet and Mamba-2 and remain competitive with same-size Transformers on language-modeling perplexity and zero-shot accuracy, with better recall than the linear recurrences.
\end{abstract}

\vspace{-1em}
{\centering\small
\href{https://github.com/ntumm120/paraTTT}{\reviewedit{\faGithub\enspace\nolinkurl{github.com/ntumm120/paraTTT}}}\par}

\section{Introduction}
\label{sec:intro}

Softmax attention scales poorly to long contexts: its compute grows quadratically with sequence length, while its key--value cache grows linearly. Recurrent language models address both problems by processing the sequence in linear time and carrying a fixed-size state while decoding. Moreover, modern recurrent architectures such as gated linear attention, Gated DeltaNet, and Mamba-2 \citep{yang2024gla,yang2025gated,dao2024transformers} admit an efficiently parallelizable structure commonly referred to as the chunkwise-parallel form, in which tokens within each chunk are processed in parallel using dense matrix multiplications, while only the recurrent state is propagated sequentially across chunks. Given this scalable structure, the central question for this class of models is how to make the recurrent memory more expressive without sacrificing parallel evaluation.

Test-time training (TTT) addresses the expressivity side of this question by treating the recurrent state as the weights of a small model that learns from the sequence through online gradient descent \citep{sun2024learning}. With a linear memory and a squared reconstruction loss, this
recovers the delta rule of \citet{widrow1960adaptive}, the update used by DeltaNet \citep{schlag2021linear,yang2024parallelizing}, whereas a nonlinear memory can adapt its features as well as its readout. TTT-MLP, Titans, and LaCT
\citep{sun2024learning,behrouz2024titans,zhang2025lact} explore such neural memories, but their gradients depend nonlinearly on the current weights: each token changes the weights at which the next token's gradient should be evaluated, creating a sequential dependency. To recover parallelism, existing methods group gradient evaluations into blocks and evaluate every gradient in a block at the weights entering it. This creates a fidelity--speed tradeoff: larger blocks reduce the sequential depth but more coarsely approximate token-by-token learning, while smaller blocks are more faithful but require more sequential steps.

\begin{figure}[t]
    \centering
    \resizebox{\textwidth}{!}{
\definecolor{teaserblue}{RGB}{28,65,125}
\begin{tikzpicture}[
  x=1cm,y=1cm,font=\footnotesize,
  title/.style={anchor=west,font=\bfseries\footnotesize,inner sep=0pt},
  sub/.style={anchor=west,font=\scriptsize,inner sep=0pt,text=black!70},
  cell/.style={draw=black!70,fill=black!15,line width=0.6pt,
    minimum width=0.36cm,minimum height=0.29cm,inner sep=0pt},
  pcell/.style={cell,draw=teaserblue,fill=teaserblue!18},
  stage/.style={draw=black!45,fill=black!2,line width=0.55pt},
  dep/.style={-stealth,draw=black!75,line width=0.7pt},
  pdep/.style={dep,draw=teaserblue},
  stage-label/.style={font=\scriptsize,anchor=west,inner sep=0pt},
  foot/.style={font=\footnotesize,inner sep=0pt}
]
\path[use as bounding box] (0,-0.98) rectangle (14.25,5.65);
\def\pb{4.9}\def\pc{9.8}
\foreach \px/\name in {0/{(a)\quad Sequential TTT},\pb/{(b)\quad Block-wise TTT}}{
  \node[title] at (\px,5.47) {\name};
}
\node[title,text=teaserblue] at (\pc,5.47) {(c)\quad ParaTTT};
\node[sub] at (0,5.05) {Sequential over tokens};
\node[sub] at (\pb,5.05) {Parallel within each block};
\node[sub,text=teaserblue] at (\pc,5.05) {Chunkwise-parallel within each round};
\foreach \px in {0,\pb,\pc}{
  \node[font=\scriptsize,text=black!55,anchor=west,inner sep=0pt] at (\px,4.62) {Tokens};
  \foreach \i in {1,...,6}{
    \pgfmathsetmacro{\xx}{\px+1.25+(\i-1)*0.48}
    \node[font=\scriptsize,text=black!65] at (\xx,4.62) {\i};
  }
  \draw[black!20,line width=0.4pt] (\px,0.98) -- (\px+4.2,0.98);
}
\draw[black!18,line width=0.4pt] (4.55,-0.90) -- (4.55,5.6);
\draw[black!18,line width=0.4pt] (9.45,-0.90) -- (9.45,5.6);

\begin{scope}
  \foreach \i in {1,...,6}{
    \pgfmathsetmacro{\xx}{1.25+(\i-1)*0.48}
    \pgfmathsetmacro{\yy}{4.15-(\i-1)*0.52}
    \node[cell,font=\scriptsize] (s\i) at (\xx,\yy) {$\dz$};
  }
  \foreach \a/\b in {1/2,2/3,3/4,4/5,5/6}
    \draw[dep] (s\a.south east) -- (s\b.north west);
  \node[foot] at (2.1,0.65) {$T$ sequential updates};
  \node[foot,font=\bfseries\footnotesize] at (2.1,0.22) {Exact};
  \node[foot] at (2.1,-0.32) {$\dz_t^\star=\dz_t(\bar\bW_{t-1})$};
  \node[font=\scriptsize] at (2.1,-0.75) {Latest online weights};
\end{scope}

\begin{scope}[shift={(\pb,0)}]
  \draw[stage] (0.99,3.91) rectangle (2.47,4.39);
  \foreach \i in {1,2,3}{
    \pgfmathsetmacro{\xx}{1.25+(\i-1)*0.48}
    \node[cell,font=\scriptsize] at (\xx,4.15) {$\dz$};
  }
  \node[stage-label] at (0,4.15) {block 1};
  \draw[stage] (2.43,2.35) rectangle (3.91,2.83);
  \foreach \i in {4,5,6}{
    \pgfmathsetmacro{\xx}{1.25+(\i-1)*0.48}
    \node[cell,font=\scriptsize] at (\xx,2.59) {$\dz$};
  }
  \node[stage-label] at (0,2.59) {block 2};
  \draw[dep] (1.73,3.91) -- (1.73,3.37) -- (3.17,3.37) -- (3.17,2.83);
  \node[font=\scriptsize,text=black!70,align=center] at (2.1,1.55) {Gradients share the block-entry weights};
  \node[foot] at (2.1,0.65) {$T/b$ sequential blocks};
  \node[foot,font=\bfseries\footnotesize] at (2.1,0.22) {Approximation: block size $b$};
  \node[foot] at (2.1,-0.32) {$\dz_t^{\mathrm{blk}}=\dz_t(\bW_s^{\mathrm{blk}})$};
  \node[font=\scriptsize] at (2.1,-0.75) {Shared weights for $s<t\le s+b$};
\end{scope}

\begin{scope}[shift={(\pc,0)}]
  \foreach \yy/\lab in {4.15/{round 1},3.11/{round 2},1.55/{round $N$}}{
    \draw[stage,draw=teaserblue!70,fill=teaserblue!3] (0.99,\yy-0.24) rectangle (3.91,\yy+0.24);
    \foreach \i in {1,...,6}{
      \pgfmathsetmacro{\xx}{1.25+(\i-1)*0.48}
      \node[pcell,font=\scriptsize] at (\xx,\yy) {$\dz$};
    }
    \node[stage-label,text=teaserblue] at (0,\yy) {\lab};
  }
  \draw[pdep] (2.45,3.91) -- (2.45,3.35);
  \draw[pdep] (2.45,2.87) -- (2.45,2.55);
  \node[text=teaserblue] at (2.45,2.33) {$\vdots$};
  \draw[pdep] (2.45,2.1) -- (2.45,1.79);
  \node[font=\scriptsize,text=teaserblue,anchor=west] at (2.60,3.63) {refine};
  \node[foot,text=teaserblue] at (2.1,0.65) {$N$ sequential rounds};
  \node[foot,text=teaserblue,font=\bfseries\footnotesize] at (2.1,0.22) {Approximation: iteration count $N$};
  \node[foot] at (2.1,-0.32) {$\dz_{1:T}^{[N]}\approx\dz_{1:T}^{\star}$};
  \node[font=\scriptsize] at (2.1,-0.75) {Refined full-sequence estimate};
\end{scope}
\end{tikzpicture}}
    \caption{\revminor{Parallelism and approximation in nonlinear TTT.
    Each cell represents a hidden pre-activation gradient $\dz_t(\bW)=\partial\ell_t/\partial\bz_t|_{\bW}$, with the layer index omitted.
    Columns index tokens; downward steps mark computational dependencies, not approximation error or elapsed time.
    (a) Exact gradients $\dz_t^\star$ use the latest online weights $\bar\bW_{t-1}$, including decay.
    (b) Block-wise TTT, as used by TTT-MLP \citep{sun2024learning} and Titans \citep{behrouz2024titans}, evaluates gradients at shared block-entry weights $\bW_s^{\mathrm{blk}}$ for $s<t\le s+b$, omitting earlier within-block updates.
    (c) ParaTTT refines the full gradient trajectory using three chunkwise-parallel passes per round.
    \rev{The block size $b$ and refinement count $N$ control the respective approximations; the exact trajectory is a fixed point of the ParaTTT iteration.}}}
    \label{fig:teaser}
\end{figure}

In this paper, we introduce \method{} (\methodshort{}), an algorithm that evaluates an MLP-based TTT memory in parallel \revminor{(\cref{fig:teaser})}. The difficulty in parallelizing the recurrence is the inner weight trajectory: the gradient step at each token is taken at weights that depend on every earlier step, so the trajectory cannot be computed without stepping through it sequentially. \methodshort{} instead refines an estimate of this trajectory, represented by the gradients of the inner loss with respect to the hidden-layer pre-activations. Each refinement consists of three parallel passes: a forward pass that constructs the features implied by the current gradient estimate, a readout pass that computes each token's loss gradient with respect to its features, and a backward pass that propagates these gradients through the hidden layers to form a new estimate. Crucially, each refinement step is a composition of \revminor{modified instances of} existing parallel recurrences such as Gated DeltaNet \citep{yang2024parallelizing,yang2025gated}, Mamba-2 \citep{dao2024transformers}, or gated linear attention \citep{yang2024gla}. \rev{The exact nonlinear sequential recurrence is a fixed point of the iteration.} \revs{Training and prefill use a finite number of these iterations, while decoding applies the exact token-by-token update.}

Our technical contributions can be summarized as follows:
\begin{itemize}
    \item We introduce \textbf{\method{} (\methodshort{})}, an algorithm for parallelizing MLP-based recurrent memories of arbitrary depth, and prove geometric convergence for \rev{a two-layer variant under sufficient conditions} (\cref{sec:method}).
    
    \item We demonstrate \methodshort{}'s superior speed--accuracy trade-off over the block-based evaluation of TTT-MLP: at 4K context, two \methodshort{} refinements match the output error of TTT-MLP's block approximation while running $32\times$ faster, and full 340M models train $4.9\times$ faster than TTT-MLP (\cref{sec:exp:efficiency,sec:exp:convergence}).
    
    \item At 340M and 1.3B parameters, \methodshort{} attains lower perplexity and stronger recall on SWDE and FDA than Gated DeltaNet and Mamba-2, \revminor{showing that the proposed nonlinear memory improves over the evaluated linear recurrent baselines} (\cref{sec:exp:results}).
\end{itemize}

\section{Background}
\label{sec:background}
We write $\bx_{1:T}$ for the input sequence of a layer and $\bq_t,\bk_t,\bv_t$ for the per-token query, key, and value projections; the fast-weight state is a matrix $\bS_t$ or, for a deep memory, a set of weights $\bW_t$; $C$ is the chunk size of the parallel kernels. Heads are independent, and we suppress the head index. \cref{app:related} extends this section.

\subsection{Linear recurrences}
\label{sec:background:linear}
Linear recurrences retain the associative-memory view of attention while replacing its growing key--value cache with a fixed-size state. Linear attention \citep{katharopoulos2020transformers,schlag2021linear} does this by dropping the softmax: it maintains a matrix state updated by an outer product,
$\bS_t=\bS_{t-1}+\bv_t\bk_t^\top$, and reads it with the query, $\bo_t=\bS_t\bq_t$. Because this state only accumulates, practical variants introduce forgetting. Gated linear attention \citep{yang2024gla} decays the state before the write,
$\bS_t=\alpha_t\bS_{t-1}+\bv_t\bk_t^\top$, using a diagonal gate, while Mamba-2 \citep{dao2024transformers} is the scalar-gated case. DeltaNet \citep{schlag2021linear,yang2024parallelizing} updates the state by replacing the additive write with one step of gradient descent on $\tfrac12\|\bS\bk_t-\bv_t\|^2$, $\bS_t=\bS_{t-1}(\bI-\beta_t\bk_t\bk_t^\top)+\beta_t\bv_t\bk_t^\top$, which erases the value stored along $\bk_t$ before writing the new one. Gated DeltaNet \citep{yang2025gated} combines this delta-rule update with decay. All of these updates are linear in the recurrent state. Consequently, the state at the end of a chunk of $C$ tokens is an affine function of the state entering the chunk. This allows tokens within each chunk to be evaluated in parallel with dense matrix multiplications, while only the chunk-boundary states are propagated sequentially with a short scan. This exact evaluation is commonly referred to as the chunkwise-parallel form.

\subsection{Test-time training}
\label{sec:background:ttt}

Test-time training \citep{sun2024learning} makes the memory more expressive than a single matrix by interpreting the recurrent state as the parameters $\bW_t$ of a small model $f_{\bW}$ trained online. At token $t$, the model takes a gradient step on the inner loss
$\ell_t(\bW)=\tfrac12\|f_{\bW}(\bk_t)-\bv_t\|^2$
and produces the layer output $f_{\bW_t}(\bq_t)$. The projections, gates, and learning rates that define this inner problem are trained end-to-end by backpropagating through the inner updates.

\textbf{TTT-linear.}
The simplest choice of inner model recovers the delta-rule recurrence described above. With $f_{\bW}(\bk)=\bW\bk$, one gradient step on the squared reconstruction loss is exactly the DeltaNet update, while adding a decay term gives Gated DeltaNet. Because this update remains linear in the recurrent state, it retains the exact chunkwise-parallel form described above. We exploit this identity for the linear readout of a nonlinear memory.

\textbf{TTT-nonlinear.}
A nonlinear inner model can adapt the features it reads and writes with, rather than only the readout; this additional expressivity motivates deep memories. TTT-MLP \citep{sun2024learning}, for example, uses a two-layer MLP as $f_{\bW}$. Exact online gradient descent evaluates each token's gradient at the weights produced by all preceding updates, which for a
nonlinear memory prevents the chunkwise-parallel rearrangement of the linear case, so TTT-MLP uses a mini-batch (dual)
approximation: for a block of $b$ tokens starting at position $s$, every gradient is evaluated at the weights $\bW_s$ entering the block,
\begingroup\small\setlength{\abovedisplayskip}{2pt}\setlength{\belowdisplayskip}{2pt}\setlength{\abovedisplayshortskip}{0pt}\setlength{\belowdisplayshortskip}{2pt}
\begin{equation*}
\text{exact: }\bW_j=\bW_{j-1}-\eta_j\nabla\ell_j(\bW_{j-1}),\qquad
\text{block: }\bW^{\mathrm{blk}}_t=\bW_s-\sum_{j=s+1}^{t}\eta_j\nabla\ell_j(\bW_s),\quad s<t\le s+b.
\end{equation*}
\endgroup
The causal partial sum lets tokens within the block be evaluated in parallel, but the gradients are stale: updates from earlier tokens in the block do not change the weights at which later gradients are evaluated. Blocks remain sequential, requiring $T/b$ steps, so larger blocks reduce sequential depth at the cost of a coarser approximation to token-by-token learning. Titans \citep{behrouz2024titans} adds momentum and weight decay under the same block form, while LaCT \citep{zhang2025lact} uses blocks of thousands of tokens together with a sliding-window attention branch for local context.

\subsection{Other nonlinear recurrent approaches}
\label{sec:background:nonlinear}

\textbf{Sequential nonlinear RNNs.}
Classic recurrent networks \citep{elman1990finding,hochreiter1997long,cho2014learning} carry a vector state through a nonlinear map.
Their standard evaluation is sequential over tokens, limiting parallelism across the sequence. This motivates attention
\citep{vaswani2017attention} and parallel linear recurrences, but the choice of transition also affects expressivity.
Under their precision and architectural assumptions, \citet{merrill2023parallelism,merrill2024illusion} place transformers and
certain SSMs in $\mathrm{TC}^0$. This rules out general permutation composition if $\mathrm{TC}^0\ne\mathrm{NC}^1$. The xLSTM family \citep{beck2024xlstm} includes both a sequential sLSTM cell and a
parallelizable mLSTM cell. More recently, M$^2$RNN \citep{mishra2026m2rnn} uses a nonlinear transition with a matrix-valued state, but pays for it with sequential forward and \revminor{backward} passes.

\textbf{Parallelizing nonlinear recurrences.}
DEER \citep{lim2024parallelizing} and ParaRNN \citep{danieli2025pararnn} cast a nonlinear recurrence
$\bh_t=g(\bh_{t-1},\bx_t)$ as a fixed-point problem over the state trajectory and solve it with Newton iterations, where each iteration reduces to a linear recurrence evaluated by a parallel scan; deep equilibrium models \citep{bai2019deep} share the fixed-point perspective. \revminor{A generic Newton step uses a transition Jacobian whose size is quadratic in the state dimension; ParaRNN makes this tractable with structured Jacobians. \method{} instead iterates on the pre-activation gradient at each token and hidden layer. For hidden width $d$, each gradient is a $d$-dimensional vector, compared with the $O(d^2)$ fast-weight state.} Conditioned on these gradients, the fast-weight trajectory becomes a linear recurrence, so each refinement can be evaluated with chunkwise linear recurrent kernels without ever forming the full weight trajectory or its transition Jacobian.
\section{Method}
\label{sec:method}

\subsection{Test-time training with a deep memory}
\label{sec:method:setup}

We consider a single head and omit the head index. As in Gated DeltaNet, the input token $\bx_t$ is mapped to a key, a query and
a value, $\bk_t,\bq_t\in\R^{d}$ and $\bv_t\in\R^{d_v}$, by linear projections followed by a short causal convolution; keys and queries
are RMS-normalized, $\|\bk_t\|=\|\bq_t\|=\sqrt d$. The token also produces a decay gate $\alpha_t\in(0,1]$ and, for each layer $l$ of the
memory defined below, a learning rate $\eta^{(l)}_t\in(0,1)$. The gate follows \citet{yang2025gated}, each $\eta^{(l)}_t$ is the sigmoid
of a learned linear function of $\bx_t$, and all layers of the memory share the decay.

\textbf{Memory.}
The recurrent state is the set of weights $\bW=(\bW^{(1)},\dots,\bW^{(L)})$ of an $L$-layer MLP. For given weights $\bW$, with
pre-activations $\bz^{(l)}$ and activations $\bh^{(l)}$, the memory computes
\begin{equation}
    \bz^{(l)}=\bW^{(l)}\bh^{(l-1)}+\bb^{(l)},\quad
    \bh^{(l)}=\sigma\big(\bz^{(l)}\big)\quad(1\le l<L),\quad
    \bh^{(0)}=\bk,\quad
    f_{\bW}(\bk)=\bW^{(L)}\bh^{(L-1)},
    \label{eq:memory}
\end{equation}
where $\sigma$ is a pointwise nonlinearity and $\bb^{(l)}$ are learned biases. The final linear layer is the \emph{readout}, and its input
$\bphi_t=\bh^{(L-1)}_t$ defines the \emph{features}. The features are additionally projected onto the unit ball,
$n(\bx)=\bx/\max(1,\|\bx\|)$, which we absorb into the last $\sigma$. This projection is used in all experiments. Our experiments use $L=2$, as in
\citet{sun2024learning}, but the method applies to any $L$. Each weight is decomposed as
$\bW^{(l)}_t=\bW^{(l)}_0+\Delta\bW^{(l)}_t$, where $\bW^{(l)}_0$ is a parameter trained by backpropagation and
$\Delta\bW^{(l)}_t$ is the recurrent state, initialized to zero at the start of every sequence.

\textbf{Sequential recurrence.}
Token $t$ first decays the state and then takes one gradient step on the reconstruction loss
$\ell_t(\bW)=\tfrac12\|f_{\bW}(\bk_t)-\bv_t\|^2$, with the layer's learning rate as the step size. For each layer $l$,
\begin{equation}
    \bar\bW^{(l)}_{t-1}
    =\bW^{(l)}_0+\alpha_t\,\Delta\bW^{(l)}_{t-1}
    \quad\text{(decay)},\qquad
    \bW^{(l)}_t
    =\bar\bW^{(l)}_{t-1}
    -\eta^{(l)}_t\,
    \frac{\partial\ell_t}{\partial\bW^{(l)}}\Big|_{\bar\bW_{t-1}}
    \quad\text{(gradient step)}.
    \label{eq:rec-steps}
\end{equation}
The layer output is $\bo_t=f_{\bW_t}(\bq_t)$: the key is evaluated under the decayed weights, before its own gradient step, and the
query after it, exactly as in Gated DeltaNet, whose memory is the linear case $L=1$ of \cref{eq:memory}.

\Cref{eq:rec-steps} is the exact token-by-token recurrence that we seek to evaluate. Computing it directly requires $T$ sequential
gradient steps because the gradient at each token depends on all preceding updates. The block approximation used by TTT-MLP,
described in \cref{sec:background:ttt}, reduces this sequential depth by evaluating multiple gradients at shared weights, but changes
the recurrence. Our goal is instead to parallelize the trajectory while retaining the exact online recurrence as the fixed point.

\begin{figure}[t]
    \centering
    \vspace{-4pt}
    \resizebox{\textwidth}{!}{
\begin{tikzpicture}[
    x=1cm, y=1cm, font=\scriptsize,
    layer/.style={draw, rounded corners=2pt, minimum height=0.62cm, text width=2.2cm, align=center, inner sep=2pt, line width=0.6pt},
    hid/.style={layer, draw=fwd!80!black, fill=fwd!18},
    hidb/.style={layer, draw=bwd!70!black, fill=bwd!18},
    ro/.style={layer, draw=rdo!80!black, fill=rdo!25},
    ghost/.style={layer, draw=black!30, fill=white, text=black!45},
    guess/.style={draw, dashed, rounded corners=2pt, minimum height=0.62cm, text width=2.2cm, align=center, inner sep=2pt, fill=white},
    flow/.style={-stealth, line width=1.1pt},
    ffwd/.style={flow, fwd!80!black},
    fbwd/.style={flow, bwd!70!black},
    frdo/.style={flow, rdo!85!black, line width=1.5pt},
    lbl/.style={font=\tiny, align=left, inner sep=1pt, text width=1.7cm},
    stepno/.style={draw, circle, fill=white, inner sep=1.4pt, font=\bfseries\footnotesize, line width=0.7pt},
    ptitle/.style={font=\bfseries\small, anchor=west, align=left, text width=3.75cm, inner sep=2pt},
    panel/.style={draw=black!25, rounded corners=4pt, fill=black!1.5, line width=0.5pt},
    pfwd/.style={panel, draw=fwd!60!black, fill=fwd!5},
    prdo/.style={panel, draw=rdo!70!black, fill=rdo!6},
    pbwd/.style={panel, draw=bwd!60!black, fill=bwd!6},
]
\def\pw{4.55}   
\def\gap{0.2}
\def\ph{5.75}   
\def\xb{1.35}   
\def\xl{2.6}   
\def\yin{0.6}
\def\yone{1.8}
\def\ydots{2.65}
\def\ytop{3.5}
\def\yro{4.7}
\def\ytitle{5.4}
\pgfmathsetmacro{\pb}{\pw+\gap}
\pgfmathsetmacro{\pc}{2*(\pw+\gap)}
\pgfmathsetmacro{\xend}{\pc+\pw}

\foreach \px/\num/\ttl/\sty in {0/1/{Parallel forward pass}/pfwd, \pb/2/{Readout, then its gradient}/prdo, \pc/3/{Parallel backward pass}/pbwd} {
  \draw[\sty] (\px,0) rectangle ++(\pw,\ph);
  \node[stepno] at (\px+0.38,\ytitle) {\num};
  \node[ptitle] at (\px+0.6,\ytitle) {\ttl};
}

\begin{scope}
  \node[guess] (g1) at (\xb,\yin) {keys $\bk_{1:T}$, current\\ backward grads $\bD^{[n-1]}$};
  \node[hid]   (l1) at (\xb,\yone) {hidden layer 1\\ $\Delta\bW^{(1)}$};
  \node at (\xb,\ydots) {$\vdots$};
  \node[hid]   (lt) at (\xb,\ytop) {hidden layer $L{-}1$\\ $\Delta\bW^{(L-1)}$};
  \node[ghost] (r1) at (\xb,\yro) {readout};
  \draw[ffwd] (g1) -- (l1);
  \draw[ffwd] (l1) -- (\xb,\ydots-0.22);
  \draw[ffwd] (\xb,\ydots+0.22) -- (lt);
  \draw[ffwd] (lt) -- node[lbl, right=1pt, pos=0.5] {features $\bphi^{[n]}$} (r1);
  \node[lbl, anchor=west] at (\xl,\yone) {\revminor{shifted GLA,}\\ then $\sigma$};
  \node[lbl, anchor=west] at (\xl,\ytop) {keys $=$ outputs of\\ the layer below};
  \node[lbl, anchor=west] at (\xl,\yin) {\revminor{update values}\\ $-\eta^{(l)}\dz^{(l)}$};
\end{scope}

\begin{scope}[shift={(\pb,0)}]
  \node[guess, draw=black!35, text=black!45] (g2) at (\xb,\yin) {keys $\bk_{1:T}$, current\\ backward grads $\bD^{[n-1]}$};
  \node[ghost] (l1b) at (\xb,\yone) {hidden layer 1};
  \node at (\xb,\ydots) {$\vdots$};
  \node[ghost] (ltb) at (\xb,\ytop) {hidden layer $L{-}1$};
  \node[ro]    (r2) at (\xb,\yro) {readout\\ gated delta rule};
  \draw[flow, black!45] (ltb) -- node[lbl, right=1pt, pos=0.5, text=black!45] {$\bphi^{[n]}$} (r2);
  \draw[frdo] ([xshift=-0.45cm]r2.south) -- ([xshift=-0.45cm]ltb.north);
  \node[lbl, anchor=west] at (\xl,\yro) {trajectory of $\bW^{(L)}$\\ on $\bphi^{[n]}$ is exact};
  \node[lbl, anchor=west, text width=1.75cm] at (\xl,\ytop+0.05) {\kgrad{}: gradient\\ at the features,\\ $\be_t\propto\partial\ell_t/\partial\bphi_t$\\ (read before token $t$'s write)};
  \node[lbl, anchor=west, text=black!45] at (\xl,\yone) {hidden layers\\ unchanged here};
\end{scope}

\begin{scope}[shift={(\pc,0)}]
  \node[ghost] (r3) at (\xb,\yro) {readout};
  \node[hidb]  (lt3) at (\xb,\ytop) {hidden layer $L{-}1$\\ $\bJ^{\top}$};
  \node at (\xb,\ydots) {$\vdots$};
  \node[hidb]  (l13) at (\xb,\yone) {hidden layer 1\\ $\bJ^{\top}$};
  \node[guess] (g3) at (\xb,\yin) {refined backward\\ grads $\bD^{[n]}$};
  \draw[fbwd] (r3) -- node[lbl, right=1pt, pos=0.5] {$\be/\eta^{(L)}$} (lt3);
  \draw[fbwd] (lt3) -- (\xb,\ydots+0.22);
  \draw[fbwd] (\xb,\ydots-0.22) -- (l13);
  \draw[fbwd] (l13) -- (g3);
  \node[lbl, anchor=west, text width=1.75cm] at (\xl,\ytop-0.1) {$\dzh^{(l)}=\bJ^{\top}\bar\bW^{\top}\dzh^{(l+1)}$\\ \revminor{(shifted transposed read)}};
\end{scope}

\draw[flow, black!60, line width=1.3pt, rounded corners=6pt] (\pc+\xb,-0.05) -- (\pc+\xb,-0.55) -- (\xb,-0.55) -- (\xb,-0.05);
\node[fill=white, inner sep=3pt, font=\scriptsize] at (\xend/2,-0.55) {\tikz\node[stepno]{4};\; repeat $N$ times, then one last parallel forward pass (keys shifted, queries unshifted) and the readout's output pass};
\node[font=\tiny, text=black!55, anchor=north east] at (\xend,-0.9) {\revminor{all recurrences use the chunkwise-parallel form over $T$ tokens}};
\end{tikzpicture}}
    \vspace{-6pt}
    \caption{One iteration of \method{}; \revminor{the recurrences are evaluated in chunkwise-parallel form over all $T$ tokens.} (1) A parallel forward pass
    evaluates the hidden layers from the current estimate of their pre-activation gradients. (2) The readout runs exactly on the
    resulting features and \kgrad{} returns each token's feature gradient. (3) A parallel backward pass propagates it through the hidden
    layers \revminor{using shifted reads of the transposed recurrences (see Step~3 in \cref{sec:method:alg} for details)} and updates the estimate. (4) After $N$ iterations a final forward pass and readout produce the output.}
    \label{fig:algorithm}
    \vspace{-8pt}
\end{figure}

\subsection{\method{}: Iterated refinement of the inner-loop weight trajectory}
\label{sec:method:alg}

We first describe the \method{} iteration, which uses a linear attention update for the hidden layers and an uncentered readout. The practical \revminor{design choices} used by our models are introduced in \cref{sec:method:practical}.

The central difficulty in parallelizing the nonlinear MLP recurrence in \cref{eq:rec-steps} is its inner weight trajectory: the
gradient step at token $t$ is taken at weights that depend on every earlier step, so the trajectory cannot be computed without
stepping through it sequentially. \method{} instead iteratively refines an estimate of this trajectory, with each refinement parallel
across the sequence length. The estimate is represented by a guess of the pre-activation gradients
$\dz^{(l)}_t=\partial\ell_t/\partial\bz^{(l)}_t$ of the hidden layers. Given such a guess, a \emph{parallel forward pass} evaluates the
weight trajectory these gradients define (Step~1), the readout is evaluated exactly on the resulting features (Step~2), and a
\emph{parallel backward pass} propagates the resulting feature gradients through the trajectory to obtain a new guess (Step~3).
At convergence, the gradients produced by the backward pass agree with those used in the forward pass
(\cref{fig:algorithm}).

To see why each pass can be parallelized, write $\bh^{(l)}_t$ for the activations of $\bk_t$ under the decayed weights
$\bar\bW_{t-1}$ and obtain $\dz^{(l)}_t$ by backpropagation through \cref{eq:memory} at $\bar\bW_{t-1}$. The gradient step in
\cref{eq:rec-steps} is the outer product of the two, so in terms of the state the recurrence reads
\begin{equation}
    \Delta\bW^{(l)}_t
    =\alpha_t\,\Delta\bW^{(l)}_{t-1}
    -\eta^{(l)}_t\,\dz^{(l)}_t\,\bh^{(l-1)\top}_t
    \qquad (l=1,\dots,L).
    \label{eq:rec-W}
\end{equation}
Given a guess of the hidden-layer gradients $\bD=(\dz^{(l)}_{1:T})_{l<L}$, one vector per token and hidden layer,
\cref{eq:rec-W} is linear in its state with known coefficients. Conditioned on $\bD$, the nonlinear recurrence therefore reduces
to a collection of linear recurrences. For a hidden layer, \cref{eq:rec-W} is exactly gated linear attention with scalar decay,
the state update of Mamba-2 \citep{yang2024gla,dao2024transformers}, with keys $\bh^{(l-1)}_t$, values
$-\eta^{(l)}_t\dz^{(l)}_t$ and gates $\alpha_t$; for the readout, where $\dz^{(L)}_t$ is the readout's own error, it is a
Gated DeltaNet \citep{yang2025gated}. Both admit exact chunkwise-parallel kernels. The nonlinearity is thus confined to the
self-consistency condition that $\bD$ determines the state trajectory while the resulting trajectory determines $\bD$.
\method{} solves this fixed-point problem by iteration from $\bD=\mathbf 0$ (\cref{fig:algorithm}; pseudocode in
\cref{alg:main}, \cref{app:alg}). Since the vectors of $\bD$ enter the scans as values, we also call them \textit{pseudo-labels}.

\textbf{Step 1: parallel forward pass.}
Given the current estimate $\bD$, the forward pass evaluates the memory MLP at all $T$ tokens, each under its own decayed weights,
layer by layer and without ever forming the fast weights. Unrolling \cref{eq:rec-W} for a hidden layer $l<L$,
\begin{equation}
    \alpha_t\Delta\bW^{(l)}_{t-1}\,\bh^{(l-1)}_t
    =-\sum_{j<t}
    \Big(\prod_{i=j+1}^{t}\alpha_i\Big)
    \eta^{(l)}_j\,\dz^{(l)}_j\,
    \big(\bh^{(l-1)\top}_j\bh^{(l-1)}_t\big),
    \label{eq:body-la}
\end{equation}
which is gated linear attention \citep{yang2024gla} with keys $\bh^{(l-1)}_j$, values $-\eta^{(l)}_j\dz^{(l)}_j$ and gates
$\alpha_j$; the sum excludes $j=t$ because key $t$ is read before its own write. We call a scan \emph{shifted} when token $t$
reads the state \emph{before} its own update, $\bar\bW_{t-1}$, and \emph{unshifted}, the standard form, when it reads the state
\emph{after} the update, $\bW_t$; keys are read shifted and queries unshifted. One chunkwise scan thus gives the fast contribution
to the pre-activations of layer $l$ at all $T$ tokens; adding $\bW^{(l)}_0\bh^{(l-1)}_t+\bb^{(l)}$ and applying $\sigma$ gives
the inputs of the next layer. The pass ends with the features $\bphi_{1:T}$ and stores the Jacobians $\bJ^{(l)}_t$ for Step~3.
When $\bD=\mathbf 0$ on the first forward pass, the fast scan is identically zero and can be skipped.

\textbf{Step 2: readout trajectory and its gradient.}
Unlike the hidden layers, the readout needs no trajectory estimate. \revminor{Once its input features are fixed, its fast weights follow a
Gated DeltaNet with keys $\bphi_t$, corrected values $\tilde\bv_t=\bv_t-\bW^{(L)}_0\bphi_t$, gates $\alpha_t$ and step sizes
$\eta^{(L)}_t$. The chunkwise gated-delta-rule kernel \citep{yang2024parallelizing,yang2025gated} computes this recurrence exactly\revs{, while}
the slow readout $\bW^{(L)}_0$ is retained in the feature gradient and added to the final output (\cref{app:dk-simple}).}

The backward pass in Step~3 additionally requires each token's loss gradient with respect to its features. We obtain this during the
readout pass:
\begin{equation}
    \be_t
    =\eta^{(L)}_t\,\bar\bW^{(L)\top}_{t-1}
    \big(\bar\bW^{(L)}_{t-1}\bphi_t-\bv_t\big)
    =\eta^{(L)}_t\,\partial\ell_t/\partial\bphi_t
    =-\,\bar\bW^{(L)\top}_{t-1}\bu_t ,
    \label{eq:dk}
\end{equation}
where $\bu_t=\eta^{(L)}_t(\bv_t-\bar\bW^{(L)}_{t-1}\bphi_t)$ is the update token $t$ writes to the readout.
The \kgrad{} kernel computes this quantity as a small modification of the existing GDN kernel
(\cref{app:dk-simple}).

\textbf{Step 3: parallel backward pass.}
The backward pass propagates each token's feature gradient through the hidden layers along the trajectory from Step~1, for all
$T$ tokens at once, giving a recomputed gradient trajectory $\hat\bD=(\dzh^{(l)}_{1:T})_{l<L}$. It requires the transposed
Jacobian of $\sigma$ and the transposed weights; the only operation involving a fast weight is the transpose of
\cref{eq:body-la},
\begin{equation}
    \big(\alpha_t\Delta\bW^{(l)}_{t-1}\big)^\top\dz_t
    =-\sum_{j<t}
    \Big(\prod_{i=j+1}^{t}\alpha_i\Big)
    \eta^{(l)}_j
    \big(\dz^{(l)\top}_j\dz_t\big)
    \bh^{(l-1)}_j ,
    \label{eq:body-la-T}
\end{equation}
\revminor{a shifted linear-attention read} with the pseudo-labels as queries and keys and the layer inputs as values, computed by the same chunkwise scan.
The estimate is then updated by a damped Picard step,
\begin{equation}
    \bD^{[n]}
    =\bD^{[n-1]}+\lambda\big(\hat\bD-\bD^{[n-1]}\big),
    \qquad \lambda=0.8 .
    \label{eq:picard}
\end{equation}

\textbf{Output.}
After $N$ iterations, a final forward pass computes the features of the keys using shifted reads and of the queries using unshifted
reads, since each query is evaluated under the updated weights. One gated-delta-rule pass with these features as keys and queries
\revminor{and corrected values $\tilde\bv_t=\bv_t-\bW^{(L)}_0\bphi_{\bk,t}$ produces the fast contribution\revs{. Finally,} adding
$\bW^{(L)}_0\bphi_{\bq,t}$ gives} the output. \Cref{sec:method:theory} shows that the exact sequential recurrence is a fixed point of this iteration and
gives conditions under which the iterates converge to it.

\textbf{Cost, training and decoding.}
Each iteration requires a constant number of chunkwise scans per memory layer, with sequential depth $O(T/C)$ for chunk size $C$
(\cref{app:kernels}). \revs{We use $N=2$ unless stated otherwise and train all parameters by backpropagation through the $N$ iterations. This finite-iteration computation approximates the sequential recurrence; its agreement with exact evaluation depends on the sequence length and training recipe (\cref{sec:exp:convergence,app:mismatch}).}
Decoding requires no iteration: with a single new token, \cref{eq:rec-steps} is one exact gradient step from the cached weights,
so $N$ affects prefill only.

\subsection{Fixed point and convergence}
\label{sec:method:theory}
\label{sec:method:fixedpoint}

\begin{proposition}[Fixed point and finite-step exactness, informal]
\label{prop:fixedpoint}
For any memory depth $L$, the gradient sequence $\bD^\star$ of the sequential recurrence
\cref{eq:rec-steps} is a fixed point of the iteration of \cref{sec:method:alg}, and the parallel evaluation at $\bD^\star$
reproduces the sequential recurrence. Without damping ($\lambda=1$), every iteration makes the estimate at one more token exact,
so $N\ge T$ iterations reproduce the sequential recurrence exactly.
\end{proposition}

\Cref{prop:fixedpoint}, proved in \cref{app:finite-exactness}, needs no contraction assumption. For a geometric rate we
analyze the iteration with $L=2$, the linear-attention hidden update \cref{eq:body-la} and the uncentered readout, with the
unit-ball projection replaced, for the analysis only, by the soft clip $n_{\mathrm{soft}}(\bx)=\bx/\sqrt{1+\|\bx\|^2}$, so that the
features $\bphi_t=\psi(\bz_t)$, $\psi=n_{\mathrm{soft}}\circ\sigma$, have a globally Lipschitz Jacobian (\cref{app:theory} restates
the setting). Write $\|\bD\|_\infty=\max_t\|\dz^{(1)}_t\|$ and $F$ for one iteration's map $\bD\mapsto\hat\bD$.

\begin{theorem}[Convergence of \method{} with soft clipping, $L=2$]
\label{thm:convergence}
Assume $\|\bk_t\|\le1$, $\|\bv_t\|\le V$, $\|\bW^{(2)}_0\|_{\mathrm{op}}\le w_0$, a shared decay $\alpha_t\le\bar\alpha<1$, a shared
learning rate $\eta^{(1)}_t=\eta^{(2)}_t\le\bar\eta\le1$, and $|\sigma'|\le c_\sigma$, $|\sigma''|\le h_\sigma$. With
$\epsilon=\bar\eta/(1-\bar\alpha)$ and $Q=V+w_0$, $F$ is globally Lipschitz in $\|\cdot\|_\infty$ with constant
\begin{equation}
    \rho=\epsilon(1+\epsilon)Q^2\big[c_\sigma^2(1+\epsilon)(5+4\epsilon)+h_\sigma(2+\epsilon)\big]
    \;\le\;(10c_\sigma^2+3h_\sigma)\,\epsilon Q^2\quad\text{for }\epsilon\le\tfrac14 .
    \label{eq:rho}
\end{equation}
If $\rho<1$, $F$ has a unique fixed point $\bD^\star$, the gradient sequence of the sequential recurrence \cref{eq:rec-steps} with
$n_{\mathrm{soft}}$ in place of the projection, and for any initialization and damping $\lambda\in(0,1]$ the iteration
\cref{eq:picard} satisfies $\|\bD^{[n]}-\bD^\star\|_\infty\le(1-\lambda(1-\rho))^n\|\bD^{[0]}-\bD^\star\|_\infty$; the features,
readout states and outputs converge at the same rate, up to constants.
\end{theorem}

The proof (\cref{app:theory}) controls the features and their changing Jacobians; SiLU satisfies the derivative assumptions, and
$\rho<1$ is independent of the sequence length. 
\revs{\Cref{app:theory-delta} extends the analysis to the hidden-layer delta rule (\cref{thm:convergence-delta}) and centering (\cref{cor:centering}), while retaining the soft clip and sufficient contraction conditions. These results identify a regime in which convergence is guaranteed. The trained models use RMS-normalized keys and a hard projection, and their learned rates and gates lie outside the contraction conditions (\cref{app:gates}). We measure their agreement with the exact recurrence directly in \cref{sec:exp:convergence,app:mismatch}.}


\subsection{\revminor{Practical design choices}}
\label{sec:method:practical}

The analysis above uses the simplest form of the iteration. Our reported models use two \revminor{design choices} that substantially improve
practical performance: a hidden-layer delta rule that preserves the target sequential recurrence, and centering, which changes the
target to the sequential recurrence with the centered readout.

\textbf{Delta rule for the hidden layers.}
Step~1 of the iteration evolves each hidden layer's fast weights by gated linear attention \cref{eq:body-la}, which sums the
writes $-\eta^{(l)}_j\dz^{(l)}_j\bh^{(l-1)\top}_j$ with a scalar decay. The same weights can instead be evolved by a gated delta
rule whose target is built from the previous iteration's read. The update is \cref{eq:rec-W} plus a correction that erases stale
writes along the iterates (\cref{app:body-delta}). This correction vanishes at the fixed point, so the target sequential recurrence
is unchanged. The delta rule costs slightly more but gives substantially better recall (\cref{tab:main}), and we therefore use it
throughout.

\textbf{Centering.}
The hidden layers receive gradient only through the readout. By \cref{eq:dk}, $\be_t$ is proportional to
$\bar\bW^{(L)\top}_{t-1}$, so a zero-initialized readout gives the hidden layers no gradient at the start of a sequence. Conversely,
a nonzero $\bW^{(L)}_0$ provides a nonzero gradient immediately, but also adds a persistent bias $\bW^{(L)}_0\bphi_t$ to the output.
We use \textit{centering} to obtain a nonzero readout Jacobian at initialization while keeping the initial output zero. Centering
trains $f_\theta-f_{\theta_0}$ in place of $f_\theta$ \citep{chizat2019lazy}; related zero-output initializations appear in
\citet{woodworth2020kernel,hu2020simple}.
We subtract from the readout its value on the features $\bphi_{0,t}$ of the slow MLP, that is,
$f_{\bW}(\bk_t)=\bW^{(L)}\bphi_t-\bW^{(L)}_0\bphi_{0,t}$. At the start of a sequence, $\bphi_t=\bphi_{0,t}$ so the readout is zero,
while its Jacobian $\bW^{(L)}_0+\Delta\bW^{(L)}$ is nonzero from the first token. Centering changes only the values passed to Step~2
and adds one term to the output (\cref{app:centering}).

\textbf{Other choices.}
We use $L=2$ with a hidden width of $d$, SiLU as $\sigma$, and a learned per-head gain on the fast part of the hidden
pre-activation. Setting the hidden learning rates to zero yields a Gated DeltaNet on the features of the frozen MLP, the $N=0$
control.

\section{Empirical study}
\label{sec:exp}
\begin{table}[t]
\centering
\vspace{-4pt}
\caption{Main results at 340M and 1.3B. Recall follows the generation protocol of \citet{arora2024simple}; NIAH is the RULER needle average at the given length. Best per column and size in bold, second best underlined.}
\label{tab:main}
\vspace{-0.5em}
\setlength{\tabcolsep}{3pt}
\resizebox{\textwidth}{!}{%
\begin{tabular}{@{}lc rr rrrrrrr rrrr rrrr@{}}
\toprule
 & & \multicolumn{2}{c}{Perplexity $\downarrow$} & \multicolumn{7}{c}{Zero-shot acc.\ $\uparrow$} & \multicolumn{4}{c}{Recall $\uparrow$} & \multicolumn{4}{c}{NIAH $\uparrow$} \\
\cmidrule(lr){3-4} \cmidrule(lr){5-11} \cmidrule(lr){12-15} \cmidrule(lr){16-19}
Model & $N$ & Wiki & Lamb & PIQA & Hella & ARC-e & Wino & ARC-c & BoolQ & SciQ & SWDE & FDA & TQA & DROP & \revminor{4K} & \revminor{8K} & \revminor{16K} & \revminor{32K} \\
\midrule
\multicolumn{19}{@{}l}{\textit{340M, 24 layers, 15.7B tokens}} \\
GDN & -- & 26.57 & 29.66 & 65.2 & 36.5 & 47.1 & 51.5 & 24.1 & 56.6 & 78.4 & 32.5 & 17.8 & 35.7 & \underline{15.3} & 25.9 & 17.0 & 13.7 & 9.3 \\
Mamba2 & -- & 26.99 & 45.87 & \underline{65.7} & 35.1 & 45.5 & 51.9 & 23.4 & 57.8 & 75.2 & 25.8 & 5.7 & 30.6 & 12.6 & 8.1 & 3.0 & 1.1 & 1.2 \\
Transformer & -- & \textbf{25.34} & 39.25 & 64.3 & 35.8 & 46.3 & \underline{52.8} & 22.4 & \underline{60.8} & 76.7 & \textbf{54.6} & \textbf{28.4} & \underline{40.5} & 14.5 & \underline{38.6} & \underline{23.2} & \textbf{16.3} & 10.5 \\
TTT-MLP & -- & 26.12 & 40.35 & 65.2 & 35.1 & 46.3 & 50.8 & 23.3 & 58.5 & 77.2 & 33.1 & 16.2 & 35.2 & 13.9 & 31.5 & 20.4 & 13.0 & \underline{12.4} \\
\revminor{LaCT (\revs{no SWA})} & -- & 463.2 & $9.4{\times}10^{4}$ & 53.7 & 25.0 & 28.8 & 50.1 & 24.1 & 60.1 & 21.2 & 0.1 & 0.0 & 0.1 & 7.6 & 0.0 & 0.0 & 0.0 & 0.0 \\
\textbf{\methodshort{} (ours)} & 2 & \underline{25.46} & \underline{26.93} & 65.3 & \textbf{37.7} & \underline{48.6} & 51.2 & 24.1 & 59.8 & \underline{79.0} & 41.9 & 20.3 & 39.6 & 14.4 & 32.6 & 20.7 & 15.6 & \textbf{16.0} \\
\quad \revminor{zero-init readout} & 2 & 26.06 & 32.62 & 64.5 & 36.8 & 48.4 & 50.6 & \textbf{25.3} & 58.2 & 78.9 & 36.6 & 19.6 & 37.0 & 14.3 & 29.0 & 21.1 & 15.0 & 12.2 \\
\quad linear-attention body & 2 & 25.62 & \textbf{25.88} & \textbf{65.8} & 37.2 & \textbf{48.7} & \textbf{53.2} & 23.5 & \textbf{61.0} & 78.9 & 28.4 & 12.9 & 33.2 & 11.2 & 7.2 & 5.5 & 4.0 & 2.8 \\
\quad trained at $N{=}5$ & 5 & 25.47 & 28.44 & 64.9 & \underline{37.6} & 46.2 & 51.4 & \underline{24.5} & 58.2 & \textbf{79.3} & \underline{45.0} & \underline{22.0} & \textbf{41.9} & \textbf{15.6} & \textbf{40.1} & \textbf{25.9} & \underline{15.9} & 5.8 \\
\midrule
\multicolumn{19}{@{}l}{\textit{1.3B, 24 layers, 100B tokens}} \\
GDN & -- & 15.78 & 10.56 & 72.2 & 54.9 & 60.0 & 58.9 & 28.1 & 53.9 & 86.7 & 59.0 & 27.0 & 52.9 & 15.2 & 50.8 & 34.1 & 18.8 & \underline{17.5} \\
Mamba2 & -- & 15.42 & 11.62 & \textbf{73.6} & 53.5 & 58.0 & 54.1 & 28.2 & 53.9 & 87.7 & 56.7 & 21.5 & \underline{54.5} & 14.8 & 33.4 & 19.6 & 11.3 & 4.9 \\
Transformer & -- & \textbf{15.04} & 10.91 & 71.5 & 53.5 & 58.0 & 56.4 & \textbf{29.9} & \underline{57.7} & 88.3 & \textbf{75.3} & \textbf{31.9} & 53.7 & 15.2 & \textbf{78.1} & \textbf{56.7} & \textbf{43.4} & 17.4 \\
\textbf{\methodshort{} (ours)} & 2 & \underline{15.21} & \underline{9.00} & 71.9 & \textbf{56.0} & \underline{61.0} & 58.6 & \textbf{29.9} & \textbf{61.3} & \textbf{89.2} & 62.8 & \underline{28.1} & 53.7 & 15.0 & 53.8 & 37.1 & 17.5 & 14.7 \\
\quad \revs{last 11B at $N{=}5$} & \revs{2} & \revs{15.33} & \revs{\textbf{8.77}} & \revs{\underline{72.6}} & \revs{\underline{55.9}} & \revs{\textbf{61.3}} & \revs{\textbf{59.7}} & \revs{\underline{29.6}} & \revs{53.8} & \revs{\underline{88.4}} & \revs{\underline{67.8}} & \revs{26.0} & \revs{\textbf{55.6}} & \revs{\underline{15.3}} & \revs{55.5} & \revs{38.4} & \revs{17.8} & \revs{14.8} \\
\quad \revs{Poisson $N$} & \revs{2} & \revs{15.28} & \revs{9.39} & \revs{72.4} & \revs{55.8} & \revs{60.3} & \revs{\underline{59.0}} & \revs{28.9} & \revs{57.4} & \revs{88.0} & \revs{63.6} & \revs{25.5} & \revs{54.1} & \revs{\textbf{16.2}} & \revs{\underline{58.8}} & \revs{\underline{40.6}} & \revs{\underline{22.7}} & \revs{\textbf{20.2}} \\
\bottomrule
\end{tabular}}
\end{table}

\textbf{Setup.}
\revs{All models use 24 pre-norm residual blocks, each with a token-mixing layer and SwiGLU MLP. The backbone width is $d=1024$ at 340M and $d=2048$ at 1.3B, with a default head dimension of 128; TTT-MLP and LaCT use the layouts in \cref{app:configs}. We use short QKV convolutions as in Gated DeltaNet \citep{yang2025gated}. \method{} has one hidden layer of width equal to the head dimension, the hidden-layer delta rule and centered readout (\cref{sec:method:practical}), and $N=2$ training iterations unless stated otherwise. We report the no-SWA LaCT configuration in \cref{app:configs} and compare LaCT, Gated DeltaNet and \method{} with the same sliding window in \cref{app:swa}.}

\subsection{Language modeling performance}
\label{sec:exp:results}

\textbf{Model set and evaluation.}
\revs{At 340M, we compare Gated DeltaNet, Mamba-2, attention, TTT-MLP, LaCT without SWA, and \method{}, including its zero-init readout, linear-attention body and $N=5$ variants. At 1.3B, we compare Gated DeltaNet, Mamba-2, attention and \method{}. Models share the training recipe, data order and seed, using 15.7B SlimPajama tokens at 340M and 100B at 1.3B (\cref{app:configs}). Comparisons match data and steps, with one seed per \method{} configuration; compute and state size are not matched. We evaluate WikiText-103 and LAMBADA perplexity, zero-shot accuracy with lm-evaluation-harness \citep{gao2023lmeval}, recall on SWDE, FDA, TriviaQA and DROP following \citet{arora2024simple,arora2024just}, and RULER NIAH \citep{hsieh2024ruler} at 4K--32K tokens, using YaRN \citep{peng2024yarn} for attention. \Cref{app:eval} details the protocol.}

\begin{figure}[t]
    \centering
    \vspace{8pt}
    \includegraphics[width=\textwidth]{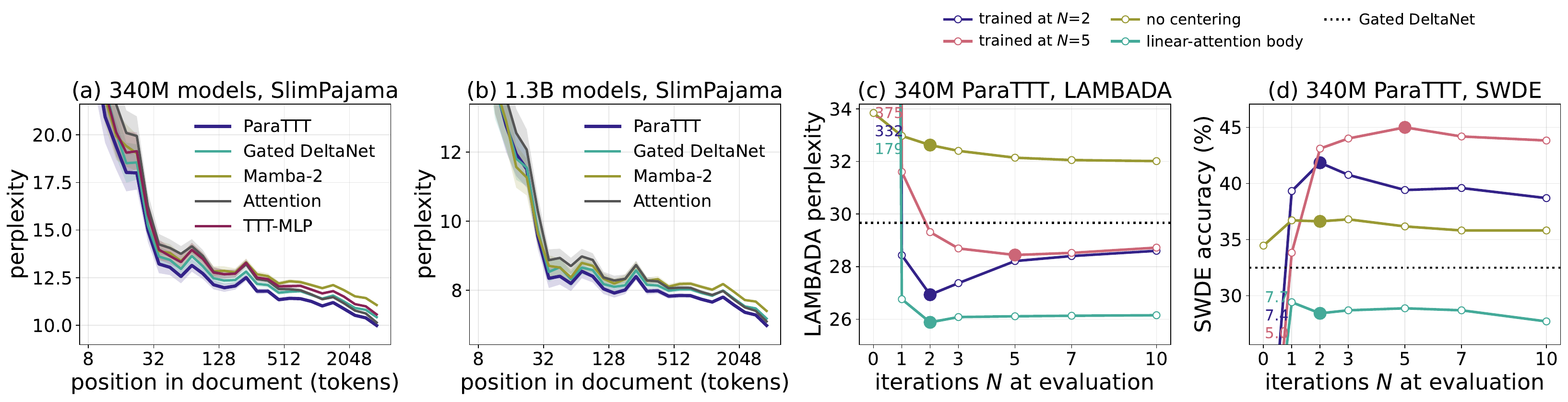}
    \vspace{-18pt}
    \caption{(a,b) Perplexity along $698$ held-out SlimPajama documents at 340M and 1.3B. (c,d) LAMBADA
    perplexity and SWDE accuracy of the 340M \method{} models against the number of iterations at evaluation; filled marker: training
    $N$; dotted line: Gated DeltaNet; off-chart loop-off values are printed at the edge. \revminor{The uncentered model uses a zero-initialized readout.}}
    \label{fig:perpos-evaln}
\end{figure}
\begin{figure}[t]
    \centering
    \vspace{4pt}
    \includegraphics[width=\textwidth]{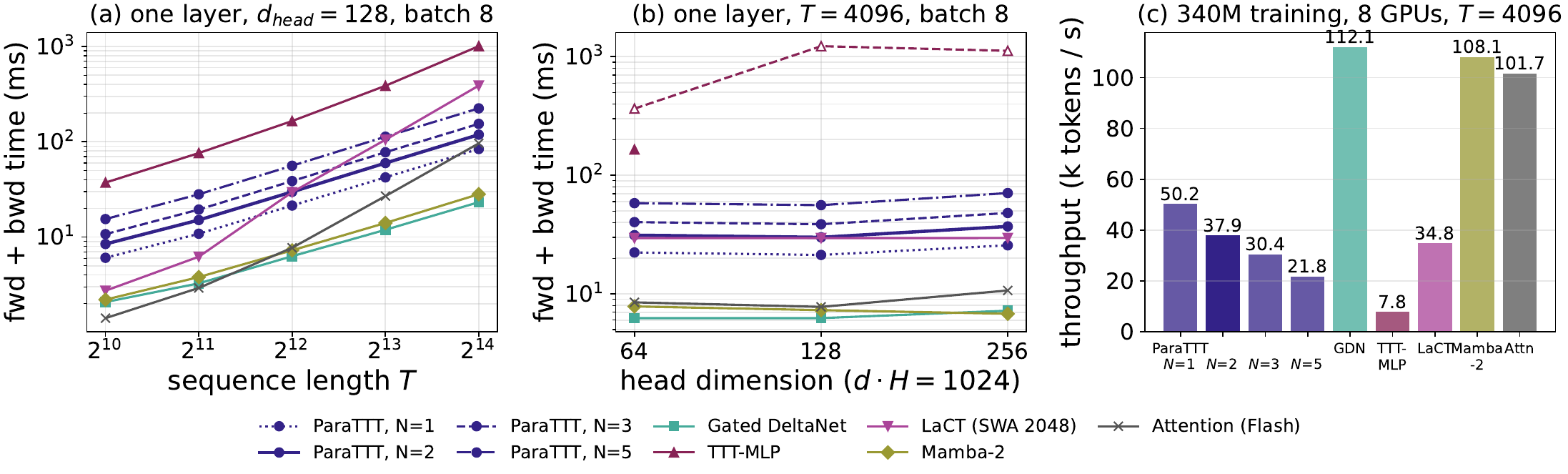}
    \vspace{-16pt}
    \caption{Efficiency. Forward+backward time of one token-mixing layer against (a) sequence length and (b) head dimension;
    (c) training throughput of the full 340M models on eight MI325X GPUs. \rev{The fused TTT-MLP implementation used in our benchmarks supports
    head dimension 64; hollow markers show a compiled scan at larger dimensions.}}
    \label{fig:efficiency}
\end{figure}

\begin{figure}[t]
    \centering
    \includegraphics[width=0.95\textwidth]{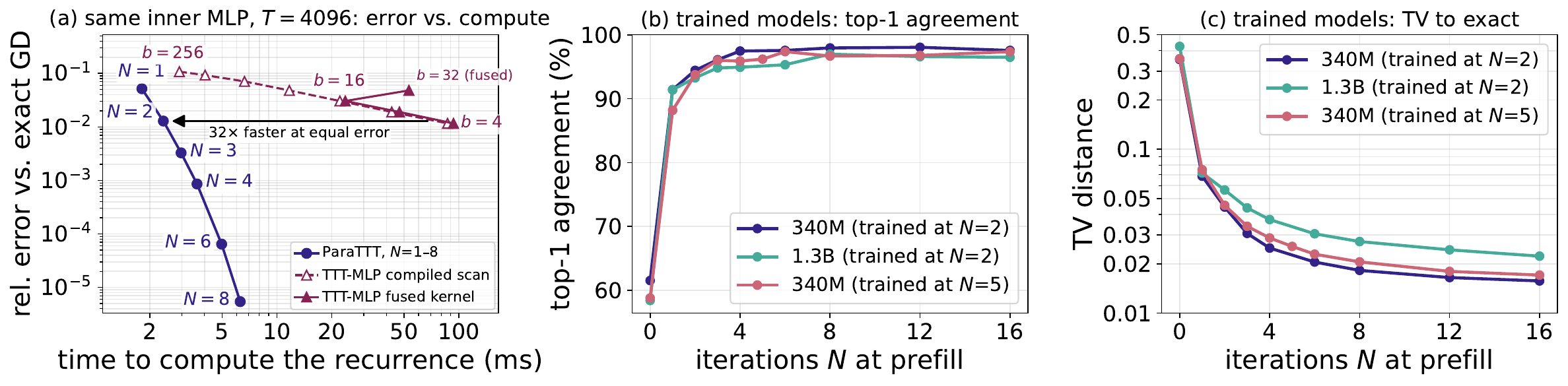}
    \vspace{-8pt}
    \caption{Convergence to the sequential recurrence. (a) One random inner MLP at $T=4096$: relative output error against
    exact online GD versus the time to compute one 340M layer's recurrence, for \method{} at $N=1$--$8$ and TTT-MLP's block form at
    $b=4$--$256$ (filled: fused kernel; hollow: compiled scan); the arrow marks equal error to $N=2$. (b,c) Trained models on real text:
    top-1 agreement and total-variation distance between the prefill at $N$ iterations and the exact recurrence.}
    \label{fig:convergence}
    \vspace{-1em}
\end{figure}

\textbf{Main results.}
\Cref{tab:main} gives the language-modeling results. \method{} has the best perplexity of the recurrent models at both scales, \revminor{the best LAMBADA perplexity among the main model comparisons, and stronger recall than both Gated DeltaNet and Mamba-2 on SWDE and FDA at both scales.} \revs{Recall tasks use exact recurrent decoding, so} \textbf{\rev{the prefill--decode discrepancy does not prevent strong recall}}. \revs{Attention performs best on many recall and long-context tasks. Despite its higher training cost, TTT-MLP matches Gated DeltaNet on WikiText-103, falls far behind on LAMBADA and offers little improvement on recall.} \revminor{We ablate the practical design choices from \cref{sec:method:practical}: the zero-init readout ablation, which removes centering and sets the initial readout to zero, degrades perplexity and recall and greatly reduces the gains from iteration (\cref{fig:perpos-evaln}(c,d)), while \rev{replacing the hidden-layer delta rule with linear attention leaves perplexity similar but substantially degrades recall.} \Cref{tab:centering} separately isolates centering from the learned initialization. \rev{Training with $N=5$ improves all four recall tasks at a proportionally higher cost per step, but raises LAMBADA perplexity from $26.93$ to $28.44$ and reduces 32K NIAH accuracy from $16.0$ to $5.8$.} \revs{The LaCT configuration without SWA performs substantially worse than the published hybrid.}} \revs{\Cref{app:swa} compares LaCT, Gated DeltaNet and \method{} with the same sliding window (\cref{tab:swa-matched}). We find that \method{} is ahead of both LaCT variants on perplexity, most recall tasks, LongBench and NIAH.}

\textbf{Perplexity as a function of token position.}
\revs{\Cref{fig:perpos-evaln}(a,b) shows that \method{} has the lowest perplexity throughout the 4096-token window at both scales, so its advantage over Gated DeltaNet persists beyond the initial tokens. Attention has the worst cold start and catches up with the recurrent models after a few hundred tokens.}

\textbf{Performance as a function of the evaluation-time iteration count.} 
\Cref{fig:perpos-evaln}(c,d) evaluates the 340M \method{} models at iteration counts other than their training $N$. Without iteration, the centered models perform poorly (LAMBADA perplexity above $300$), whereas \revminor{the zero-init readout ablation} performs well but gains little from additional iterations, suggesting that \textbf{\revminor{centering is necessary to unlock the full gains from refinement}}. Models perform best near their training $N$: over-iterating the model trained at $N=2$ degrades performance slowly, while the model trained at $N=5$ exhibits little degradation. This suggests that the model adapts to the finite-$N$ prefill used during training, even though exact recurrent decoding remains strong on the recall tasks. 
\revs{The stronger degradation at 1.3B (\cref{tab:evaln}) motivates training the model with more than one iteration count. We test two recipes: one switches to $N=5$ near the end of training, while the other samples $N$ independently per layer throughout training (details in \cref{app:mismatch}). Both reduce sensitivity to the evaluation-time $N$, with the randomized recipe giving the more stable sweep. At the usual $N=2$ evaluation, the randomized model also improves long-context retrieval and exceeds Gated DeltaNet on NIAH and LongBench at every tested length; the final phase at $N=5$ improves SWDE (\cref{tab:main,tab:longbench}). These gains come with some trade-offs: both variants score lower on FDA, and randomized $N$ raises LAMBADA perplexity relative to the original $N=2$-trained model. The extra training cost is modest (\cref{app:variant-cost}).}


\subsection{Efficiency}
\label{sec:exp:efficiency}

\revs{On MI325X GPUs in bf16, we measure forward+backward time for one token-mixing layer ($d=1024$, 8 heads of dimension 128, batch 8) and full-model training throughput on eight GPUs (\cref{app:kernels}). \Cref{fig:efficiency} compares \method{} at $N\in\{1,2,3,5\}$ with Gated DeltaNet \citep{yang2025gated}, Mamba-2 \citep{dao2024transformers}, FlashAttention, TTT-MLP at its default block size of 16 \citep{sun2024learning}, and published LaCT with a 2048-token sliding window \citep{zhang2025lact}.}

\textbf{Results.}
\method{} scales linearly in $T$ like the linear recurrences it is built from: one layer at $N=2$ costs $4.7\times$ a GDN layer, and each additional iteration adds a fixed number of chunkwise scans. TTT-MLP's block-sequential form is $5.5\times$ slower than \method{} at $T=4096$ and $8.5\times$ at $T=16{,}384$. At the model level, \method{} at $N=2$ trains at $37.9$k tokens/s per node against $7.8$k for TTT-MLP, a $4.9\times$ speedup, while retaining roughly one-third the throughput of the linear baselines. \revs{During decode, the iterations are replaced by one exact recurrent update per token. At batch size 1, \method{}'s latency remains nearly constant across the tested context lengths and is close to Gated DeltaNet's, while remaining less than half that of TTT-MLP and LaCT (\cref{app:decode-latency}).}

\subsection{Convergence analysis}
\label{sec:exp:convergence}

\textbf{\revminor{ParaTTT better approximates the exact online recurrence than TTT-MLP.}}
\revs{\Cref{fig:convergence}(a) compares the output error relative to exact online gradient descent against recurrence evaluation time for \method{} and TTT-MLP on the same random inner MLP ($d=64$, eight heads, step size $0.2$, decay $0.98$). We use the faster of TTT-MLP's fused kernel and compiled scan at each block size. Reducing TTT-MLP's error requires smaller blocks and a longer sequential loop; \method{} instead increases the number of parallel refinements. At $T=4096$, $N=2$ matches TTT-MLP's error at $b\approx5$ in $32\times$ less time, and $N=8$ approaches the bf16 noise floor. \Cref{app:convergence-sweep} repeats this comparison at other sequence lengths.}

\textbf{\revs{Agreement with the exact recurrence in trained models.}}
\revs{\Cref{fig:convergence}(b,c) compares finite-iteration prefill with the exact token-by-token recurrence on 512-token sequences, using top-1 agreement and total-variation distance between next-token distributions. A few iterations give close agreement, with diminishing gains thereafter, despite the trained models lying outside \cref{thm:convergence}'s assumptions. The longer-document tests in \cref{app:mismatch} show a larger discrepancy that increasing $N$ only partly recovers over the tested range. Varying $N$ during training reduces the perplexity penalty from exact recurrent decoding, but does not eliminate the mismatch.}

\section{Conclusion}
\label{sec:conclusion}

We introduce ParaTTT, an algorithm for parallelizing test-time training layers with arbitrarily deep memories through iterative refinement of the inner-loop gradient trajectory. The exact online gradient descent recurrence is a fixed point, and we prove geometric convergence for a two-layer variant under sufficient contraction conditions. Training and prefill use a finite-iteration approximation, while decoding applies exact recurrent updates. With two iterations at 4K context, full 340M models train $4.9\times$ faster than TTT-MLP, while models at 340M and 1.3B achieve strong language-modeling performance and stronger recall on most tasks than the linear recurrent baselines. These results narrow the long-standing tradeoff between the expressivity of nonlinear recurrence and the efficiency of parallel sequence processing.

\section*{Acknowledgements}
This work was supported by the Department of the Air Force Artificial Intelligence Accelerator and was accomplished under Cooperative Agreement Number FA8750-19-2-1000. The views and conclusions contained in this document are those of the authors and should not be interpreted as representing the official policies, either expressed or implied, of the Department of the Air Force or the U.S. Government. The U.S. Government is authorized to reproduce and distribute reprints for Government purposes notwithstanding any copyright notation herein.

\bibliography{references}

@inproceedings{vaswani2017attention,
  title={Attention is all you need},
  author={Vaswani, Ashish and Shazeer, Noam and Parmar, Niki and Uszkoreit, Jakob and Jones, Llion and Gomez, Aidan N and Kaiser, {\L}ukasz and Polosukhin, Illia},
  booktitle={Advances in Neural Information Processing Systems},
  year={2017},
  url={https://proceedings.neurips.cc/paper/2017/hash/3f5ee243547dee91fbd053c1c4a845aa-Abstract.html}
}

@inproceedings{katharopoulos2020transformers,
  title={Transformers are {RNN}s: Fast autoregressive transformers with linear attention},
  author={Katharopoulos, Angelos and Vyas, Apoorv and Pappas, Nikolaos and Fleuret, Fran{\c{c}}ois},
  booktitle={International Conference on Machine Learning},
  year={2020},
  url={https://proceedings.mlr.press/v119/katharopoulos20a.html}
}

@inproceedings{schlag2021linear,
  title={Linear transformers are secretly fast weight programmers},
  author={Schlag, Imanol and Irie, Kazuki and Schmidhuber, J{\"u}rgen},
  booktitle={International Conference on Machine Learning},
  year={2021},
  url={https://proceedings.mlr.press/v139/schlag21a.html}
}

@inproceedings{yang2024gla,
  title={Gated linear attention transformers with hardware-efficient training},
  author={Yang, Songlin and Wang, Bailin and Shen, Yikang and Panda, Rameswar and Kim, Yoon},
  booktitle={International Conference on Machine Learning},
  year={2024},
  url={https://proceedings.mlr.press/v235/yang24ab.html}
}

@inproceedings{yang2024parallelizing,
  title={Parallelizing linear transformers with the delta rule over sequence length},
  author={Yang, Songlin and Wang, Bailin and Zhang, Yu and Shen, Yikang and Kim, Yoon},
  booktitle={Advances in Neural Information Processing Systems},
  year={2024},
  url={https://arxiv.org/abs/2406.06484}
}

@inproceedings{yang2025gated,
  title={Gated delta networks: Improving {Mamba2} with delta rule},
  author={Yang, Songlin and Kautz, Jan and Hatamizadeh, Ali},
  booktitle={International Conference on Learning Representations},
  year={2025},
  url={https://openreview.net/forum?id=r8H7xhYPwz}
}

@inproceedings{dao2024transformers,
  title={Transformers are {SSM}s: Generalized models and efficient algorithms through structured state space duality},
  author={Dao, Tri and Gu, Albert},
  booktitle={International Conference on Machine Learning},
  year={2024},
  url={https://proceedings.mlr.press/v235/dao24a.html}
}

@inproceedings{sun2024learning,
  title={Learning to (learn at test time): {RNN}s with expressive hidden states},
  author={Sun, Yu and Li, Xinhao and Dalal, Karan and Xu, Jiarui and Vikram, Arjun and Zhang, Genghan and Dubois, Yann and Chen, Xinlei and Wang, Xiaolong and Koyejo, Sanmi and Hashimoto, Tatsunori and Guestrin, Carlos},
  year={2025},
  url={https://proceedings.mlr.press/v267/sun25h.html},
  booktitle={International Conference on Machine Learning}
}

@inproceedings{behrouz2024titans,
  title={Titans: Learning to memorize at test time},
  author={Behrouz, Ali and Zhong, Peilin and Mirrokni, Vahab},
  year={2025},
  url={https://proceedings.neurips.cc/paper_files/paper/2025/hash/a4ca07aa108036f80cbb5b82285fd4b1-Abstract-Conference.html},
  booktitle={Advances in Neural Information Processing Systems}
}

@inproceedings{zhang2025lact,
  title={Test-time training done right},
  author={Zhang, Tianyuan and Bi, Sai and Hong, Yicong and Zhang, Kai and Luan, Fujun and Yang, Songlin and Sunkavalli, Kalyan and Freeman, William T and Tan, Hao},
  year={2026},
  url={https://proceedings.iclr.cc/paper_files/paper/2026/hash/ffbfff81fb78e4bb558273b91e12e318-Abstract-Conference.html},
  booktitle={International Conference on Learning Representations}
}

@inproceedings{lim2024parallelizing,
  title={Parallelizing non-linear sequential models over the sequence length},
  author={Lim, Yi Heng and Zhu, Qi and Selfridge, Joshua and Kasim, Muhammad Firmansyah},
  booktitle={International Conference on Learning Representations},
  year={2024},
  url={https://openreview.net/forum?id=E34AlVLN0v}
}

@inproceedings{danieli2025pararnn,
  title={{ParaRNN}: Unlocking parallel training of nonlinear {RNN}s for large language models},
  author={Danieli, Federico and Rodr{\'i}guez, Pau and Sarabia, Miguel and Suau, Xavier and Zappella, Luca},
  booktitle={International Conference on Learning Representations},
  year={2026},
  url={https://openreview.net/forum?id=mX8b64iUaa}
}

@inproceedings{bai2019deep,
  title={Deep equilibrium models},
  author={Bai, Shaojie and Kolter, J Zico and Koltun, Vladlen},
  booktitle={Advances in Neural Information Processing Systems},
  year={2019},
  url={https://proceedings.neurips.cc/paper/2019/hash/01386bd6d8e091c2ab4c7c7de644d37b-Abstract.html}
}

@inproceedings{arora2024simple,
  title={Simple linear attention language models balance the recall-throughput tradeoff},
  author={Arora, Simran and Eyuboglu, Sabri and Zhang, Michael and Timalsina, Aman and Alberti, Silas and Zou, James and Rudra, Atri and R{\'e}, Christopher},
  booktitle={International Conference on Machine Learning},
  year={2024},
  url={https://proceedings.mlr.press/v235/arora24a.html}
}

@inproceedings{hsieh2024ruler,
  title={{RULER}: What's the real context size of your long-context language models?},
  author={Hsieh, Cheng-Ping and Sun, Simeng and Kriman, Samuel and Acharya, Shantanu and Rekesh, Dima and Jia, Fei and Zhang, Yang and Ginsburg, Boris},
  year={2024},
  url={https://openreview.net/forum?id=kIoBbc76Sy},
  booktitle={First Conference on Language Modeling (COLM)}
}

@inproceedings{peng2024yarn,
  title={{YaRN}: Efficient context window extension of large language models},
  author={Peng, Bowen and Quesnelle, Jeffrey and Fan, Honglu and Shippole, Enrico},
  booktitle={International Conference on Learning Representations},
  year={2024},
  url={https://openreview.net/forum?id=wHBfxhZu1u}
}

@misc{gao2023lmeval,
  title={A framework for few-shot language model evaluation},
  author={Gao, Leo and Tow, Jonathan and Abbasi, Baber and Biderman, Stella and Black, Sid and DiPofi, Anthony and Foster, Charles and Golding, Laurence and Hsu, Jeffrey and Le Noac'h, Alain and Li, Haonan and McDonell, Kyle and Muennighoff, Niklas and Ociepa, Chris and Phang, Jason and Reynolds, Laria and Schoelkopf, Hailey and Skowron, Aviya and Sutawika, Lintang and Tang, Eric and Thite, Anish and Wang, Ben and Wang, Kevin and Zou, Andy},
  publisher={Zenodo},
  version={v0.4.0},
  doi={10.5281/zenodo.10256836},
  year={2023},
  url={https://zenodo.org/records/10256836}
}

@inproceedings{chizat2019lazy,
  title={On lazy training in differentiable programming},
  author={Chizat, L{\'e}na{\"\i}c and Oyallon, Edouard and Bach, Francis},
  booktitle={Advances in Neural Information Processing Systems},
  year={2019},
  url={https://proceedings.neurips.cc/paper/2019/hash/ae614c557843b1df326cb29c57225459-Abstract.html}
}

@inproceedings{woodworth2020kernel,
  title={Kernel and rich regimes in overparametrized models},
  author={Woodworth, Blake and Gunasekar, Suriya and Lee, Jason D and Moroshko, Edward and Savarese, Pedro and Golan, Itay and Soudry, Daniel and Srebro, Nathan},
  booktitle={Conference on Learning Theory},
  year={2020},
  url={https://proceedings.mlr.press/v125/woodworth20a.html}
}

@inproceedings{hu2020simple,
  title={Simple and effective regularization methods for training on noisily labeled data with generalization guarantee},
  author={Hu, Wei and Li, Zhiyuan and Yu, Dingli},
  booktitle={International Conference on Learning Representations},
  year={2020},
  url={https://openreview.net/forum?id=Hke3gyHYwH}
}

@inproceedings{widrow1960adaptive,
  title={Adaptive switching circuits},
  author={Widrow, Bernard and Hoff, Jr., Marcian E.},
  booktitle={1960 IRE WESCON Convention Record, Part 4},
  pages={96--104},
  year={1960},
  url={https://isl.stanford.edu/~widrow/papers/c1960adaptiveswitching.pdf}
}

@article{elman1990finding,
  title={Finding structure in time},
  author={Elman, Jeffrey L.},
  journal={Cognitive Science},
  volume={14},
  number={2},
  pages={179--211},
  year={1990},
  url={https://doi.org/10.1207/s15516709cog1402_1}
}

@article{hochreiter1997long,
  title={Long short-term memory},
  author={Hochreiter, Sepp and Schmidhuber, J{\"u}rgen},
  journal={Neural Computation},
  volume={9},
  number={8},
  pages={1735--1780},
  year={1997},
  url={https://doi.org/10.1162/neco.1997.9.8.1735}
}

@inproceedings{cho2014learning,
  title={Learning phrase representations using {RNN} encoder--decoder for statistical machine translation},
  author={Cho, Kyunghyun and van Merri{\"e}nboer, Bart and Gulcehre, Caglar and Bahdanau, Dzmitry and Bougares, Fethi and Schwenk, Holger and Bengio, Yoshua},
  booktitle={Proceedings of the 2014 Conference on Empirical Methods in Natural Language Processing (EMNLP)},
  year={2014},
  url={https://aclanthology.org/D14-1179/},
  pages={1724--1734},
  doi={10.3115/v1/D14-1179}
}

@inproceedings{beck2024xlstm,
  title={{xLSTM}: Extended long short-term memory},
  author={Beck, Maximilian and P{\"o}ppel, Korbinian and Spanring, Markus and Auer, Andreas and Prudnikova, Oleksandra and Kopp, Michael and Klambauer, G{\"u}nter and Brandstetter, Johannes and Hochreiter, Sepp},
  booktitle={Advances in Neural Information Processing Systems},
  volume={37},
  year={2024},
  url={https://proceedings.neurips.cc/paper_files/paper/2024/hash/c2ce2f2701c10a2b2f2ea0bfa43cfaa3-Abstract-Conference.html}
}

@inproceedings{liu2023transformers,
  title={Transformers learn shortcuts to automata},
  author={Liu, Bingbin and Ash, Jordan T. and Goel, Surbhi and Krishnamurthy, Akshay and Zhang, Cyril},
  booktitle={International Conference on Learning Representations},
  year={2023},
  url={https://openreview.net/forum?id=De4FYqjFueZ}
}

@inproceedings{merrill2024illusion,
  title={The illusion of state in state-space models},
  author={Merrill, William and Petty, Jackson and Sabharwal, Ashish},
  booktitle={International Conference on Machine Learning},
  year={2024},
  url={https://proceedings.mlr.press/v235/merrill24a.html}
}

@article{merrill2023parallelism,
  title={The parallelism tradeoff: Limitations of log-precision transformers},
  author={Merrill, William and Sabharwal, Ashish},
  journal={Transactions of the Association for Computational Linguistics},
  volume={11},
  pages={531--545},
  year={2023},
  url={https://aclanthology.org/2023.tacl-1.31/}
}

@inproceedings{grazzi2024unlocking,
  title={Unlocking state-tracking in linear {RNNs} through negative eigenvalues},
  author={Grazzi, Riccardo and Siems, Julien and Zela, Arber and Franke, J{\"o}rg K. H. and Hutter, Frank and Pontil, Massimiliano},
  booktitle={International Conference on Learning Representations},
  year={2025},
  url={https://proceedings.iclr.cc/paper_files/paper/2025/hash/5a0ce3abb720b740419e193c87afd080-Abstract-Conference.html}
}

@inproceedings{tumma2026preconditioned,
  title={Preconditioned {DeltaNet}: Curvature-aware sequence modeling for linear recurrences},
  author={Tumma, Neehal and Loo, Noel and Rus, Daniela},
  year={2026},
  url={https://icml.cc/virtual/2026/poster/63733},
  booktitle={International Conference on Machine Learning}
}

@article{arora2024just,
  title={Just read twice: Closing the recall gap for recurrent language models},
  author={Arora, Simran and Timalsina, Aman and Singhal, Aaryan and Spector, Benjamin and Eyuboglu, Sabri and Zhao, Xinyi and Rao, Ashish and Rudra, Atri and R{\'e}, Christopher},
  journal={arXiv preprint arXiv:2407.05483},
  year={2024},
  url={https://arxiv.org/abs/2407.05483}
}

@article{mishra2026m2rnn,
  title={{M$^2$RNN}: Non-linear {RNN}s with matrix-valued states for scalable language modeling},
  author={Mishra, Mayank and Tan, Shawn and Stoica, Ion and Gonzalez, Joseph E. and Dao, Tri},
  journal={arXiv preprint arXiv:2603.14360},
  year={2026},
  url={https://arxiv.org/abs/2603.14360}
}

@article{hatamizadeh2026gdn2,
  title={Gated {DeltaNet}-2: Decoupling erase and write in linear attention},
  author={Hatamizadeh, Ali and Choi, Yejin and Kautz, Jan},
  journal={arXiv preprint arXiv:2605.22791},
  year={2026},
  url={https://arxiv.org/abs/2605.22791}
}

@inproceedings{lahoti2026mamba3,
  title={Mamba-3: Improved sequence modeling using state space principles},
  author={Lahoti, Aakash and Li, Kevin Y. and Chen, Berlin and Wang, Caitlin and Bick, Aviv and Kolter, J. Zico and Dao, Tri and Gu, Albert},
  year={2026},
  booktitle={International Conference on Learning Representations},
  url={https://proceedings.iclr.cc/paper_files/paper/2026/hash/8abd2043b71a074278d5f687947bff9c-Abstract-Conference.html}
}

@inproceedings{peng2024eagle,
  title={Eagle and {Finch}: {RWKV} with matrix-valued states and dynamic recurrence},
  author={Peng, Bo and Goldstein, Daniel and Anthony, Quentin and Albalak, Alon and Alcaide, Eric and Biderman, Stella and Cheah, Eugene and Ferdinan, Teddy and Hou, Haowen and {Kranthi Kiran GV} and Krishna, Satyapriya and McClelland, Jr., Ronald and Muennighoff, Niklas and Obeid, Fares and Saito, Atsushi and Song, Guangyu and Tu, Haoqin and Zhang, Ruichong and Zhao, Bingchen and Zhao, Qihang and Zhu, Jian and Zhu, Rui-Jie},
  year={2024},
  booktitle={First Conference on Language Modeling (COLM)},
  url={https://openreview.net/pdf/9f3d829735b231725665ff226a4f732293ad4aa8.pdf}
}

@inproceedings{peng2025rwkv7,
  title={{RWKV}-7 ``{Goose}'' with expressive dynamic state evolution},
  author={Peng, Bo and Zhang, Ruichong and Goldstein, Daniel and Alcaide, Eric and Du, Xingjian and Hou, Haowen and Lin, Jiaju and Liu, Jiaxing and Lu, Janna and Merrill, William and Song, Guangyu and Tan, Kaifeng and Utpala, Saiteja and Wilce, Nathan and Wind, Johan S. and Wu, Tianyi and Wuttke, Daniel and Zhou-Zheng, Christian},
  year={2025},
  booktitle={Second Conference on Language Modeling (COLM)},
  url={https://openreview.net/forum?id=ayB1PACN5j}
}

@inproceedings{huang2026mdn,
  title={{MDN}: Parallelizing stepwise momentum for delta linear attention},
  author={Huang, Yulong and Liu, Xiang and Huang, Hongxiang and Lin, Xiaopeng and Liu, Zunchang and Chu, Xiaowen and Xie, Zeke and Cheng, Bojun},
  year={2026},
  booktitle={International Conference on Machine Learning},
  url={https://icml.cc/virtual/2026/poster/63901}
}

@article{kimi2025linear,
  title={Kimi {Linear}: An expressive, efficient attention architecture},
  author={{Kimi Team}},
  journal={arXiv preprint arXiv:2510.26692},
  year={2025},
  url={https://arxiv.org/abs/2510.26692}
}

@inproceedings{siems2025deltaproduct,
  title={{DeltaProduct}: Improving state-tracking in linear {RNN}s via {Householder} products},
  author={Siems, Julien and Carstensen, Timur and Zela, Arber and Hutter, Frank and Pontil, Massimiliano and Grazzi, Riccardo},
  booktitle={Advances in Neural Information Processing Systems},
  year={2025},
  url={https://proceedings.neurips.cc/paper_files/paper/2025/hash/e1ea2520fbf9cd1600b287dde67e0a3c-Abstract-Conference.html}
}

@inproceedings{vonoswald2026mesanet,
  title={{MesaNet}: Sequence modeling by locally optimal test-time training},
  author={von Oswald, Johannes and Scherrer, Nino and Kobayashi, Seijin and Versari, Luca and Yang, Songlin and Schlegel, Maximilian and Maile, Kaitlin and Schimpf, Yanick and Sieberling, Oliver and Meulemans, Alexander and Saurous, Rif A. and Lajoie, Guillaume and Frenkel, Charlotte and Pascanu, Razvan and Ag{\"u}era y Arcas, Blaise and Sacramento, Jo{\~a}o},
  booktitle={International Conference on Learning Representations},
  year={2026},
  url={https://proceedings.iclr.cc/paper_files/paper/2026/hash/2c3d5a4eff20d24390a445762d61bfbb-Abstract-Conference.html}
}

@inproceedings{hu2025comba,
  title={Improving bilinear {RNN}s with closed-loop control},
  author={Hu, Jiaxi and Pan, Yongqi and Du, Jusen and Lan, Disen and Tang, Xiaqiang and Wen, Qingsong and Liang, Yuxuan and Sun, Weigao},
  year={2025},
  booktitle={Advances in Neural Information Processing Systems},
  url={https://proceedings.neurips.cc/paper_files/paper/2025/hash/9a439efaa34fe37177eba00737624824-Abstract-Conference.html}
}

@article{karami2025lattice,
  title={Lattice: Learning to efficiently compress the memory},
  author={Karami, Mahdi and Pascanu, Razvan and Mirrokni, Vahab},
  journal={arXiv preprint arXiv:2504.05646},
  year={2025},
  url={https://arxiv.org/abs/2504.05646}
}

@article{wang2025testtime,
  title={Test-time regression: a unifying framework for designing sequence models with associative memory},
  author={Wang, Ke Alexander and Shi, Jiaxin and Fox, Emily B.},
  journal={arXiv preprint arXiv:2501.12352},
  year={2025},
  url={https://arxiv.org/abs/2501.12352}
}

@article{behrouz2025atlas,
  title={{ATLAS}: Learning to optimally memorize the context at test time},
  author={Behrouz, Ali and Li, Zeman and Kacham, Praneeth and Daliri, Majid and Deng, Yuan and Zhong, Peilin and Razaviyayn, Meisam and Mirrokni, Vahab},
  journal={arXiv preprint arXiv:2505.23735},
  year={2025},
  url={https://arxiv.org/abs/2505.23735}
}

@inproceedings{feng2026inplace,
  title={In-place test-time training},
  author={Feng, Guhao and Luo, Shengjie and Hua, Kai and Zhang, Ge and Huang, Wenhao and He, Di and Cai, Tianle},
  year={2026},
  booktitle={International Conference on Learning Representations},
  url={https://proceedings.iclr.cc/paper_files/paper/2026/hash/ba6f5879d7f382241316f7dda7e6b8ac-Abstract-Conference.html}
}

@article{tandon2025e2ettt,
  title={End-to-end test-time training for long context},
  author={Tandon, Arnuv and Dalal, Karan and Li, Xinhao and Koceja, Daniel and R{\o}d, Marcel and Buchanan, Sam and Wang, Xiaolong and Leskovec, Jure and Koyejo, Sanmi and Hashimoto, Tatsunori and Guestrin, Carlos and McCaleb, Jed and Choi, Yejin and Sun, Yu},
  journal={arXiv preprint arXiv:2512.23675},
  year={2025},
  url={https://arxiv.org/abs/2512.23675}
}

@inproceedings{gonzalez2024elk,
  title={Towards scalable and stable parallelization of nonlinear {RNN}s},
  author={Gonzalez, Xavier and Warrington, Andrew and Smith, Jimmy T. H. and Linderman, Scott W.},
  booktitle={Advances in Neural Information Processing Systems},
  year={2024},
  url={https://arxiv.org/abs/2407.19115}
}

@inproceedings{gonzalez2025predictability,
  title={Predictability enables parallelization of nonlinear state space models},
  author={Gonzalez, Xavier and Kozachkov, Leo and Zoltowski, David M. and Clarkson, Kenneth L. and Linderman, Scott W.},
  booktitle={Advances in Neural Information Processing Systems},
  year={2025},
  url={https://proceedings.neurips.cc/paper_files/paper/2025/hash/1b7a239766f041abcd4d286ec5548037-Abstract-Conference.html}
}

@inproceedings{merrill2025depth,
  title={A little depth goes a long way: The expressive power of log-depth transformers},
  author={Merrill, William and Sabharwal, Ashish},
  booktitle={Advances in Neural Information Processing Systems},
  year={2025},
  url={https://proceedings.neurips.cc/paper_files/paper/2025/hash/88dd7aa6979e352fda7c4952ca8eac59-Abstract-Conference.html}
}

@inproceedings{schoene2025implicit,
  title={Implicit language models are {RNN}s: Balancing parallelization and expressivity},
  author={Sch{\"o}ne, Mark and Rahmani, Babak and Kremer, Heiner and Falck, Fabian and Ballani, Hitesh and Gladrow, Jannes},
  booktitle={International Conference on Machine Learning},
  year={2025},
  url={https://proceedings.mlr.press/v267/schone25a.html}
}

@inproceedings{dehghani2019universal,
  title={Universal transformers},
  author={Dehghani, Mostafa and Gouws, Stephan and Vinyals, Oriol and Uszkoreit, Jakob and Kaiser, {\L}ukasz},
  booktitle={International Conference on Learning Representations},
  year={2019},
  url={https://arxiv.org/abs/1807.03819}
}

@inproceedings{giannou2023looped,
  title={Looped transformers as programmable computers},
  author={Giannou, Angeliki and Rajput, Shashank and Sohn, Jy-yong and Lee, Kangwook and Lee, Jason D. and Papailiopoulos, Dimitris},
  booktitle={International Conference on Machine Learning},
  year={2023},
  url={https://proceedings.mlr.press/v202/giannou23a.html}
}

@inproceedings{saunshi2025reasoning,
  title={Reasoning with latent thoughts: On the power of looped transformers},
  author={Saunshi, Nikunj and Dikkala, Nishanth and Li, Zhiyuan and Kumar, Sanjiv and Reddi, Sashank J.},
  booktitle={International Conference on Learning Representations},
  year={2025},
  url={https://proceedings.iclr.cc/paper_files/paper/2025/hash/2676109d49d1eb26d6bc584a8f556305-Abstract-Conference.html}
}

@inproceedings{geiping2025scaling,
  title={Scaling up test-time compute with latent reasoning: A recurrent depth approach},
  author={Geiping, Jonas and McLeish, Sean and Jain, Neel and Kirchenbauer, John and Singh, Siddharth and Bartoldson, Brian R. and Kailkhura, Bhavya and Bhatele, Abhinav and Goldstein, Tom},
  booktitle={Advances in Neural Information Processing Systems},
  year={2025},
  url={https://proceedings.neurips.cc/paper_files/paper/2025/hash/3b01972cf31e6fa0fe29e4b8b5c2a0a1-Abstract-Conference.html}
}

@article{zhu2025ouro,
  title={Scaling latent reasoning via looped language models},
  author={Zhu, Rui-Jie and Wang, Zixuan and Hua, Kai and Zhang, Tianyu and Li, Ziniu and Que, Haoran and Wei, Boyi and Wen, Zixin and Yin, Fan and Xing, He and Li, Lu and Shi, Jiajun and Ma, Kaijing and Li, Shanda and Kergan, Taylor and Smith, Andrew and Qu, Xingwei and Hui, Mude and Wu, Bohong and Min, Qiyang and Huang, Hongzhi and Zhou, Xun and Ye, Wei and Liu, Jiaheng and Yang, Jian and Shi, Yunfeng and Lin, Chenghua and Zhao, Enduo and Cai, Tianle and Zhang, Ge and Huang, Wenhao and Bengio, Yoshua and Eshraghian, Jason},
  journal={arXiv preprint arXiv:2510.25741},
  year={2025},
  url={https://arxiv.org/abs/2510.25741}
}

@article{movahedi2026fixedpoint,
  title={Fixed-point reasoners: Stable and adaptive deep looped transformers},
  author={Movahedi, Sajad and Milovanovi{\'c}, Vera and Feigin, Shlomo Libo and Theus, Alexander and Hofmann, Thomas and Boeva, Valentina and Rusch, T. Konstantin and Orvieto, Antonio},
  journal={arXiv preprint arXiv:2606.18206},
  year={2026},
  url={https://arxiv.org/abs/2606.18206}
}

@inproceedings{liu2026secretly,
  title={Test-time training with {KV} binding is secretly linear attention},
  author={Liu, Junchen and Elflein, Sven and Litany, Or and Gojcic, Zan and Li, Ruilong},
  year={2026},
  booktitle={International Conference on Machine Learning},
  url={https://icml.cc/virtual/2026/poster/64516}
}

@article{werbos1990bptt,
  title={Backpropagation through time: what it does and how to do it},
  author={Werbos, Paul J.},
  journal={Proceedings of the IEEE},
  volume={78},
  number={10},
  pages={1550--1560},
  year={1990},
  url={https://doi.org/10.1109/5.58337}
}

@article{rumelhart1986learning,
  title={Learning representations by back-propagating errors},
  author={Rumelhart, David E. and Hinton, Geoffrey E. and Williams, Ronald J.},
  journal={Nature},
  volume={323},
  number={6088},
  pages={533--536},
  year={1986},
  url={https://doi.org/10.1038/323533a0}
}

@article{labovich2026stability,
  title={Stability and generalization in looped transformers},
  author={Labovich, Asher},
  journal={arXiv preprint arXiv:2604.15259},
  year={2026},
  url={https://arxiv.org/abs/2604.15259}
}

@inproceedings{hao2025coconut,
  title={Training large language models to reason in a continuous latent space},
  author={Hao, Shibo and Sukhbaatar, Sainbayar and Su, DiJia and Li, Xian and Hu, Zhiting and Weston, Jason and Tian, Yuandong},
  booktitle={Second Conference on Language Modeling (COLM)},
  year={2025},
  url={https://arxiv.org/abs/2412.06769}
}

@inproceedings{wu2025pccot,
  title={Parallel continuous chain-of-thought with {J}acobi iteration},
  author={Wu, Haoyi and Teng, Zhihao and Tu, Kewei},
  booktitle={Proceedings of the 2025 Conference on Empirical Methods in Natural Language Processing (EMNLP)},
  pages={914--926},
  year={2025},
  url={https://aclanthology.org/2025.emnlp-main.47/},
  doi={10.18653/v1/2025.emnlp-main.47}
}

@article{cai2026t2mlr,
  title={{T$^2$MLR}: Transformer with temporal middle-layer recurrence},
  author={Cai, Ziyang and Zhu, Xingyu and Dong, Yihe and He, Yinghui and Arora, Sanjeev},
  journal={arXiv preprint arXiv:2607.15178},
  year={2026},
  url={https://arxiv.org/abs/2607.15178}
}

@article{wang2026fullbandwidth,
  title={Full-bandwidth transformer},
  author={Wang, Xi and Cai, Ziyang and Zhan, Zheng and Dong, Harry and Fan, Ying and de Rosa, Gustavo and Pearce, Tim and Langford, John},
  journal={arXiv preprint arXiv:2608.08888},
  year={2026},
  url={https://arxiv.org/abs/2608.08888}
}

@misc{parnichkun2025quantifyingmemoryutilizationeffective,
      title={Quantifying Memory Utilization with Effective State-Size}, 
      author={Rom N. Parnichkun and Neehal Tumma and Armin W. Thomas and Alessandro Moro and Qi An and Taiji Suzuki and Atsushi Yamashita and Michael Poli and Stefano Massaroli},
      year={2025},
      eprint={2504.19561},
      archivePrefix={arXiv},
      primaryClass={cs.LG},
      url={https://arxiv.org/abs/2504.19561}, 
}

@misc{tumma2023leveraginglowranksparserecurrent,
      title={Leveraging Low-Rank and Sparse Recurrent Connectivity for Robust Closed-Loop Control}, 
      author={Neehal Tumma and Mathias Lechner and Noel Loo and Ramin Hasani and Daniela Rus},
      year={2023},
      eprint={2310.03915},
      archivePrefix={arXiv},
      primaryClass={cs.LG},
      url={https://arxiv.org/abs/2310.03915}, 
}
\bibliographystyle{iclr2027_conference}

\newpage
\appendix
\begingroup
\raggedbottom
\section{Extended related work}
\label{app:related}

\paragraph{Linear recurrences.}
The lineage in \cref{sec:background:linear} runs from linear attention \citep{katharopoulos2020transformers} through gated
linear attention \citep{yang2024gla}, Mamba-2 \citep{dao2024transformers}, DeltaNet \citep{schlag2021linear,yang2024parallelizing}
and Gated DeltaNet \citep{yang2025gated}; \citet{wang2025testtime} give the unifying view of all of these as test-time regression.
The RWKV family follows the same path from a vector-valued to a matrix-valued state \citep{peng2024eagle} and, in RWKV-7, to a
generalized delta rule with vector-valued gates and in-context learning rates \citep{peng2025rwkv7}. The most recent linear
recurrences refine the delta rule's write and forget. KDA \citep{kimi2025linear} replaces Gated DeltaNet's scalar decay by a
channel-wise one; Gated DeltaNet-2 \citep{hatamizadeh2026gdn2} further decouples the erase gate from the write gate, recovering
both KDA and Gated DeltaNet as special cases; Mamba-3 \citep{lahoti2026mamba3} revisits the discretization of the state-space
recurrence and adds complex-valued and multi-input multi-output updates; DeltaProduct \citep{siems2025deltaproduct} takes several
delta steps per token so that the transition is a product of generalized Householder matrices; Comba \citep{hu2025comba} adds output feedback
to the bilinear state transition; Lattice \citep{karami2025lattice} writes only the component of a token orthogonal to the stored
memory slot, with an approximate chunkwise evaluation; Momentum DeltaNet \citep{huang2026mdn} gives the online gradient step a momentum term; Preconditioned DeltaNet
\citep{tumma2026preconditioned} preconditions it with an online estimate of the curvature; and MesaNet
\citep{vonoswald2026mesanet} solves the in-context regression to optimality with a conjugate-gradient solver. These methods retain matrix-valued memories \revminor{\citep{parnichkun2025quantifyingmemoryutilizationeffective}} but use different update and evaluation schemes: momentum introduces an auxiliary
state, MesaNet adds a per-query solve, and Lattice approximates a state-dependent update. \method{} uses chunkwise linear recurrent
kernels as building blocks. Extending its readout to these other updates is a possible direction, but would require checking the
corresponding solver and gradient computation.

\paragraph{Test-time training.}
TTT-Linear and TTT-MLP \citep{sun2024learning} introduced the layer whose state is the weights of a small model trained online,
evaluated in the mini-batch form of \cref{sec:background:ttt}. It is worth writing this form out, since it is the approximation that
\method{} removes. Let a block consist of tokens $s+1,\dots,s+b$ and let $\bW_s$ be the weights entering it (TTT-MLP uses no decay,
so we drop $\alpha$). Exact online gradient descent evaluates the gradient of token $j$ at the weights left by token $j-1$,
$\bW_j=\bW_{j-1}-\eta_j\,\nabla\ell_j(\bW_{j-1})$; the block form evaluates every gradient of the block at the weights that entered it
and scores token $t$ with the causal partial sum of these \emph{stale} gradients,
\begin{equation}
    \bW^{\text{blk}}_t \;=\; \bW_s-\sum_{j=s+1}^{t}\eta_j\,\nabla\ell_j(\bW_s),\qquad s<t\le s+b .
    \label{eq:block-form}
\end{equation}
For a squared-loss MLP with a linear readout, the readout update has an attention-style form inside the block. In the next display,
$\bW$ denotes the readout weights only. With $\bphi_j$ the features of token $j$, computed with the
block-start hidden weights, the gradient at $\bW_s$ is the outer product $(\bW_s\bphi_j-\bv_j)\bphi_j^\top$ of a residual with a
feature. Let $\bphi_{\bq,t}$ denote the query features under the causally updated hidden weights at token $t$; then the output is
\begin{equation}
    \bo_t \;=\; \bW^{\text{blk}}_t\bphi_{\bq,t}\;=\;\bW_s\bphi_{\bq,t}-\sum_{j=s+1}^{t}\eta_j\,\big(\bW_s\bphi_j-\bv_j\big)\,\big(\bphi_j^\top\bphi_{\bq,t}\big),
    \label{eq:block-la}
\end{equation}
an attention-style readout with keys $\bphi_j$, values the residuals $\bW_s\bphi_j-\bv_j$ and queries $\bphi_{\bq,t}$, so the readout sum can be evaluated with matrix multiplications once the features are given. The price is in the residuals: exact online descent would use the preceding readout weights and the current online features,
so that the write of token $j$ erases what earlier tokens of the block stored under similar keys (the delta rule), whereas the block
form's residuals see none of the block's earlier writes and the writes simply add up, as in linear attention without a delta rule.
The hidden-layer gradients are also evaluated at block-start weights. This freezes the features used to compute those gradients,
but does not freeze the hidden weights or query features used for each token's output. At $b=1$ the block form recovers online
descent; at $b=T$ all gradient evaluations share the initial weights. TTT-MLP's default $b=16$ and LaCT's larger blocks trade
sequential depth against update frequency, and \cref{fig:convergence}(a) measures the resulting error against the exact trajectory.
\method{} instead targets the $b=1$ recurrence by iterating on the gradients.

\citet{liu2026secretly} study LaCT-style and vision TTT layers. They report that adding inner steps only at inference can hurt
downstream performance, that retraining with gradient ascent can preserve performance, and that replacing queries with keys can
have little effect in their experiments. Their attention-style representation allows updated hidden features; it is not limited
to a frozen-feature model. They also obtain parallel variants through further simplifications, including restricting adaptation
to the final layer. These findings motivate testing which parts of nonlinear adaptation matter. Our experiments address this
question for the online recurrence, but do not establish that the block approximation explains their observations.

Titans \citep{behrouz2024titans} adds momentum and weight decay to
the inner optimizer and ATLAS \citep{behrouz2025atlas} optimizes the memory over a window of past tokens rather than the current
one; LaCT \citep{zhang2025lact} makes the block very large so that the inner update is a dense matrix multiplication, and pairs it
with a sliding-window attention branch (\cref{app:swa}). Two recent lines move the fast weights into the backbone. In-place TTT
\citep{feng2026inplace} treats the down-projection of every MLP block as the fast weights, trains it with an objective aligned with
next-token prediction, and updates it chunk-wise; end-to-end TTT \citep{tandon2025e2ettt} keeps a standard transformer with
sliding-window attention and adapts selected MLP weights on the context by next-token prediction, meta-learned so that the
test-time updates help. In its reported configuration, adaptation is restricted to the last quarter of the blocks. Both inherit the block structure: the weights are frozen inside a chunk or a window and updated between
them. \method{} is orthogonal to these choices of what to adapt and with which loss; it changes how the inner trajectory of a
nonlinear memory is computed, so that every token's update is taken at the weights left by the previous token, and it does so with
the parallel kernels of the linear recurrences above.

\paragraph{Sequential nonlinear RNNs.}
Nonlinear recurrent networks \revminor{\citep{elman1990finding,hochreiter1997long,cho2014learning,tumma2023leveraginglowranksparserecurrent}} are trained by backpropagation through
time (BPTT) \citep{rumelhart1986learning,werbos1990bptt}, which unrolls the recurrence over the sequence and runs the backward pass
through every step in reverse. Their straightforward forward evaluation and BPTT are sequential over tokens, limiting sequence parallelism compared with
attention \citep{vaswani2017attention}. The
sLSTM cell of xLSTM \citep{beck2024xlstm} keeps a sequential nonlinear transition next to a parallel linear one. M$^2$RNN
\citep{mishra2026m2rnn} is the most direct recent attempt to make the sequential path competitive: it keeps a matrix-valued state,
updated by a $\tanh$ of a linear map of the previous state plus an outer-product write with a learned forget gate, so that the
state size scales as it does in linear attention, and it evaluates the recurrence sequentially in a Triton kernel that parallelizes
over batch and heads while looping over time; the backward pass is full BPTT over the 4096-token training sequence, recomputing the
forward states and running the reverse loop over every step. The nonlinear transition prevents direct evaluation by the linear scans used by SSMs. The paper reports perfect length
generalization on its evaluated state-tracking tasks, with training throughput below its linear recurrent baselines. \method{} targets
the same expressivity, a nonlinear transition on a large state, but it obtains a parallel evaluation of it, and its nonlinearity acts
on the features that the memory reads and writes with rather than on the state transition directly.

\paragraph{Parallelizing nonlinear recurrences and its limits.}
DEER \citep{lim2024parallelizing} cast the evaluation of a nonlinear recurrence as a fixed-point problem over the whole trajectory
and solved it with Newton's method, each step a linear recurrence evaluated by a parallel scan; ELK and quasi-DEER
\citep{gonzalez2024elk} stabilize the Newton iteration through its connection to Kalman smoothing and make it scale by using diagonal
Jacobians; ParaRNN \citep{danieli2025pararnn} designs LSTM and GRU cells with structured Jacobians so that the Newton step is cheap
and trains 7B-parameter nonlinear RNNs this way; \citet{gonzalez2025predictability} show that the number of iterations these methods
need is governed by the predictability of the dynamics, measured by the largest Lyapunov exponent, and is polylogarithmic in the
sequence length for contractive systems. Our \cref{thm:convergence} is a statement of the same kind for our iteration: the
contraction condition bounds the learning rate times the memory horizon, which is the quantity that controls how far a token's
update can propagate. A small iteration count improves parallelism, but its expressivity requires a separate analysis. An associative scan generally
has logarithmic circuit depth; a fixed number of scans alone does not establish a $\mathrm{TC}^0$ upper bound.
The bounds of \citet{merrill2023parallelism,merrill2024illusion} apply to the architectures and precision assumptions they analyze.
For transformers, \citet{merrill2025depth} show that logarithmic depth suffices to recognize regular languages. We do not claim a
circuit-complexity classification for \method{}. Our finite-step result instead shows that $N\ge T$ recovers its sequential
recurrence in the undamped case (\cref{prop:fixedpoint}). The practical question, which \cref{sec:exp:convergence} addresses, is how
many iterations are needed for the tasks that matter.

\paragraph{Looped and implicit models.}
A second family iterates a block of layers in depth rather than a recurrence in time. Universal transformers
\citep{dehghani2019universal} apply the same layer repeatedly with adaptive halting; deep equilibrium models \citep{bai2019deep}
take the limit and define the output as the fixed point of a layer, found by a root solver and differentiated implicitly; looped
transformers were shown to be programmable computers \citep{giannou2023looped} and to match the reasoning ability of much deeper
unlooped models \citep{saunshi2025reasoning}, and recent language models scale test-time compute by unrolling a recurrent block to
a chosen depth \citep{geiping2025scaling,zhu2025ouro}. Fixed-point reasoners \citep{movahedi2026fixedpoint} use convergence to a
fixed point as the halting criterion of a looped transformer and add damping to remove oscillations of the iterates, and
\citet{labovich2026stability} \revminor{analyzes} when the fixed points of a looped transformer are reachable, input-dependent and stable
under backpropagation, and \revminor{finds} that recall connections with outer normalization improve stability and generalization in their
studied architectures. Closest to our
work, \citet{schoene2025implicit} iterate a state-space model over the whole sequence until convergence to a fixed point and show
that the implicit model implements the nonlinear state transition of an RNN; like \method{}, they observe that approximate
convergence suffices during training, and they scale the approach to 1.3B parameters. The essential difference is at decoding time.
In recurrent-depth models, the chosen number of loops is part of the token computation; in fixed-point models, the output is
defined by an equilibrium. Both require repeated computation per generated token, including in the sequential mode of
\citet{schoene2025implicit}. For \method{} the loop is only a method for computing a prefill; the object
being computed is the exact sequential recurrence of \cref{eq:rec-steps}, which for one new token is a single gradient step from
the cached weights. Decoding therefore requires no iteration at all (\cref{sec:method:alg}), and $N$ affects the cost and the
accuracy of the prefill only.

\paragraph{\revminor{Parallelizing continuous chain of thought.}}
\revminor{A third family feeds a hidden state of one decoding step back into the input of the next, a nonlinear recurrence in time
whose state is a full residual-stream vector rather than a token. Coconut \citep{hao2025coconut} feeds the last hidden state of
a ``continuous thought'' position back as the next input embedding; because each thought depends on the previous one, $n$
thoughts cost $n+1$ sequential forward passes at training time, which the authors note is the main obstacle to parallelism.
Parallel continuous chain of thought \citep{wu2025pccot} removes the sequential dependence with the same tool we use: starting
from dummy thought inputs, all thought positions are updated in parallel by nonlinear Jacobi iteration, each pass feeding the
previous pass's hidden states forward by one position, and the iteration is exact after as many passes as there are thoughts
(their Theorem~1 is the finite-step exactness of our \cref{prop:fixedpoint}). They train and decode with a few extra passes (three,
for 24 thoughts) and observe that the thought vectors do not in fact converge after training and that training with more than about six
passes becomes unstable. Two recent language models apply the same construction to the whole sequence. T$^2$MLR \citep{cai2026t2mlr} fuses a cache
of a middle-layer representation from the previous token into an earlier layer of the current token, and the full-bandwidth
transformer \citep{wang2026fullbandwidth} fuses the previous token's top-layer state with the next token embedding through a
gated linear unit. Both train by Jacobi iteration over the sequence, re-running the affected layers in parallel with the
previous pass's states shifted right by one position, and both decode sequentially with one pass per token, as \method{} does.
They differ from each other in how the iteration is optimized: the full-bandwidth transformer uses at most three passes with
gradients through all of them, on a schedule in which most batches use a single pass, whereas T$^2$MLR runs sixteen passes but
backpropagates only through the last four, detaching the earlier ones in the manner of truncated backpropagation through time,
and reports that this approximation matches exact recurrent training to within $0.005$ in validation loss at 135M to 1B parameters.
The recurrences in these models run through many layers of a transformer, so each Jacobi pass is a forward pass of the
network and the number of passes is kept small. In \method{} the recurrence is a
gradient step on the weights of one fast-weight layer, the iteration acts on that layer's pre-activation gradients, and each
pass is a chunkwise scan inside one layer's kernel, so the iteration is cheap enough that we backpropagate through
all $N$ passes and the analysis of \cref{sec:method:theory} applies to the converged object.}

\clearpage
\endgroup

\section{Proofs}
\label{app:theory}
\paragraph{Setting, restated.}
We analyze the iteration with $L=2$, the linear-attention hidden update \cref{eq:body-la}, and the uncentered readout.
For this analysis only, replace the unit-ball projection used in the experiments with the soft clip
$n_{\mathrm{soft}}(\bx)=\bx/\sqrt{1+\|\bx\|^2}$. This keeps the features bounded and, under the activation assumptions below, makes their Jacobian globally Lipschitz.
Write $\psi=n_{\mathrm{soft}}\circ\sigma$, where $\sigma$ acts coordinatewise. The features of this smooth variant are
\begin{equation*}
    \bphi_t=\psi(\bz_t),\qquad
    \bz_t=\bW^{(1)}_0\bk_t+\bb^{(1)}+\alpha_t\Delta\bW^{(1)}_{t-1}\bk_t.
\end{equation*}
Step 3 uses the exact Jacobian $\mathbf{H}_t=D\psi(\bz_t)$ at the current pre-activation. Write
$\|\bD\|_\infty=\max_t\|\dz^{(1)}_t\|$ and $F$ for one iteration's map $\bD\mapsto\hat\bD$.

\begin{theorem*}[Theorem~\ref{thm:convergence}, restated]
For the soft-clipped variant defined above, assume $\|\bk_t\|\le1$, $\|\bv_t\|\le V$, and $\|\bW^{(2)}_0\|_{\mathrm{op}}\le w_0$; shared decay
$0\le\alpha_t\le\bar\alpha<1$ and shared learning rates $0<\eta^{(1)}_t=\eta^{(2)}_t=\eta_t\le\bar\eta\le1$; and
$\sigma\in C^2(\R)$ with $|\sigma'(x)|\le c_\sigma$ and $|\sigma''(x)|\le h_\sigma$ for every $x\in\R$.
Set $\epsilon=\bar\eta/(1-\bar\alpha)$ and $Q=V+w_0$. Then $F$ is globally Lipschitz in $\|\cdot\|_\infty$ with constant
\begin{equation*}
    \rho=\epsilon(1+\epsilon)Q^2
    \big[c_\sigma^2(1+\epsilon)(5+4\epsilon)+h_\sigma(2+\epsilon)\big].
\end{equation*}
For $\epsilon\le\tfrac14$, $\rho\le(10c_\sigma^2+3h_\sigma)\epsilon Q^2$.
If $\rho<1$, $F$ has a unique fixed point $\bD^\star$, the gradient sequence of the sequential recurrence \cref{eq:rec-steps} with $n_{\mathrm{soft}}$ in place of the projection.
For any initialization and damping $\lambda\in(0,1]$, \cref{eq:picard} converges geometrically:
\[
    \|\bD^{[n]}-\bD^\star\|_\infty\le\big(1-\lambda(1-\rho)\big)^n\|\bD^{[0]}-\bD^\star\|_\infty.
\]
The features, readout states and outputs converge to those of the sequential recurrence at the same rate, up to constant factors.
\end{theorem*}

\paragraph{Proof strategy.}
We compare two guesses of the hidden-gradient sequence and follow their differences through one iteration.
First, the hidden scan turns the guesses into pre-activations and features. Next, these features determine the readout trajectory
and its feature gradients. Finally, the backward pass multiplies each feature gradient by the activation Jacobian.
Both factors in this last product change between the two guesses, so we must control both changes.
The lemmas below give these bounds in the order they are needed.

\paragraph{Notation.}
We use the Euclidean norm for vectors and the operator norm for matrices. For a sequence, $\|\cdot\|_\infty$ is the maximum
of the vector norms over tokens. There is one hidden layer, so we omit its layer index and write $\bD=(\dz_t)_{t\le T}$.
A prime denotes the quantity obtained from a second guess $\bD'$, with the keys, values and model parameters held fixed.

Past writes are multiplied by the decay factors
\[
    \gamma_{j,t}=\prod_{i=j+1}^{t-1}\alpha_i.
\]
The assumption $\alpha_i\le\bar\alpha<1$ bounds their total contribution, independently of sequence length:
\[
    \bar\eta\sum_{j<t}\gamma_{j,t}
    \le\bar\eta\sum_{r\ge0}\bar\alpha^r
    =\frac{\bar\eta}{1-\bar\alpha}=\epsilon.
\]
We will use this geometric-series bound for both the hidden and readout states.
For brevity, write $n=n_{\mathrm{soft}}$ throughout the proof.

\begin{lemma}[Soft clip]
\label{lem:softclip}
For $n(\bx)=\bx/\sqrt{1+\|\bx\|^2}$, the output and Jacobian satisfy
\[
    \|n(\bx)\|\le1,\qquad \|Dn(\bx)\|\le1,\qquad
    \|Dn(\bx)-Dn(\by)\|\le2\|\bx-\by\|.
\]
\end{lemma}
\begin{proof}
The output bound follows directly from the definition. To bound the derivatives, set
$s=\sqrt{1+\|\bx\|^2}$ and $\ba=n(\bx)$. Differentiation gives
\[
    Dn(\bx)=s^{-1}\bI-s^{-3}\bx\bx^\top
           =s^{-1}(\bI-\ba\ba^\top).
\]
Along $\bx$, this matrix has eigenvalue $s^{-3}$; in directions perpendicular to $\bx$, it has eigenvalue $s^{-1}$.
At $\bx=0$ it is the identity. Thus $\|Dn(\bx)\|\le s^{-1}\le1$, which also shows that $n$ is nonexpansive.

It remains to bound the change in the Jacobian. In a direction $\bh$, the derivative of the scalar factor $s^{-1}$ has magnitude
at most $\|\bx\|\|\bh\|/s^3$. The derivative of $\ba\ba^\top$ has norm at most
$2\|\ba\|\|Dn(\bx)\bh\|$. Applying the product rule therefore gives
\[
    \|D(Dn)(\bx)[\bh]\|
    \le\frac{3\|\bx\|}{(1+\|\bx\|^2)^{3/2}}\|\bh\|
    \le\frac{2}{\sqrt3}\|\bh\|\le2\|\bh\|.
\]
Here we used $\max_{r\ge0}r/(1+r^2)^{3/2}=2/(3\sqrt3)$.
Integrating this bound along the segment from $\bx$ to $\by$ proves the final inequality.
\end{proof}

\begin{lemma}[Composite activation]
\label{lem:activation}
The full activation $\psi=n\circ\sigma$ satisfies
\[
    \|D\psi(\bz)\|\le c_\sigma,\qquad
    \|D\psi(\bz)-D\psi(\bz')\|\le(2c_\sigma^2+h_\sigma)\|\bz-\bz'\|.
\]
\end{lemma}
\begin{proof}
The chain rule factors the Jacobian into the derivatives of the soft clip and the coordinatewise activation:
\[
    D\psi(\bz)=\bN(\bz)\bJ(\bz),\qquad
    \bN(\bz)=Dn(\sigma(\bz)),\qquad
    \bJ(\bz)=\operatorname{diag}\sigma'(\bz).
\]
Since $\|\bN\|\le1$ and $\|\bJ\|\le c_\sigma$, their product has norm at most $c_\sigma$.
For its variation, the activation assumptions and \cref{lem:softclip} give
\begin{align*}
    \|\bN(\bz)-\bN(\bz')\|&\le2\|\sigma(\bz)-\sigma(\bz')\|
        \le2c_\sigma\|\bz-\bz'\|,\\
    \|\bJ(\bz)-\bJ(\bz')\|&\le h_\sigma\|\bz-\bz'\|.
\end{align*}
Now split the difference of the products as
$\bN\bJ-\bN'\bJ'=(\bN-\bN')\bJ+\bN'(\bJ-\bJ')$.
The first term contributes at most $2c_\sigma^2\|\bz-\bz'\|$, and the second contributes at most
$h_\sigma\|\bz-\bz'\|$. Adding them proves the claim.
\end{proof}

\begin{lemma}[Features]
\label{lem:features}
For any two gradient guesses, let $\delta=\|\bD-\bD'\|_\infty$. Then
\[
    \|\bphi_t\|\le1,\qquad
    \|\bz_t-\bz'_t\|\le\epsilon\delta,\qquad
    \|\bphi_t-\bphi'_t\|\le c_\sigma\epsilon\delta.
\]
\end{lemma}
\begin{proof}
Only the fast weights depend on the gradient guess. Their contribution to the pre-activation is, by \cref{eq:body-la},
\[
    \bu_t=\alpha_t\Delta\bW^{(1)}_{t-1}\bk_t
    =-\alpha_t\sum_{j<t}\gamma_{j,t}\eta_j(\bk_j^\top\bk_t)\dz_j.
\]
The normalized keys satisfy $|\bk_j^\top\bk_t|\le1$. Subtracting the two sums and using $\alpha_t\le1$ gives
\[
    \|\bz_t-\bz'_t\|=\|\bu_t-\bu'_t\|
    \le\bar\eta\sum_{j<t}\gamma_{j,t}\|\dz_j-\dz'_j\|
    \le\epsilon\delta.
\]
Applying the $c_\sigma$-Lipschitz map $\psi$ gives the feature-difference bound.
The bound $\|\bphi_t\|\le1$ follows from the soft clip.
\end{proof}

\begin{lemma}[Readout trajectory]
\label{lem:readout}
Let $\bS_t=\Delta\bW^{(2)}_t$ be the readout's fast weights for features satisfying $\|\bphi_t\|\le1$. Then
\[
    \|\bS_t\|\le\epsilon Q,\qquad
    \|\bar\bW^{(2)}_{t-1}\|\le(1+\epsilon)Q.
\]
For two such feature sequences, writing $\delta_\phi=\|\bphi-\bphi'\|_\infty$, we also have
\[
    \|\bS_t-\bS'_t\|\le2\epsilon(1+\epsilon)Q\,\delta_\phi.
\]
\end{lemma}
\begin{proof}
Separate each readout update into the part carried from the previous state and the new write:
\begin{equation}
    \bS_t=\alpha_t\bS_{t-1}(\bI-\eta_t\bphi_t\bphi_t^\top)
           +\eta_t(\bv_t-\bW^{(2)}_0\bphi_t)\bphi_t^\top,
    \qquad \bS_0=\mathbf0.
    \label{eq:S-rec}
\end{equation}
We first bound the state itself, then compare the states generated by different features.

\emph{State size.}
Because $\eta_t\|\bphi_t\|^2\le1$, the matrix $\bI-\eta_t\bphi_t\bphi_t^\top$ has operator norm at most one.
The new write has norm at most $\bar\eta(V+w_0)=\bar\eta Q$. Hence
\[
    \|\bS_t\|\le\bar\alpha\|\bS_{t-1}\|+\bar\eta Q.
\]
Starting from zero and summing the geometric series gives $\|\bS_t\|\le\epsilon Q$.
Adding the slow weights then gives
\[
    \|\bar\bW^{(2)}_{t-1}\|
    =\|\bW^{(2)}_0+\alpha_t\bS_{t-1}\|
    \le w_0+\epsilon Q\le(1+\epsilon)Q.
\]

\emph{Sensitivity to the features.}
Write $\delta_t=\|\bS_t-\bS'_t\|$. Two elementary bounds control the feature-dependent terms in \cref{eq:S-rec}:
\begin{align*}
    \|\bphi_t\bphi_t^\top-\bphi'_t\bphi_t'^\top\|&\le2\delta_\phi,\\
    \|(\bv_t-\bW^{(2)}_0\bphi_t)\bphi_t^\top
       -(\bv_t-\bW^{(2)}_0\bphi'_t)\bphi_t'^\top\|&\le(V+2w_0)\delta_\phi.
\end{align*}
The first follows by changing one factor of the outer product at a time. For the second, the value term contributes
$V\delta_\phi$ and the quadratic feature term contributes $2w_0\delta_\phi$.
Subtracting the two recurrences now gives
\begin{align*}
    \delta_t
    &\le\bar\alpha\delta_{t-1}
      +2\bar\eta\bar\alpha\|\bS'_{t-1}\|\delta_\phi
      +\bar\eta(V+2w_0)\delta_\phi\\
    &\le\bar\alpha\delta_{t-1}+2\bar\eta(1+\epsilon)Q\,\delta_\phi.
\end{align*}
In the last line we used the state-size bound and $V+2w_0\le2Q$.
Since $\delta_0=0$, another geometric sum yields
$\delta_t\le2\epsilon(1+\epsilon)Q\,\delta_\phi$.
\end{proof}

\begin{lemma}[Feature gradient]
\label{lem:gradient}
For features in the unit ball, let
\[
    \bg_t=\bar\bW^{(2)\top}_{t-1}
           (\bar\bW^{(2)}_{t-1}\bphi_t-\bv_t).
\]
Its size and sensitivity to the feature sequence satisfy
\[
    \|\bg_t\|\le B,\qquad
    \|\bg_t-\bg'_t\|\le A\|\bphi-\bphi'\|_\infty,
\]
where
\[
    B=(1+\epsilon)(2+\epsilon)Q^2,\qquad
    A=(1+\epsilon)(1+7\epsilon+4\epsilon^2)Q^2.
\]
\end{lemma}
\begin{proof}
Fix a token and abbreviate its decayed readout by $\bar\bW$ and its residual by $\br=\bar\bW\bphi_t-\bv_t$.
Set $M=(1+\epsilon)Q$. By \cref{lem:readout},
\[
    \|\bar\bW\|\le M,\qquad \|\br\|\le M+V.
\]
Therefore $\|\bg_t\|\le M(M+V)\le B$, using $V\le Q$.

For the difference bound, there are two sources of change: the readout weights and the features at this token.
Writing $\Delta_W=\|\bar\bW-\bar\bW'\|$ and $\delta_\phi=\|\bphi-\bphi'\|_\infty$, we have
\[
    \br-\br'=(\bar\bW-\bar\bW')\bphi_t
                   +\bar\bW'(\bphi_t-\bphi'_t),
    \qquad
    \|\br-\br'\|\le\Delta_W+M\delta_\phi.
\]
Splitting the gradient difference in the same way gives
\begin{align*}
    \|\bg_t-\bg'_t\|
    &\le\Delta_W\|\br\|+\|\bar\bW'\|\,\|\br-\br'\|\\
    &\le(2M+V)\Delta_W+M^2\delta_\phi.
\end{align*}
The readout lemma supplies $\Delta_W\le2\epsilon(1+\epsilon)Q\delta_\phi$.
Substituting this and $2M+V\le(3+2\epsilon)Q$ yields
\[
    \|\bg_t-\bg'_t\|
    \le\big[2\epsilon(1+\epsilon)(3+2\epsilon)+(1+\epsilon)^2\big]Q^2\delta_\phi
    =A\delta_\phi.
\]
\end{proof}

\begin{proof}[Proof of \cref{thm:convergence}]
\emph{Bounding one iteration.}
Let $\delta=\|\bD-\bD'\|_\infty$. The backward pass returns
$F_t(\bD)=\mathbf{H}_t^\top\bg_t$, where $\mathbf{H}_t=D\psi(\bz_t)$.
The preceding lemmas give the four bounds needed to compare these products:
\begin{align*}
    \|\mathbf{H}_t\|&\le c_\sigma,&
    \|\mathbf{H}_t-\mathbf{H}'_t\|&\le(2c_\sigma^2+h_\sigma)\epsilon\delta,\\
    \|\bg'_t\|&\le B,&
    \|\bg_t-\bg'_t\|&\le A c_\sigma\epsilon\delta.
\end{align*}
Here $A$ bounds the sensitivity of the feature gradient and $B$ bounds its size, as defined in \cref{lem:gradient}.
The change in the gradient contributes the first term below; the change in the Jacobian contributes the second:
\begin{align*}
    \|F_t(\bD)-F_t(\bD')\|
    &\le\|\mathbf{H}_t\|\,\|\bg_t-\bg'_t\|
         +\|\mathbf{H}_t-\mathbf{H}'_t\|\,\|\bg'_t\|\\
    &\le\epsilon\big[c_\sigma^2 A+(2c_\sigma^2+h_\sigma)B\big]\delta.
\end{align*}
Substituting $A$ and $B$ and collecting terms gives exactly the constant $\rho$ in \cref{eq:rho}.
Taking the maximum over tokens proves $\|F(\bD)-F(\bD')\|_\infty\le\rho\delta$.
The simpler bound for $\epsilon\le1/4$ follows from
\[
    (1+\epsilon)^2(5+4\epsilon)\le\frac{75}{8}<10,
    \qquad (1+\epsilon)(2+\epsilon)\le\frac{45}{16}<3.
\]

\emph{Identifying the limit.}
If $\rho<1$, the map is a contraction on the complete space $(\R^{T\times \revminor{d}},\|\cdot\|_\infty)$.
Banach's theorem therefore gives a unique fixed point $\bD^\star$.
At that point, every guessed gradient equals the actual gradient returned by the backward pass.
This identifies the resulting trajectory with the sequential recurrence using the soft clip: both start from zero fast weights;
if their states agree before a token, their features and gradients agree at that token, so their updates give the same next states.
Induction over tokens proves agreement of the whole trajectory.

\emph{Damping and the convergence rate.}
The damped map is $G(\bD)=(1-\lambda)\bD+\lambda F(\bD)$.
It has the same fixed point and satisfies
\[
    \|G(\bD)-G(\bD')\|_\infty
    \le(1-\lambda+\lambda\rho)\|\bD-\bD'\|_\infty.
\]
Applying this inequality repeatedly with $\bD'=\bD^\star$ gives the stated geometric rate.

\emph{Features, states and outputs.}
The feature and readout lemmas show that these quantities change by at most a constant times the error in $\bD$.
The same is true for query-side features: their unshifted hidden scan has gain at most
$\epsilon\max_t\|\bq_t\|$, followed by the $c_\sigma$-Lipschitz map $\psi$.
Finally, the output is $\bo_t=\bW^{(2)}_t\bphi_t^{\bq}$, a product of bounded quantities that are Lipschitz in $\bD$.
Thus their errors all inherit the same geometric rate, up to constant factors.
\end{proof}

\paragraph{SiLU satisfies the assumptions.}
For $\sigma(x)=x s(x)$ with $s(x)=(1+e^{-x})^{-1}$,
$\sigma'(x)=s(x)+x s'(x)$ and $\sigma''(x)=2s'(x)+x s'(x)(1-2s(x))$.
Since $s'(x)\le\min(1/4,e^{-|x|})$ and $|x|e^{-|x|}\le1/e$, valid global bounds are
$c_\sigma=1+1/e$ and $h_\sigma=1/2+1/e$.

\paragraph{Fixed point and finite-step exactness.}
\label{app:finite-exactness}
This is the formal version of \cref{prop:fixedpoint}.
\begin{proposition}[Finite-step exactness of the undamped iteration]
\label{prop:finite-exactness}
For the iteration of \cref{sec:method:alg} with $\lambda=1$ and any initialization, the estimated hidden gradients at tokens
$1,\ldots,\min(n,T)$ equal their sequential values after $n$ iterations. Consequently, for $N\ge T$, the final evaluation reproduces
the sequential recurrence exactly in exact arithmetic. The argument applies at any memory depth, with either projection or soft
clipping, using the same derivative convention in the parallel and sequential evaluations, and requires no contraction condition. For
$0<\lambda<1$, finite-step exactness fails in general: already at token $1$, $\bD^{[n]}_1-\bD^\star_1=(1-\lambda)^n(\bD^{[0]}_1-\bD^\star_1)$.
\end{proposition}
\begin{proof}
Group all hidden-layer gradient estimates at token $t$ into $\bD_t$.
The iteration is strictly causal: $F_t(\bD)$ depends only on $\bD_1,\ldots,\bD_{t-1}$.
To see this at arbitrary depth, proceed layer by layer in the forward pass. At token $s\le t$, each hidden scan uses only writes
$j<s$, and its layer inputs depend only on estimates at tokens preceding their own positions. Thus all key-side features through
token $t$ depend only on $\bD_{<t}$. The readout before token $t$ is determined by earlier features and updates. The backward pass
uses these same hidden and readout states, so it introduces no dependence on $\bD_t$ or later estimates.

The sequential gradient sequence $\bD^\star$ satisfies $F(\bD^\star)=\bD^\star$: at $\bD^\star$ the shifted scans of Step 1 sum the
writes of tokens $j<t$ and reproduce the decayed weights $\bar\bW^{(l)}_{t-1}$ of \cref{eq:rec-steps} at every token, Step 2 reproduces
$\bar\bW^{(L)}_{t-1}$, and Step 3 is the backpropagation of the sequential model, so $\dzh=\dz$; the parallel evaluation at $\bD^\star$
therefore coincides with the sequential recurrence, and damping changes the iterates but not this fixed point.
Since $F_1$ has no dependence on $\bD$, the first undamped iteration gives $\bD^{[1]}_1=\bD^\star_1$.
Suppose the first $n$ tokens are exact after $n$ iterations. For every $t\le\min(n+1,T)$, all arguments of $F_t$ then agree with
those of $F_t(\bD^\star)$, hence $\bD^{[n+1]}_t=\bD^\star_t$. Induction proves the prefix claim and gives
$\bD^{[N]}=\bD^\star$ whenever $N\ge T$. The final forward and readout passes therefore reproduce the sequential states and outputs.
The argument uses only causality and identical local derivative evaluations, so it applies to either clipping choice; at a
nondifferentiable point of the hard projection, both evaluations must use the same fixed derivative convention.
\end{proof}

\begin{revblocks}
\paragraph{Convergence with the hidden-layer delta rule.}
\label{app:theory-delta}
\Cref{thm:convergence} analyzes the linear-attention hidden update. We now prove the analogous statement for the hidden-layer delta
rule of \cref{sec:method:practical} (\cref{app:body-delta}), which the reported models use, and then extend both theorems to the
centered readout (\cref{cor:centering}). The setting is otherwise that of
\cref{thm:convergence}: $L=2$, the uncentered readout and the soft clip. The notation and
\cref{lem:softclip,lem:activation,lem:readout,lem:gradient} are reused unchanged; only the feature bound of \cref{lem:features}
is replaced.

With one hidden layer, the keys of the hidden delta rule are the layer inputs $\bk_t$, which do not depend on the iterate. Write
$\kappa_t=\|\bk_t\|$ and $\hat\bk_t=\bk_t/\kappa_t$, and take the write rate $\beta_t=\eta_t$ as in \cref{app:body-delta}.
Substituting the target $\by_t$ into \cref{eq:body-delta}, a forward pass with gradient guess $\bD=(\dz_t)_{t\le T}$ and previous
reads $\tilde{\mathbf R}=(\tilde\br_t)_{t\le T}$ evolves the hidden fast weights from $\Delta\bW^{(1)}_0=\mathbf 0$ by
\begin{equation}
    \Delta\bW^{(1)}_t
    =\alpha_t\Delta\bW^{(1)}_{t-1}\big(\bI-\eta_t\hat\bk_t\hat\bk_t^\top\big)
     +\eta_t\big(\tilde\br_t-\kappa_t\dz_t\big)\hat\bk_t^\top,
    \qquad
    \br_t=\alpha_t\Delta\bW^{(1)}_{t-1}\hat\bk_t ,
    \label{eq:body-delta-rec}
\end{equation}
and the pre-activations are $\bz_t=\bW^{(1)}_0\bk_t+\bb^{(1)}+\kappa_t\br_t$. Steps~2 and~3 do not involve the hidden update rule,
so the backward pass again returns $\hat\bD_t=\mathbf{H}_t^\top\bg_t$ with $\mathbf{H}_t=D\psi(\bz_t)$ and $\bg_t$ as in
\cref{lem:gradient}. The iteration now carries the reads as well as the gradients. Write $\Psi$ for the map of one iteration,
$\Psi(\bD,\tilde{\mathbf R})=(\hat\bD,\mathbf R)$ with $\mathbf R=(\br_t)_{t\le T}$. As in the algorithm, the gradients are damped
by \cref{eq:picard} and the reads are replaced:
\begin{equation}
    \bD^{[n]}=\bD^{[n-1]}+\lambda\big(\hat\bD-\bD^{[n-1]}\big),\qquad
    \mathbf R^{[n]}=\mathbf R,
    \qquad\text{where }(\hat\bD,\mathbf R)=\Psi\big(\bD^{[n-1]},\mathbf R^{[n-1]}\big),
    \label{eq:picard-delta}
\end{equation}
and the algorithm starts from $\bD^{[0]}=\mathbf 0$, $\mathbf R^{[0]}=\mathbf 0$. For $\lambda\in(0,1]$ we measure pairs in the norm
$\|(\bD,\mathbf R)\|_\lambda=\|\bD\|_\infty+\lambda\|\mathbf R\|_\infty$.

\begin{theorem}[Convergence of \method{} with the hidden-layer delta rule, $L=2$]
\label{thm:convergence-delta}
Consider the soft-clipped variant with the hidden-layer delta rule \cref{eq:body-delta-rec}. Let the assumptions of
\cref{thm:convergence} hold, with nonzero keys, $0<\|\bk_t\|\le1$, and let $\epsilon$, $Q$ and $\rho$ be as defined there. Then:
\begin{enumerate}
    \item For any two pairs $(\bD,\tilde{\mathbf R})$ and $(\bD',\tilde{\mathbf R}')$, with
    $\delta_D=\|\bD-\bD'\|_\infty$ and $\delta_R=\|\tilde{\mathbf R}-\tilde{\mathbf R}'\|_\infty$, the outputs of one iteration satisfy
    \[
        \|\hat\bD-\hat\bD'\|_\infty\le\rho\,(\delta_D+\delta_R),\qquad
        \|\mathbf R-\mathbf R'\|_\infty\le\epsilon\,(\delta_D+\delta_R).
    \]
    In particular, $\Psi$ is globally Lipschitz with constant $\rho+\epsilon$ in the norm $\|\bD\|_\infty+\|\mathbf R\|_\infty$.
    \item If $\rho+\epsilon<1$, $\Psi$ has a unique fixed point $(\bD^\star,\mathbf R^\star)$: $\bD^\star$ is the gradient sequence of the
    sequential recurrence \cref{eq:rec-steps} with $n_{\mathrm{soft}}$ in place of the projection, the same as in
    \cref{thm:convergence}, and $\br^\star_t=\alpha_t\Delta\bW^{(1)}_{t-1}\hat\bk_t$ are the key reads of that recurrence. For any
    initialization and damping $\lambda\in(0,1]$, the iteration \cref{eq:picard-delta} converges geometrically:
    \[
        \big\|(\bD^{[n]},\mathbf R^{[n]})-(\bD^\star,\mathbf R^\star)\big\|_\lambda
        \le\big(1-\lambda(1-\rho-\epsilon)\big)^n\,
        \big\|(\bD^{[0]},\mathbf R^{[0]})-(\bD^\star,\mathbf R^\star)\big\|_\lambda .
    \]
    The features, readout states and outputs converge to those of the sequential recurrence at the same rate, up to constant factors.
\end{enumerate}
\end{theorem}

The statement has the same form as \cref{thm:convergence}, with $\rho+\epsilon$ in place of $\rho$: the additional $\epsilon$ is the
gain of the reads that the delta rule carries from one iteration to the next. As before, the condition is sufficient, does not depend on the
sequence length, and does not cover the hard projection; centering is treated in \cref{cor:centering}.

\begin{lemma}[Hidden delta-rule trajectory]
\label{lem:delta-reads}
For two pairs $(\bD,\tilde{\mathbf R})$ and $(\bD',\tilde{\mathbf R}')$, with $\delta_D$ and $\delta_R$ as in
\cref{thm:convergence-delta}, the trajectories of \cref{eq:body-delta-rec} satisfy, for every token $t$,
\[
    \|\Delta\bW^{(1)}_t-\Delta\bW'^{(1)}_t\|\le\epsilon(\delta_D+\delta_R),\qquad
    \|\br_t-\br'_t\|\le\epsilon(\delta_D+\delta_R),\qquad
    \|\bz_t-\bz'_t\|\le\epsilon(\delta_D+\delta_R),
\]
and therefore $\|\bphi_t-\bphi'_t\|\le c_\sigma\epsilon(\delta_D+\delta_R)$, while $\|\bphi_t\|\le1$.
\end{lemma}
\begin{proof}
Let $\mathbf{E}_t=\Delta\bW^{(1)}_t-\Delta\bW'^{(1)}_t$. The keys, gates and learning rates are the same for both pairs, so
subtracting the two copies of \cref{eq:body-delta-rec} gives a linear recurrence for the difference,
\[
    \mathbf{E}_t
    =\alpha_t\mathbf{E}_{t-1}\big(\bI-\eta_t\hat\bk_t\hat\bk_t^\top\big)
     +\eta_t\big[(\tilde\br_t-\tilde\br'_t)-\kappa_t(\dz_t-\dz'_t)\big]\hat\bk_t^\top,
    \qquad \mathbf{E}_0=\mathbf 0 .
\]
Because $\hat\bk_t$ is a unit vector and $0<\eta_t\le1$, the matrix $\bI-\eta_t\hat\bk_t\hat\bk_t^\top$ has eigenvalues $1-\eta_t$
and $1$, hence operator norm at most one. The second term is an outer product with a unit vector, so its norm is at most
$\bar\eta(\delta_R+\kappa_t\delta_D)\le\bar\eta(\delta_D+\delta_R)$, using $\kappa_t\le1$. Therefore
\[
    \|\mathbf{E}_t\|\le\bar\alpha\|\mathbf{E}_{t-1}\|+\bar\eta(\delta_D+\delta_R),
\]
and summing the geometric series from $\mathbf{E}_0=\mathbf 0$ gives $\|\mathbf{E}_t\|\le\epsilon(\delta_D+\delta_R)$ for every $t$.
The reads and pre-activations differ by $\br_t-\br'_t=\alpha_t\mathbf{E}_{t-1}\hat\bk_t$ and
$\bz_t-\bz'_t=\kappa_t(\br_t-\br'_t)$, and $\alpha_t,\kappa_t\le1$, which gives the next two bounds. The feature bound follows
because $\psi$ is $c_\sigma$-Lipschitz (\cref{lem:activation}), and $\|\bphi_t\|\le1$ by the soft clip.
\end{proof}

\begin{proof}[Proof of \cref{thm:convergence-delta}]
\emph{Bounding one iteration.}
The bound on the reads is part of \cref{lem:delta-reads}. For the gradients, \cref{lem:delta-reads} plays the role of
\cref{lem:features}, with $\delta_D+\delta_R$ in place of $\delta$; the rest of the argument is that of \cref{thm:convergence}.
By \cref{lem:activation} and \cref{lem:delta-reads},
$\|\mathbf{H}_t\|\le c_\sigma$ and $\|\mathbf{H}_t-\mathbf{H}'_t\|\le(2c_\sigma^2+h_\sigma)\epsilon(\delta_D+\delta_R)$.
\Cref{lem:readout,lem:gradient} hold for any feature sequences in the unit ball, so $\|\bg'_t\|\le B$ and
$\|\bg_t-\bg'_t\|\le A\|\bphi-\bphi'\|_\infty\le Ac_\sigma\epsilon(\delta_D+\delta_R)$. Hence
\begin{align*}
    \|\hat\bD_t-\hat\bD'_t\|
    &\le\|\mathbf{H}_t\|\,\|\bg_t-\bg'_t\|+\|\mathbf{H}_t-\mathbf{H}'_t\|\,\|\bg'_t\|\\
    &\le\epsilon\big[c_\sigma^2A+(2c_\sigma^2+h_\sigma)B\big](\delta_D+\delta_R)
    =\rho\,(\delta_D+\delta_R),
\end{align*}
where the last equality is the computation in the proof of \cref{thm:convergence}. Taking the maximum over tokens and adding the
two bounds proves the first part.

\emph{Damping and the convergence rate.}
Fix $\lambda\in(0,1]$ and let $G_\lambda(\bD,\tilde{\mathbf R})=\big((1-\lambda)\bD+\lambda\hat\bD,\ \mathbf R\big)$ be one step of
\cref{eq:picard-delta}. By the first part,
\begin{align*}
    \big\|G_\lambda(\bD,\tilde{\mathbf R})-G_\lambda(\bD',\tilde{\mathbf R}')\big\|_\lambda
    &\le(1-\lambda)\delta_D+\lambda\rho(\delta_D+\delta_R)+\lambda\epsilon(\delta_D+\delta_R)\\
    &=\theta\,\delta_D+\lambda(\rho+\epsilon)\,\delta_R ,
    \qquad \theta=1-\lambda(1-\rho-\epsilon).
\end{align*}
Since $\theta-(\rho+\epsilon)=(1-\lambda)(1-\rho-\epsilon)\ge0$ when $\rho+\epsilon\le1$, the right-hand side is at most
$\theta(\delta_D+\lambda\delta_R)=\theta\,\|(\bD,\tilde{\mathbf R})-(\bD',\tilde{\mathbf R}')\|_\lambda$. If $\rho+\epsilon<1$, then
$\theta<1$ and $G_\lambda$ is a contraction on the complete space
$\big(\R^{T\times d}\times\R^{T\times d},\|\cdot\|_\lambda\big)$. Banach's theorem gives a unique fixed point, and applying the
inequality repeatedly with the second pair equal to the fixed point gives the stated rate. A pair is fixed by $G_\lambda$ exactly
when $\mathbf R=\tilde{\mathbf R}$ and $\hat\bD=\bD$, that is, when it is fixed by $\Psi$, so the fixed point does not depend on
$\lambda$.

\emph{Identifying the limit.}
At a fixed point, $\tilde\br_t=\br_t=\alpha_t\Delta\bW^{(1)}_{t-1}\hat\bk_t$ for every $t$. Substituting this into
\cref{eq:body-delta-rec}, the two terms containing the read cancel and, since $\kappa_t\hat\bk_t=\bk_t$,
\[
    \Delta\bW^{(1)}_t=\alpha_t\Delta\bW^{(1)}_{t-1}-\eta_t\,\dz_t\,\bk_t^\top ,
\]
which is the hidden-layer recurrence \cref{eq:rec-W}. Moreover $\hat\bD=\bD$, so every guessed gradient equals the gradient returned
by the backward pass on this trajectory. The induction over tokens in the proof of \cref{thm:convergence} then identifies the
trajectory with the sequential recurrence using the soft clip. Hence $\bD=\bD^\star$, and the fixed reads are the key reads
$\br^\star_t$ of that recurrence.

\emph{Features, states and outputs.}
By \cref{lem:delta-reads}, the hidden fast weights, pre-activations and features of the final forward pass differ from those of
the sequential recurrence by at most a constant times
$\|\bD^{[N]}-\bD^\star\|_\infty+\|\mathbf R^{[N]}-\mathbf R^\star\|_\infty
\le\lambda^{-1}\|(\bD^{[N]},\mathbf R^{[N]})-(\bD^\star,\mathbf R^\star)\|_\lambda$,
and \cref{lem:readout} transfers this bound to the readout states. The query-side features read the updated weights
$\Delta\bW^{(1)}_t$, whose error is bounded by the same lemma, so their pre-activations differ by at most
$\epsilon\max_t\|\bq_t\|$ times the same quantity. The output is again a product of bounded quantities that are Lipschitz in
$(\bD,\mathbf R)$, so all of them inherit the geometric rate, up to constant factors.
\end{proof}

\begin{remark}[Finite-step exactness with the delta rule]
The map $\Psi$ is strictly causal: by \cref{eq:body-delta-rec}, $\br_t$ and $\bz_t$ depend only on $\dz_j$ and $\tilde\br_j$ with
$j<t$, and so does $\hat\bD_t$. The induction in the proof of \cref{prop:finite-exactness} therefore applies to the pair
$(\bD,\mathbf R)$: without damping, the gradients and reads at tokens $1,\dots,\min(n,T)$ are exact after $n$ iterations, with no
contraction condition.
\end{remark}

\paragraph{Centering.}
The centered readout of \cref{sec:method:practical} changes only the values that the readout fits, by a term that does not depend
on the iterate, so both theorems carry over.

\begin{corollary}[Centered readout]
\label{cor:centering}
\Cref{thm:convergence,thm:convergence-delta} hold for the centered readout \cref{eq:center} with $Q$ replaced by
$Q_{\mathrm c}=V+2w_0$ in the constant $\rho$. The fixed point is then the gradient sequence, and for the delta rule the key reads,
of the sequential recurrence with the centered readout and $n_{\mathrm{soft}}$ in place of the projection.
\end{corollary}
\begin{proof}
Let $\bphi_{0,t}=\psi(\bW^{(1)}_0\bk_t+\bb^{(1)})$ be the features of the slow MLP. They are determined by the keys and the slow
parameters, so they do not depend on the gradient guess or the reads, and $\|\bphi_{0,t}\|\le1$ by the soft clip. By
\cref{eq:center}, the centered inner loss of token $t$ is
\[
    \ell_t=\tfrac12\big\|\bar\bW^{(2)}_{t-1}\bphi_t-\bW^{(2)}_0\bphi_{0,t}-\bv_t\big\|^2
          =\tfrac12\big\|\bar\bW^{(2)}_{t-1}\bphi_t-\bv^{\mathrm c}_t\big\|^2,
    \qquad \bv^{\mathrm c}_t=\bv_t+\bW^{(2)}_0\bphi_{0,t},
\]
which is the uncentered loss with the values $\bv_t$ replaced by the fixed sequence $\bv^{\mathrm c}_t$. The centered sequential
recurrence is therefore the uncentered recurrence \cref{eq:rec-steps} run on the values $\bv^{\mathrm c}_t$. The same holds for
every step of the iteration: the corrected values of \cref{app:centering} are
$\bv_t-\bW^{(2)}_0(\bphi_t-\bphi_{0,t})=\bv^{\mathrm c}_t-\bW^{(2)}_0\bphi_t$, those of the uncentered readout on
$\bv^{\mathrm c}_t$, so the readout trajectory, its feature gradient and the backward pass coincide with their uncentered
counterparts, with either hidden update. The centered output differs from the uncentered output on $\bv^{\mathrm c}_t$ only by
the term $-\bW^{(2)}_0\bphi_{0,\bq,t}$, which does not depend on the iterate. The values enter the proofs of
\cref{thm:convergence,thm:convergence-delta} only through the bound $\|\bv_t\|\le V$. Since
$\|\bv^{\mathrm c}_t\|\le V+w_0$, both theorems apply with $V+w_0$ in place of $V$, that is, with $Q$ replaced by
$V+2w_0$.
\end{proof}
\end{revblocks}

\section{Algorithm details}
\label{app:alg}

\begin{algorithm}[h]
\caption{\method{} \revminor{with linear-attention hidden updates and an uncentered readout} (one head; all operations act on the full sequence of $T$ tokens). \revminor{Practical modifications are in \cref{sec:method:practical}.}}
\label{alg:main}
\begin{algorithmic}[1]
\Require keys $\bk_{1:T}$, queries $\bq_{1:T}$, values $\bv_{1:T}$; gates $\alpha_{1:T}$, learning rates $\eta^{(1..L)}_{1:T}$; weights $\bW^{(l)}_0,\bb^{(l)}$; iterations $N$; damping $\lambda$
\State $\dz^{(l)}_{1:T}\gets\mathbf 0$ for every hidden layer $l<L$ \Comment{initial estimate of the backward gradients}
\For{$n=1,\dots,N$}
    \Statex \hspace{\algorithmicindent}\stepfwd{Step 1: parallel forward pass}
    \State $\bh^{(0)}\gets\bk$
    \For{$l=1,\dots,L-1$}
        \State $\ba^{(l)}\gets\bW^{(l)}_0\bh^{(l-1)}+\bb^{(l)}+\textsc{ShiftedScan}\big(\text{keys }\bh^{(l-1)},\ \text{values }{-\eta^{(l)}}\dz^{(l)},\ \text{gates }\alpha\big)$
        \State $\bh^{(l)}\gets\sigma(\ba^{(l)})$; store the Jacobian $\bJ^{(l)}$
    \EndFor
    \Statex \hspace{\algorithmicindent}\steprdo{Step 2: readout trajectory and its gradient}
    \State \revminor{$\tilde\bv\gets\bv-\bW^{(L)}_0\bh^{(L-1)}$}
    \State \revminor{$\be\gets\textsc{DeltaRuleGrad}\big(\bh^{(L-1)},\tilde\bv,\alpha,\eta^{(L)},\bW^{(L)}_0\big)$} \Comment{\cref{eq:dk}\revminor{; includes slow offset}}
    \Statex \hspace{\algorithmicindent}\stepbwd{Step 3: parallel backward pass}
    \State $\bg\gets\be/\eta^{(L)}$
    \For{$l=L-1,\dots,1$}
        \State $\dzh^{(l)}\gets\bJ^{(l)\top}\bg$
        \If{$l>1$}
            \State $\bg\gets\bW^{(l)\top}_0\dzh^{(l)}$
            \State $\bg\gets\bg+\textsc{TransposedScan}\big(\text{keys }\dz^{(l)},\ \text{values }{-\eta^{(l)}}\bh^{(l-1)},\ \text{gates }\alpha\big)\,\dzh^{(l)}$
        \EndIf
    \EndFor
    \State $\dz^{(l)}\gets\dz^{(l)}+\lambda\,(\dzh^{(l)}-\dz^{(l)})$ for every hidden layer $l$
\EndFor
\Statex \textit{Output}
\State $\bphi_{\bk}\gets$ features of $\bk$ from a final Step 1 (shifted); $\bphi_{\bq}\gets$ features of $\bq$ from the same scans, unshifted
\State \revminor{$\tilde\bv\gets\bv-\bW^{(L)}_0\bphi_{\bk}$}
\State \revminor{$\bo_{\mathrm{fast}}\gets\textsc{DeltaRule}\big(\text{queries }\bphi_{\bq},\ \text{keys }\bphi_{\bk},\ \text{values }\tilde\bv,\ \text{gates }\alpha,\ \text{step sizes }\eta^{(L)}\big)$}
\State \Return \revminor{$\bo\gets\bW^{(L)}_0\bphi_{\bq}+\bo_{\mathrm{fast}}$}
\end{algorithmic}
\end{algorithm}

\subsection{Shifted and unshifted scans}
\label{app:shift}
Every scan in \cref{alg:main} is one of two forms of the same recurrence, which differ only in whether token $t$ reads the state
before or after its own write. For a gated linear attention with keys $\bk_j$, values $\bv_j$ and gates $\alpha_j$, the state is
$\bS_t=\alpha_t\bS_{t-1}+\bv_t\bk_t^\top$, and the two reads of a query $\bq_t$ are
\begin{equation}
\begin{aligned}
    \text{unshifted:}&\quad \bo_t=\bS_t\bq_t=\sum_{j\le t}\Big(\prod_{i=j+1}^{t}\alpha_i\Big)\bv_j\,(\bk_j^\top\bq_t),\\
    \text{shifted:}&\quad \bo_t=\alpha_t\bS_{t-1}\bq_t=\sum_{j<t}\Big(\prod_{i=j+1}^{t}\alpha_i\Big)\bv_j\,(\bk_j^\top\bq_t).
\end{aligned}
    \label{eq:shift}
\end{equation}
The unshifted read is the standard one, in which the sum includes the current token $j=t$; the shifted read excludes it. We need both
because the sequential recurrence \cref{eq:rec-steps} evaluates everything about the key of token $t$ at the decayed weights
$\bar\bW_{t-1}$, before its own gradient step, and the query at $\bW_t$, after it. The hidden-layer pre-activation of key $t$,
\cref{eq:body-la}, is therefore a shifted read with $\bq_t=\bh^{(l-1)}_t$ and $\bv_j=-\eta^{(l)}_j\dz^{(l)}_j$; the transposed read of
Step 3, \cref{eq:body-la-T}, is a shifted read with the roles of keys and values exchanged; \kgrad{} reads the readout state before
token $t$'s write, \cref{eq:dk-chunk}, which is again shifted; and the final output pass reads the query-side features under the
updated weights, which is unshifted. \textbf{The shift is also what makes the fixed-point argument of \cref{sec:method:fixedpoint} work}: with a
shifted read no token's update depends on its own write, so all $T$ updates can be evaluated simultaneously from an estimate of
$\bD$; with an unshifted read the update of token $t$ would depend on $\dz_t$ itself and the parallel evaluation would not reproduce the
sequential recurrence. The use of shifted kernels for keys is consistent with standard formulation of online gradient descent used in GDN and TTT-MLP, where the gradient (on the keys) is evaluated after applying the decay (shifted), \revminor{and} the readout from the query is applied using the \revminor{updated} state (unshifted).

In the chunkwise form the difference is confined to the intra-chunk term. Writing the chunk's queries, keys and values as matrices
and $\bGamma_{\!/}$ for the matrix of decay ratios, the two reads are
\begin{equation}
    \text{unshifted:}\quad \bO=\bGamma\bQ\revminor{\bS_c^\top}+\operatorname{tril}\big(\bQ\bK^\top\odot\bGamma_{\!/}\big)\bV,
    \quad
    \text{shifted:}\quad \bO=\bGamma\bQ\revminor{\bS_c^\top}+\operatorname{tril}\big(\bQ\bK^\top\odot\bGamma_{\!/},-1\big)\bV,
    \label{eq:shift-chunk}
\end{equation}
where $\operatorname{tril}(\cdot,-1)$ is the strictly lower-triangular part. The boundary state $\bS_c$ contains only tokens of earlier
chunks, so the inter-chunk term is identical in both forms; only the causal mask of the intra-chunk attention changes, from
$j\le t$ to $j<t$. This is how the shifted kernels are implemented: we copied FLA's chunkwise gated-linear-attention and
gated-delta-rule kernels and changed the intra-chunk mask in the forward output kernel and its two backward kernels (for the
query/key gradients and for the value gradient) from a causal to a strictly causal comparison of the token indices, leaving the
inter-chunk state recurrence, the WY solve and the chunk-boundary bookkeeping untouched. The same one-line change applied to the
delta-rule kernel's final read gives the shifted delta-rule scan used for the hidden layer in \cref{app:body-delta} and, with the
query and value operands exchanged, \kgrad{}. At decode time ($T=1$) the shifted read is the read of the cached state before the
token's update, so no special case is needed.

\Cref{alg:main} is implemented with three chunkwise kernels of the FLA library and their transposes: the gated-linear-attention
scan for the hidden layers (\cref{eq:body-la} and its transpose \cref{eq:body-la-T}), the gated-delta-rule scan for the hidden layer
in the delta-rule form of \cref{eq:body-delta}, and the gated-delta-rule kernel with the query and value operands of its final read
exchanged for \kgrad{} (\cref{app:dk-simple}). The shift is implemented inside the chunk: the intra-chunk attention is strictly
lower-triangular for the key-side read and lower-triangular including the diagonal for the query-side read, and the inter-chunk state
is the same for both. Every iteration recomputes the hidden-layer scans from the current pseudo-labels. \revminor{In the reported
one-hidden-layer configuration, the slow hidden contribution $\bW^{(1)}_0\bk_t$} and the iteration-0 features are computed once.
\revminor{For deeper memories, the inputs of subsequent hidden layers change with the iteration, so their slow contributions are recomputed.} Training backpropagates through all $N$ iterations with the
kernels' chunkwise backward passes; iteration $n$ stores the pseudo-labels, the features and the Jacobians of $\sigma$, and the
readout states at chunk boundaries are recomputed in the backward pass rather than stored.

\subsection{Delta rule for the hidden layers}
\label{app:body-delta}
Step 1 evolves each hidden layer's fast weights by the gated linear attention \cref{eq:body-la}, which sums the writes
$-\eta^{(l)}_j\dz^{(l)}_j\bh^{(l-1)\top}_j$ without ever erasing. The same weights can be evolved by a gated delta rule. With the unit key
$\hat\bk_t=\bh^{(l-1)}_t/\|\bh^{(l-1)}_t\|$ and write rate $\beta_t=\eta^{(l)}_t$, the delta rule is
\begin{align}
    \Delta\bW^{(l)}_t &\;=\; \alpha_t\Delta\bW^{(l)}_{t-1} \;+\; \beta_t\big(\by_t-\br_t\big)\hat\bk_t^\top,
    \label{eq:body-delta}\\
    \br_t&\;=\;\alpha_t\Delta\bW^{(l)}_{t-1}\hat\bk_t,\qquad
    \by_t\;=\;\tilde\br_t-\frac{\eta^{(l)}_t\|\bh^{(l-1)}_t\|}{\beta_t}\,\dz^{(l)}_t ,\nonumber
\end{align}
where $\br_t$ is the read of the key from the decayed weights and the target $\by_t$ is built from $\tilde\br_t$, the same read taken
on the trajectory of the previous iteration. Substituting $\by_t$,
\begin{equation}
    \Delta\bW^{(l)}_t \;=\; \alpha_t\Delta\bW^{(l)}_{t-1} \;-\; \eta^{(l)}_t\,\dz^{(l)}_t\,\bh^{(l-1)\top}_t \;+\; \beta_t\big(\tilde\br_t-\br_t\big)\hat\bk_t^\top ,
\end{equation}
which is \cref{eq:rec-W} plus a correction that erases the part of the previous iteration's read that the current trajectory does not
reproduce. At the fixed point $\tilde\br_t=\br_t$ and the correction vanishes, so the two forms agree there; along the iterates the
delta-rule form erases stale writes rather than accumulating them. \revminor{It is computed with a chunkwise gated-delta-rule kernel;} \cref{tab:main} compares the two.

\subsection{The readout gradient}
\label{app:dk-simple}

Following the notation of \citet{yang2024parallelizing,tumma2026preconditioned}, write $\bS_t=\Delta\bW^{(L)\top}_t$ for the
\emph{fast} part of the readout, which decays as $\bar\bS_{t-1}=\alpha_t\bS_{t-1}$, and keep the slow readout $\bW^{(L)}_0$ as a fixed
offset, so that the weight token $t$ reads is $\bar\bS_{t-1}+\bW^{(L)\top}_0$. (Equivalently, the full state $\bS_t+\bW^{(L)\top}_0$
decays toward $\bW^{(L)\top}_0$: $\alpha_t(\bS_{t-1}+\bW^{(L)\top}_0)+(1-\alpha_t)\bW^{(L)\top}_0$.) With the corrected values
$\tilde\bv_t=\bv_t-\bW^{(L)}_0\bphi_t$ (or their centered form, \cref{app:centering}) and $\bu_t=\eta^{(L)}_t(\tilde\bv_t-\bar\bS_{t-1}^\top\bphi_t)$
the value that token $t$ writes, the fast readout's recurrence is $\bS_t=\bar\bS_{t-1}+\bphi_t\bu_t^\top$, and token $t$'s inner
loss and the gradient that Step 3 needs are
\begin{equation}
    \ell_t \;=\;\tfrac12\big\|\bar\bS_{t-1}^\top\bphi_t-\tilde\bv_t\big\|^2,\qquad
    \be_t \;=\;\eta^{(L)}_t\,\frac{\partial\ell_t}{\partial\bphi_t}\;=\;-\,\big(\bar\bS_{t-1}+\bW^{(L)\top}_0\big)\,\bu_t .
    \label{eq:dk-simple}
\end{equation}
The chunkwise gated-delta-rule kernel of \citet{yang2024parallelizing,yang2025gated}, as implemented in FLA, already forms every
quantity on the right: the chunk-boundary states, the intra-chunk decays and the corrected values $\bu_j$ of a chunk, which it obtains
jointly from the WY representation. \kgrad{} therefore reuses that kernel and changes only its final read, exchanging the roles of the
queries and the values; its cost equals one delta-rule pass. The chunkwise form follows.

\subsection{Chunkwise computation of the readout gradient}
\label{app:dk}

We use the chunk notation of \citet{yang2024parallelizing,yang2025gated,tumma2026preconditioned}. Partition the sequence into
chunks of $C$ tokens and, for a chunk, stack its features, values, step sizes and outputs as \revminor{$\bPhi\in\R^{C\times d}$, $\bV\in\R^{C\times d_v}$,}
$\bbeta=(\eta^{(L)}_t)_{t\in c}$ and $\bO$, with $\bS_c$ the readout state at the chunk's start and $\gamma_t=\prod_{i\le t}\alpha_i$ the
cumulative decay within the chunk. \revminor{Write $\gamma_{t,j}=\gamma_t/\gamma_j$, $\bGamma=\operatorname{diag}(\gamma_t)$,
$(\bGamma_{\!/})_{tj}=\gamma_{t,j}$, $\bGamma_C=\operatorname{diag}(\gamma_C/\gamma_j)$} and $\bL=\operatorname{diag}(\bbeta)$. In this transposed convention the gated delta rule advances the state by
$\bS_t=(\bI-\eta_t\bphi_t\bphi_t^\top)\,\bar\bS_{t-1}+\eta_t\bphi_t\tilde\bv_t^\top$, and the transitions of a chunk, applied in order
from the left, have the WY form $(\bI-\eta_C\bphi_C\bphi_C^\top)\cdots(\bI-\eta_1\bphi_1\bphi_1^\top)=\bI-\bPhi^\top\bU$ with
\begin{equation}
    \bU \;=\;\revminor{\bT_0}\,\bL\,\bPhi,\qquad \revminor{\bT_0}\;=\;\big(\bI+\operatorname{tril}(\bL\,\bPhi\bPhi^\top,-1)\big)^{-1},
    \label{eq:wy}
\end{equation}
a $C\times C$ triangular solve against the Gram matrix of the chunk's features. \revminor{Including the decay gives the solve
\[
    \bT=\bGamma\bT_0\bGamma^{-1}
    =\big(\bI+\operatorname{tril}((\bL\,\bPhi\bPhi^\top)\odot\bGamma_{\!/},-1)\big)^{-1}.
\]}
The rows $\bu_j$ of the corrected values
\begin{equation}
    \bU_{\bv}\;=\;\bT\,\bL\,\big(\tilde\bV-\revminor{\bGamma}\bPhi\,\bS_c\big)
    \label{eq:wy-values}
\end{equation}
are exactly the writes $\bu_j=\eta_j(\tilde\bv_j-\bar\bS_{j-1}^\top\bphi_j)$ of \cref{app:dk-simple}, expressed without the token-by-token
state, and the state entering token $t$ of the chunk is
\begin{equation}
    \bar\bS_{t-1} \;=\; \gamma_t\,\bS_{c} \;+\; \sum_{\substack{j\in c\\ j<t}}\gamma_{t,j}\,\bphi_j\,\bu_j^\top .
    \label{eq:chunk-state}
\end{equation}
The kernel computes, per chunk, $\bT$, $\bU_{\bv}$, the next boundary state
$\bS_{c+1}=\gamma_C\bS_c+\bPhi^\top\bGamma_{C}\bU_{\bv}$ (the inter-chunk recurrence, a short scan over chunks), and the output
\begin{equation}
    \bO \;=\; \bGamma\,\bPhi_{\bq}\,\bS_c \;+\; \operatorname{tril}\big(\bPhi_{\bq}\bPhi^\top\odot\bGamma_{\!/}\big)\,\bU_{\bv},
    \label{eq:chunk-out}
\end{equation}
where the $\operatorname{tril}$ mask is causal; this is \cref{eq:chunk-state}
substituted into $\bo_t=\bS_t^\top\bphi_{\bq,t}$ (the slow term $\bW^{(L)}_0\bphi_{\bq,t}$ is added outside the kernel), a product with
the boundary state plus a causal intra-chunk attention with scores $\bphi_{\bq,t}^\top\bphi_j$ and values $\bu_j$.

The feature gradient $\be_t=-(\bar\bS_{t-1}+\bW^{(L)\top}_0)\bu_t$ of \cref{eq:dk-simple} substitutes the same \cref{eq:chunk-state} and
gives, per token and per chunk,
\begin{equation}
\begin{aligned}
    \be_t &\;=\; -\,\big(\gamma_t\,\bS_c+\bW^{(L)\top}_0\big)\,\bu_t \;-\; \sum_{\substack{j\in c\\ j<t}}\gamma_{t,j}\,\big(\bu_j^\top\bu_t\big)\,\bphi_j ,\\
    \bE &\;=\; -\,\bGamma\,\bU_{\bv}\,\bS_c^\top-\bU_{\bv}\bW^{(L)}_0 \;-\; \operatorname{tril}\big(\bU_{\bv}\bU_{\bv}^\top\odot\bGamma_{\!/},-1\big)\,\bPhi ,
\end{aligned}
    \label{eq:dk-chunk}
\end{equation}
a product with the boundary state plus a causal intra-chunk attention in which the corrected values act as both queries and keys
and the features as values. Comparing \cref{eq:chunk-out,eq:dk-chunk}: the two expressions share $\bS_c$, $\bGamma$, $\bT$ and
$\bU_{\bv}$, and differ only in which operands enter the final read, $(\bPhi_{\bq},\bPhi,\bU_{\bv})$ for the output against
$(\bU_{\bv},\bU_{\bv},\bPhi)$ for the gradient, and in the strict lower-triangular mask, which excludes $j=t$ and implements the
shifted arrangement. \kgrad{} is therefore the gated-delta-rule kernel with the query and value operands of its final read
exchanged, with the offset $\bW^{(L)\top}_0$ entering its final read only; its cost equals that of one delta-rule pass. Because
$\bT$ depends on the features, which change with every iteration, it is recomputed in each of the $N$ iterations and in the final
scans. \revminor{In the reported one-hidden-layer configuration, the hidden delta-rule solve (\cref{app:body-delta}) is computed once
because its keys are the fixed inputs $\bk_t$. In deeper memories, this reuse applies only to the first hidden layer; subsequent
hidden-layer inputs change with the iteration.} The transposed reads of Step 3, \cref{eq:body-la-T}, have the same
structure with plain linear attention: values $-\eta^{(l)}_j\bh^{(l-1)}_j$ in place of $\bU_{\bv}$, and no triangular solve.

\subsection{Centering}
\label{app:centering}
The centered readout of \cref{sec:method:practical} is
\begin{equation}
    f_{\bW}(\bk_t)
    =\bW^{(L)}\bphi_t-\bW^{(L)}_0\bphi_{0,t}
    =\Delta\bW^{(L)}\bphi_t
    +\bW^{(L)}_0\big(\bphi_t-\bphi_{0,t}\big),
    \label{eq:center}
\end{equation}
the fast readout applied to the features plus the slow readout applied to their drift from the slow features. With it, the inner loss of token $t$ is
$\ell_t=\tfrac12\|\Delta\bar\bW^{(L)}_{t-1}\bphi_t+\bW^{(L)}_0(\bphi_t-\bphi_{0,t})-\bv_t\|^2$. Writing
$\tilde\bv_t=\bv_t-\bW^{(L)}_0(\bphi_t-\bphi_{0,t})$ for the corrected values, this is the uncentered loss with $\bv_t$ replaced by
$\tilde\bv_t$, so the fast readout $\Delta\bW^{(L)}$ follows the same gated delta rule as before on the corrected values, and its
trajectory is computed by the same kernel. The feature gradient of \cref{eq:dk} becomes
\begin{equation}
    \be_t \;=\; \eta^{(L)}_t\big(\Delta\bar\bW^{(L)}_{t-1}+\bW^{(L)}_0\big)^{\top}\big(\Delta\bar\bW^{(L)}_{t-1}\bphi_t-\tilde\bv_t\big),
\end{equation}
which \kgrad{} returns when its readout weight is offset by $\bW^{(L)}_0$; the offset enters the kernel's final read only. The output
is $\bo_t=\Delta\bW^{(L)}_t\bphi_{\bq,t}+\bW^{(L)}_0(\bphi_{\bq,t}-\bphi_{0,\bq,t})$, one gated-delta-rule pass on the corrected values plus
a matrix product. The slow features $\bphi_{0}$ of the keys are exactly those produced by iteration $0$, where $\bD=\mathbf 0$, so they
cost nothing extra; the slow features of the queries need one additional pass through the slow MLP. The fixed point of
\cref{sec:method:fixedpoint} is unchanged in form: it is the gradient sequence of the sequential recurrence with the centered readout.
$\bW^{(L)}_0$ is trained by backpropagation like every other slow parameter; because the forward pass depends on it through
$\bphi_t-\bphi_{0,t}$, its gradient is a true gradient and no stop-gradient is involved.

\section{Model configurations and training recipe}
\label{app:configs}
The reported experiments use the hard unit-ball projection $\bx/\max(1,\|\bx\|)$ specified below.
The soft clip is used only for the theoretical analysis in \cref{sec:method:theory,app:theory}.

All models share one backbone: a stack of pre-norm residual blocks, each a token-mixing layer followed by a SwiGLU MLP, with RMSNorm
and tied embeddings, 24 layers at both scales, $d=1024$ \revminor{at 340M and $d=2048$ at 1.3B. \rev{The default head layouts are 8 heads of
dimension 128 at 340M and 16 heads of dimension 128 at 1.3B; TTT-MLP and LaCT use the separate layouts below.}} The token-mixing layer is \method{}, Gated DeltaNet, Mamba-2, softmax attention with RoPE, or TTT-MLP; the recurrent
layers apply a short causal convolution of width 4 to the query, key and value projections and an output gate. Mamba-2 uses
expansion 1 and 4 groups, TTT-MLP its published layout of 16 heads of dimension 64 with a block of 16 tokens, and LaCT its published
recipe with a chunk of 2048 tokens, without the sliding-window branch, so that it is a pure fast-weight model. \method{} uses the
memory of \cref{sec:method:setup} with one hidden layer of width \revminor{$d_{\mathrm{head}}=128$} and SiLU, the delta rule for the hidden layer and centering
(\cref{sec:method:practical}), a per-token learning rate that is the sigmoid of a linear function of the input, the decay gate of Gated
DeltaNet shared by both layers, damping $\lambda=0.8$, and $N=2$ iterations during training unless stated. Every \method{} model starts
from the initialization of the Gated DeltaNet model of the same size, so that the two differ only in the memory.
\Cref{tab:configs} lists the models of \cref{tab:main}. All models are trained with FLAME on SlimPajama (parquet shards of the
100B-token subset) with the Llama tokenizer, sequence length 4096, 8 sequences per GPU with two accumulation steps on 8 GPUs (0.5M tokens
per step), AdamW ($\beta=(0.9,0.95)$, $\epsilon=10^{-8}$, weight decay $0.01$), peak learning rate $4\times10^{-4}$ with 1024 warmup steps and
cosine decay to $4\times10^{-5}$, gradient clipping at norm 1, bf16, seed 42 and the same data order. The 340M models train for 30k
steps and the 1.3B models for 190k steps; the 1.3B \method{} model trains on two nodes with a sharding that reproduces the one-node
data order. Every \method{} model is initialized from the Gated DeltaNet initialization of the same size (the slow weights of the
memory are new parameters, initialized with the standard scheme and zero fast weights), so that at initialization the two models differ only in the
memory.

\begin{table}[h]
\centering
\small
\setlength{\tabcolsep}{3pt}
\renewcommand{\arraystretch}{1.15}
\begin{tabular*}{\textwidth}{@{\extracolsep{\fill}}p{0.165\textwidth}rp{0.225\textwidth}p{0.27\textwidth}p{0.19\textwidth}@{}}
\toprule
\raggedright Model & Params & \raggedright Mixing layer & \raggedright Memory / state & \raggedright Notes \tabularnewline
\midrule
\multicolumn{5}{@{}l}{\textit{340M, 24 layers, $d=1024$}} \\
\raggedright Gated DeltaNet & 342M & \raggedright 8 heads $\times$ 128 & \raggedright $128\times128$ per head & \raggedright conv 4, output gate \tabularnewline
\raggedright Mamba-2 & 342M & \raggedright 8 heads $\times$ 128, expand 1, 4 groups & \raggedright $128\times128$ per head & \raggedright scalar decay \tabularnewline
\raggedright Transformer & 341M & \raggedright 8 heads $\times$ 128, RoPE & \raggedright KV cache & \raggedright YaRN for RULER \tabularnewline
\raggedright TTT-MLP & 349M & \raggedright 16 heads $\times$ 64, block 16 & \raggedright MLP $64\to256\to64$ per head & \raggedright inner LR base 0.1 \tabularnewline
\raggedright \revminor{LaCT fast-weight branch only} & 340M & \raggedright 4 heads, chunk 2048 & \raggedright SwiGLU fast weights & \raggedright Muon updates, conv 4 \tabularnewline
\raggedright \method{} & 348M & \raggedright 8 heads $\times$ 128 & \raggedright MLP $128\to128\to128$ per head & \raggedright $N=2$, $\lambda=0.8$, delta-rule body, centered \tabularnewline
\midrule
\multicolumn{5}{@{}l}{\textit{1.3B, 24 layers, $d=2048$}} \\
\raggedright Gated DeltaNet & 1.30B & \raggedright 16 heads $\times$ 128 &  &  \tabularnewline
\raggedright Mamba-2 & 1.30B & \raggedright 16 heads $\times$ 128, expand 1, 4 groups &  &  \tabularnewline
\raggedright Transformer & 1.30B & \raggedright 16 heads $\times$ 128, RoPE &  &  \tabularnewline
\raggedright \method{} & 1.31B & \raggedright 16 heads $\times$ 128 & \raggedright MLP $128\to128\to128$ per head & \raggedright $N=2$ \tabularnewline
\bottomrule
\end{tabular*}
\caption{Models of \cref{tab:main}. Parameter counts include embeddings. The \method{} memory has a hidden layer of width $d_{\text{head}}$
with SiLU, a bias, the unit-ball projection on the features, and a learned per-head gain on the fast pre-activation; its learning rates
are $\eta_t=\operatorname{sigmoid}(\mathbf{w}_\eta^\top\bx_t+b_\eta)$ per head and layer, its decay is the Gated DeltaNet gate shared by both layers.}
\label{tab:configs}
\end{table}

\section{Evaluation protocol}
\label{app:eval}
\paragraph{Perplexity and zero-shot.} \revminor{WikiText-103} word perplexity and LAMBADA (OpenAI) perplexity, and accuracy on PIQA,
HellaSwag (normalized), ARC-easy, ARC-challenge (normalized), WinoGrande, BoolQ and SciQ, all zero-shot with lm-evaluation-harness
\citep{gao2023lmeval}. \method{} models use the chunkwise prefill at their training $N$ unless stated otherwise. \revs{Both new 1.3B training variants use $N=2$ for the main task evaluations, including recall, NIAH and LongBench; \cref{tab:evaln} separately varies the evaluation-time iteration count.}
\paragraph{Recall.} SWDE, FDA, TriviaQA (the cloze version of \citet{arora2024just}) and DROP are generation tasks: the model
greedily decodes an answer after the prompt and the score is whether the gold answer appears in the generation, case-insensitively,
on the full evaluation sets. \method{} prefills the prompt with the chunkwise iteration at \revs{the evaluation-time $N$ specified above} and decodes token by token
with the exact recurrence; TTT-MLP decodes with a block-faithful cache at batch 1; the other models use their standard cached
decoding.
\paragraph{RULER.} The eight needle-in-a-haystack tasks of RULER \citep{hsieh2024ruler} (single 1--3, multikey 1--3, multiquery,
multivalue) at \revminor{4K, 8K, 16K} and \revminor{32K} tokens with 100 samples per task; the table reports the average. Attention models use YaRN
\citep{peng2024yarn} with a scale of the evaluation length over 4096.
\paragraph{Synthetic tasks.} Two-layer models of width 256 (head dimension 128) with the same backbone as the language models, trained
from scratch for 4k steps with AdamW, batch 128, over learning rates $\{3\times10^{-4},10^{-3},3\times10^{-3}\}$ and three seeds; the
long-horizon settings (parity at 512 and 1024, MQAR at 512 with 32 pairs) are also trained for 8k steps at
$\{10^{-4},3\times10^{-4}\}$, and MQAR with 64 pairs at width 512. \Cref{fig:synthetics} reports, per model and setting, the learning rate
(and budget) with the best mean final accuracy over the three seeds. \method{} models train and evaluate at $N=2$ (or $N=5$).
\revs{The group word problems of \cref{tab:state-tracking} use the same models and learning rates with batch 64, for 4k steps or, where
stated, 16k steps; \method{} is trained and evaluated at each listed $N$, and every entry uses the learning rate with the best mean
final accuracy over the three seeds.}
\paragraph{Per-position perplexity.} 698 held-out SlimPajama documents (validation and test splits) of at least 4097 tokens, each
contributing its first 4096 tokens, so that every position is the same number of tokens into a document; the mean NLL per position
is binned in 27 log-spaced bins and reported as perplexity with one standard error.

\section{Additional results}
\label{app:results}

\subsection{Synthetic nonlinear-state tasks}
\label{app:synthetics}
\begin{figure}[h]
    \centering
    \includegraphics[width=\textwidth]{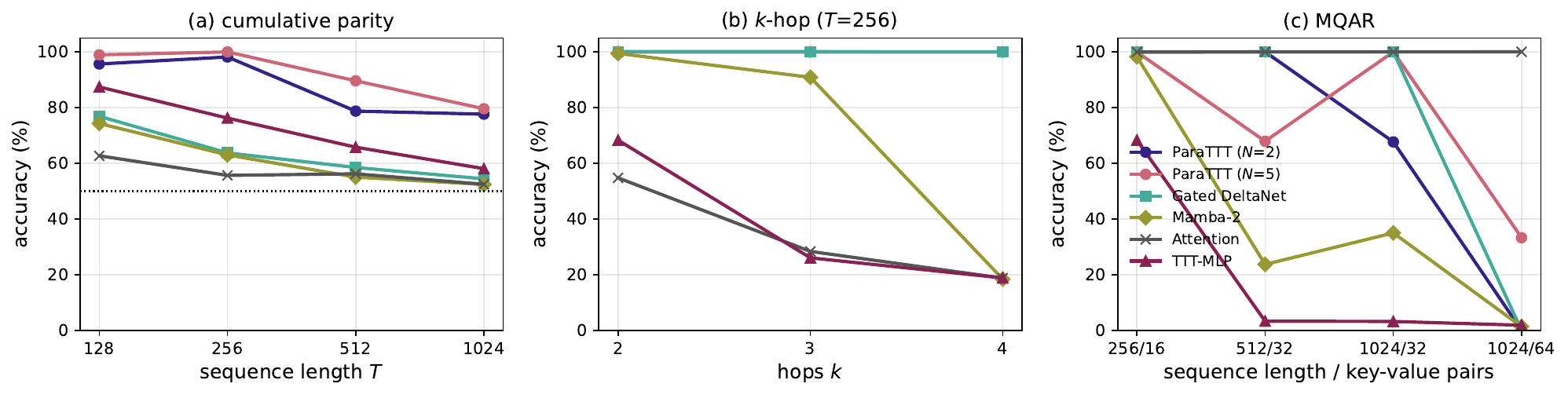}
    \caption{Synthetic nonlinear-state tasks (two layers, width \revminor{256 except width 512 for MQAR with 64 pairs;} mean over three seeds at the best learning rate). (a) Cumulative
    parity against sequence length; the dotted line is chance. (b) $k$-hop pointer following at $T=256$. (c) Multi-query associative
    recall for increasing sequence length and number of key-value pairs.}
    \label{fig:synthetics}
\end{figure}

Three synthetic tasks isolate what a nonlinear memory adds, each learned from scratch by two-layer models of width 256
\revminor{(width 512 for MQAR with 64 pairs)} with the backbone of \cref{app:configs} (protocol in \cref{app:eval}; \cref{fig:synthetics}). \emph{Cumulative parity} asks, at every position,
for the parity of the number of ones seen so far; \emph{$k$-hop} asks for the token reached by following $k$ pointer links back through
the sequence; \emph{multi-query associative recall} (MQAR) asks for the values of previously stored keys.

Parity probes whether a memory can maintain a state that flips with selected inputs. It is not a universal separation between
nonlinear and linear memories: negative transition eigenvalues allow linear RNNs to solve it. The relevant limitation concerns
finite-precision linear RNNs restricted to nonnegative real eigenvalues at arbitrary sequence lengths \citep{grazzi2024unlocking}.
Our finite-length experiments test the architectures and training settings used here. In \cref{fig:synthetics}(a), Gated DeltaNet
and Mamba-2 fall from $74$--$77\%$ at length 128 to $63\%$ at 256 and $52$--$55\%$ at 1024, approaching chance, and the two-layer
attention model stays below $63\%$ at every length, whereas \method{} reaches $96$--$100\%$ at lengths up to 256, $79$--$90\%$ at 512
and $78$--$80\%$ at 1024, with the model trained at $N=5$ best at every length.
Turning the loop off ($N=0$) drops \method{} to chance, indicating that the iteration is important in this setup. TTT-MLP sits
between the two groups ($87\%$ at length 128, $58\%$ at 1024). Its block form evaluates gradients at the weights entering each
16-token block, although each token's readout incorporates preceding within-block updates \citep{sun2024learning}.
The frozen gradient evaluations differ from the online recurrence that \method{} targets; these results alone do not establish
that this difference causes the parity gap.

The other two tasks are solvable by a linear memory and serve as controls. On $k$-hop (\cref{fig:synthetics}b) the recurrent models
with an erasing update, \method{} and Gated DeltaNet, solve every $k$, Mamba-2 breaks at $k=4$, and the two-layer attention model cannot
compose hops at this depth. MQAR (\cref{fig:synthetics}c) is a capacity test: Gated DeltaNet and \method{} solve it up to 32 pairs, with
\method{} more seed-sensitive at length 1024 and the $N=5$ model the more reliable of the two, and only attention solves 64 pairs, which
\revminor{exceeds the empirical capacity of the recurrent models in this setting.} \method{} thus keeps the linear recurrences' recall while achieving
stronger state-tracking performance in these experiments.

\begin{revblocks}
\paragraph{State tracking in groups.}
To test the nonlinear memory on tasks that require successive state updates, we add group word problems
\citep{liu2023transformers,merrill2024illusion}: the tokens are uniformly random elements of a finite group, and the target at every
position is the running product of the elements seen so far. For $\mathbb{Z}_2$ this is the cumulative parity task above; $S_3$ is the
smallest non-abelian group. We train \method{} at $N\in\{1,2,5,8,10\}$ and evaluate each model at its training $N$, with Gated
DeltaNet and attention as baselines, all with the two-layer models of width 256 used above (protocol in \cref{app:eval}).
\Cref{tab:state-tracking} reports the mean and the range over three seeds.

On parity, \method{} with two iterations is already well above the linear baseline, and models trained with more iterations solve the task. The shorter $S_3$ task gives a similar separation: every \method{} seed at $N=2$ outperforms every Gated DeltaNet seed in the standard training budget. Giving Gated DeltaNet four times as many steps does not close this gap, while the longer-trained \method{} models with more iterations approach perfect accuracy. More iterations often help, although the shorter runs vary substantially across seeds and their mean accuracy is not monotonic in $N$. The longer $S_3$ task remains unsolved by all models in the tested budget.
\end{revblocks}

\begin{table}[h]
\centering
\small
\setlength{\tabcolsep}{4pt}
\begin{tabular}{@{}lccccccc@{}}
\toprule
 & \multicolumn{5}{c}{\method{}, trained and evaluated at $N$} & & \\
\cmidrule(lr){2-6}
Task & $N=1$ & 2 & 5 & 8 & 10 & GDN & Attention \\
\midrule
Parity ($\mathbb{Z}_2$), length 256 & 78.1 & 97.6 & 100.0 & 100.0 & 100.0 & 63.3 & 55.6 \\
 & {\scriptsize 69.7--93.7} & {\scriptsize 96.1--99.3} & {\scriptsize 100--100} & {\scriptsize 100--100} & {\scriptsize 100--100} & {\scriptsize 59.7--67.2} & {\scriptsize 55.4--56.0} \\
$S_3$, length 64, 4k steps & 48.7 & 62.7 & 81.1 & 92.9 & 80.8 & 38.1 & 31.5 \\
 & {\scriptsize 39.0--55.4} & {\scriptsize 55.7--69.0} & {\scriptsize 53.1--98.0} & {\scriptsize 90.3--97.7} & {\scriptsize 43.3--99.8} & {\scriptsize 34.4--43.4} & {\scriptsize 30.6--32.2} \\
$S_3$, length 64, 16k steps & 62.1 & 81.0 & 98.7 & 99.9 & 99.9 & 47.4 & 40.6 \\
 & {\scriptsize 50.1--81.4} & {\scriptsize 71.8--85.9} & {\scriptsize 97.9--99.7} & {\scriptsize 99.8--99.9} & {\scriptsize 99.7--100} & {\scriptsize 45.7--48.2} & {\scriptsize 39.9--41.2} \\
\bottomrule
\end{tabular}
\caption{\revs{Group word problems: accuracy (\%) of the running product at every position, mean over three seeds with the range of the
seeds below. Two-layer models of width 256; \method{} is trained and evaluated at the given $N$.}}
\label{tab:state-tracking}
\end{table}

\paragraph{Evaluation at other iteration counts.} \Cref{tab:evaln} gives the full sweeps behind \cref{fig:perpos-evaln}(c,d),
including the 1.3B model, on \revminor{WikiText-103,} LAMBADA and SWDE.
\revs{The last two blocks are the two 1.3B variants of \cref{sec:exp:results}, which change only the number of iterations during training.}

\begin{table}[h]
\centering
\small
\setlength{\tabcolsep}{5pt}
\begin{tabular}{@{}llrrrrrrr@{}}
\toprule
Model & Metric & $N=0$ & 1 & 2 & 3 & 5 & 7 & 10 \\
\midrule
340M, trained at $N=2$ & \revminor{WikiText-103} & 61.3 & 25.76 & \textbf{25.46} & 25.54 & 25.67 & 25.75 & 25.87 \\
 & LAMBADA & 332 & 28.43 & \textbf{26.93} & 27.37 & 28.21 & 28.40 & 28.60 \\
 & SWDE & 7.4 & 39.3 & \textbf{41.9} & 40.8 & 39.4 & 39.6 & 38.7 \\
340M, trained at $N=5$ & \revminor{WikiText-103} & 61.7 & 26.29 & 25.63 & 25.51 & \textbf{25.47} & 25.49 & 25.52 \\
 & LAMBADA & 375 & 31.60 & 29.31 & 28.69 & \textbf{28.44} & 28.52 & 28.72 \\
 & SWDE & 5.0 & 33.8 & 43.1 & 44.0 & \textbf{45.0} & 44.2 & 43.8 \\
340M, \revminor{zero-init readout} & \revminor{WikiText-103} & 26.28 & 26.09 & 26.06 & 26.06 & 26.08 & 26.09 & 26.12 \\
 & LAMBADA & 33.84 & 32.97 & 32.62 & 32.40 & 32.14 & 32.05 & 32.01 \\
 & SWDE & 34.5 & 36.7 & 36.6 & 36.8 & 36.2 & 35.8 & 35.8 \\
340M, linear-attention body & \revminor{WikiText-103} & 55.4 & 26.43 & \textbf{25.62} & 25.75 & 26.00 & 25.74 & 25.64 \\
 & LAMBADA & 179 & 26.76 & \textbf{25.88} & 26.08 & 26.11 & 26.13 & 26.15 \\
 & SWDE & 7.7 & 29.4 & 28.4 & 28.7 & 28.9 & 28.7 & 27.7 \\
1.3B, trained at $N=2$ & \revminor{WikiText-103} & 73.1 & 15.36 & \textbf{15.21} & 15.32 & 15.49 & 15.58 & 15.70 \\
 & LAMBADA & 348 & \textbf{8.94} & 9.00 & 9.40 & 9.91 & 10.16 & 10.38 \\
 & SWDE & 6.8 & 62.4 & \textbf{62.8} & 58.2 & 53.6 & 48.9 & 44.4 \\
\revs{1.3B, last 11B at $N=5$} & \revs{WikiText-103} & \revs{105.1} & \revs{15.56} & \revs{15.33} & \revs{15.25} & \revs{\textbf{15.24}} & \revs{15.25} & \revs{15.31} \\
 & \revs{LAMBADA} & \revs{387} & \revs{8.82} & \revs{\textbf{8.77}} & \revs{8.88} & \revs{9.13} & \revs{9.33} & \revs{9.56} \\
 & \revs{SWDE} & \revs{5.1} & \revs{65.3} & \revs{\textbf{67.8}} & \revs{67.7} & \revs{66.0} & \revs{64.9} & \revs{62.6} \\
\revs{1.3B, Poisson $N$} & \revs{WikiText-103} & \revs{52.3} & \revs{15.30} & \revs{\textbf{15.28}} & \revs{\textbf{15.28}} & \revs{15.29} & \revs{15.31} & \revs{15.33} \\
 & \revs{LAMBADA} & \revs{122} & \revs{\textbf{9.35}} & \revs{9.39} & \revs{9.36} & \revs{9.41} & \revs{9.46} & \revs{9.47} \\
 & \revs{SWDE} & \revs{7.6} & \revs{62.4} & \revs{\textbf{63.6}} & \revs{\textbf{63.6}} & \revs{\textbf{63.6}} & \revs{62.9} & \revs{61.9} \\
\bottomrule
\end{tabular}
\caption{Perplexity \revminor{(WikiText-103,} LAMBADA) and SWDE accuracy of the \method{} models as a function of the number of iterations at
evaluation. Bold: best $N$ for the centered models.}
\label{tab:evaln}
\end{table}

\paragraph{LongBench.}
\cref{tab:longbench} reports LongBench at four context lengths for every model in \cref{tab:main}.

\begin{table}[htbp]
\centering
\small
\begin{tabular}{@{}lrrrr@{}}
\toprule
Model & \revminor{4K} & \revminor{8K} & \revminor{16K} & \revminor{32K} \\
\midrule
\multicolumn{5}{@{}l}{\textit{340M, 24 layers, 15.7B tokens}} \\
GDN & 11.9 & 12.1 & 12.3 & 12.2 \\
Mamba2 & 11.8 & 11.9 & 11.9 & 11.9 \\
Transformer & 10.9 & 11.5 & 10.9 & 10.2 \\
TTT-MLP & 9.8 & 9.0 & 8.1 & 8.1 \\
\revminor{LaCT fast-weight branch only} & 4.9 & 4.5 & 4.2 & 4.2 \\
\textbf{\methodshort{} (ours)} & 12.4 & 12.5 & \underline{12.4} & 12.1 \\
\quad \revminor{zero-init readout} & \underline{12.6} & \underline{12.6} & \textbf{12.5} & \textbf{12.5} \\
\quad linear-attention body & 9.5 & 9.6 & 9.3 & 9.7 \\
\quad trained at \revminor{$N=5$} & \textbf{12.7} & \textbf{12.7} & \textbf{12.5} & \underline{12.4} \\
\midrule
\multicolumn{5}{@{}l}{\textit{1.3B, 24 layers, 100B tokens}} \\
GDN & 17.4 & 17.7 & 17.8 & 17.7 \\
Mamba2 & 15.1 & 15.1 & 15.1 & 15.2 \\
Transformer & \textbf{19.3} & \textbf{20.8} & \textbf{21.0} & \textbf{19.2} \\
\textbf{\methodshort{} (ours)} & 16.4 & 16.9 & 17.2 & 17.3 \\
\quad \revs{last 11B at $N=5$} & \revs{17.6} & \revs{17.6} & \revs{17.9} & \revs{17.8} \\
\quad \revs{Poisson $N$} & \revs{\underline{18.8}} & \revs{\underline{18.5}} & \revs{\underline{18.0}} & \revs{\underline{18.0}} \\
\bottomrule
\end{tabular}
\caption{LongBench: average score over the 16 English tasks (100 samples per task) at context lengths \revminor{4K--32K,} for the models of
\cref{tab:main}. Best per column and size in bold, second best underlined. All models trained at \revminor{4K;} \method{} is within a point of
Gated DeltaNet at every length and scale for the original $N=2$ models, and attention leads at 1.3B. \revs{The two 1.3B variants are those of \cref{tab:main}, evaluated at $N=2$; the Poisson-$N$ variant is above Gated DeltaNet at every length.}}
\label{tab:longbench}
\end{table}

\subsection{\revs{Comparisons with sliding-window attention}}
\label{app:swa}
The published TTT architectures that scale beyond the original TTT-Linear and TTT-MLP do not evaluate their memory alone. LaCT
\citep{zhang2025lact} pairs its large-chunk fast-weight update with a sliding-window attention branch in every layer and reports
only the combination; Titans \citep{behrouz2024titans} reports its long-context results for the hybrids that add attention to the
memory (MAC, MAG and MAL). A window of 2048 tokens covers half of a 4096-token training context, so the attention branch alone
already accounts for local modeling, and the fast-weight memory only has to contribute beyond the window.

\revs{The LaCT configuration without SWA in \cref{tab:main} performs poorly on perplexity and recall. Its large-chunk update provides one reason to expect local attention to matter: the gradients for a chunk are evaluated at its incoming weights, so the update does not follow the token-by-token recurrence within that chunk. However, this row alone does not isolate the contribution of SWA; its configuration is documented separately in \cref{app:configs}. We therefore compare all three recurrences with the same sliding-window branch below. The \method{} models in \cref{tab:main} remain pure recurrences, with an update at every token and no attention branch.}

\begin{table}[t]
\centering

\setlength{\tabcolsep}{3.5pt}
\resizebox{\textwidth}{!}{%
\begin{tabular}{@{}l rr rrrr rrrr rrrr@{}}
\toprule
 & \multicolumn{2}{c}{Perplexity $\downarrow$} & \multicolumn{4}{c}{Recall $\uparrow$} & \multicolumn{4}{c}{LongBench $\uparrow$} & \multicolumn{4}{c}{NIAH $\uparrow$} \\
\cmidrule(lr){2-3} \cmidrule(lr){4-7} \cmidrule(lr){8-11} \cmidrule(l){12-15}
Model & Wiki & Lamb & SWDE & FDA & TQA & DROP & 4K & 8K & 16K & 32K & 4K & 8K & 16K & 32K \\
\midrule
\multicolumn{15}{@{}l}{\textit{With a 2048-token sliding-window attention branch}} \\
LaCT (as published) & 25.45 & 32.74 & 53.2 & 26.8 & 37.2 & 13.7 & 11.3 & 10.3 & 9.0 & 8.6 & 31.3 & 12.4 & 1.8 & 0.0 \\
LaCT + norm & 24.99 & 33.68 & 58.4 & 28.4 & 38.9 & \textbf{15.1} & 12.0 & 11.4 & 11.1 & 10.7 & 32.5 & 16.4 & 6.1 & 1.4 \\
GDN & 26.52 & 29.79 & 37.3 & 18.2 & 37.6 & 13.7 & 12.8 & 13.0 & 12.9 & 13.0 & 28.7 & 18.8 & 13.7 & 13.5 \\
GDN + norm & 25.37 & 25.82 & 58.5 & 27.2 & 38.9 & 15.0 & 12.6 & 12.7 & 12.6 & 12.7 & 32.3 & 16.6 & 6.9 & 5.2 \\
\methodshort{} ($N=2$) & 25.37 & 29.31 & 46.4 & 20.7 & \textbf{40.7} & 14.9 & 12.9 & 12.5 & 12.2 & 12.1 & \textbf{36.3} & \textbf{21.9} & \textbf{17.6} & \textbf{16.1} \\
\methodshort{} + norm ($N=2$) & \textbf{24.67} & \textbf{24.10} & \textbf{59.8} & \textbf{29.9} & 39.5 & 14.6 & \textbf{14.7} & \textbf{15.0} & \textbf{14.5} & \textbf{14.5} & 33.8 & 16.8 & 8.4 & 6.0 \\
\midrule
\multicolumn{15}{@{}l}{\textit{Without a window (\cref{tab:main,tab:longbench})}} \\
LaCT & 463.2 & $9.4{\times}10^{4}$ & 0.1 & 0.0 & 0.1 & 7.6 & 4.9 & 4.5 & 4.2 & 4.2 & 0.0 & 0.0 & 0.0 & 0.0 \\
GDN & 26.57 & 29.66 & 32.5 & 17.8 & 35.7 & 15.3 & 11.9 & 12.1 & 12.3 & 12.2 & 25.9 & 17.0 & 13.7 & 9.3 \\
\methodshort{} ($N=2$) & 25.46 & 26.93 & 41.9 & 20.3 & 39.6 & 14.4 & 12.4 & 12.5 & 12.4 & 12.1 & 32.6 & 20.7 & 15.6 & 16.0 \\
\bottomrule
\end{tabular}}
\caption{\revs{340M models with and without a 2048-token sliding-window attention branch; same recipe, data and seed as
\cref{tab:main}. ``+ norm'' applies RMS normalization to the attention's queries, keys and values. LaCT without a window is the
configuration of \cref{tab:main}, detailed in \cref{app:configs}. LaCT (as published) is a second run of the published hybrid configuration;
the first run reaches $25.51$ on WikiText-103 and $33.3$ on LAMBADA. Best per column among the models with a window in bold.}}
\label{tab:swa-matched}
\end{table}

\begin{revblocks}
\paragraph{Evaluating all recurrences with SWA.} We give \method{} and Gated DeltaNet the same kind of SWA branch as LaCT: a 2048-token sliding-window attention whose output is summed with the recurrent output. The runs share the training recipe, data and seed of \cref{tab:main}, but are not matched in training compute. We also train each model with per-head RMS normalization of the attention queries, keys and values (``+ norm''). \Cref{tab:swa-matched} reports the results.

With a normalized window, \method{} is ahead of both LaCT variants on perplexity, most recall tasks, LongBench and NIAH; DROP is the exception. The benefit over LaCT therefore remains when both models have access to local attention.

The comparison also shows that Gated DeltaNet and \method{} benefit more from SWA on SWDE and FDA when the branch is normalized. Their short convolutions are initialized with small weights (standard deviation $0.02$), which can leave the attention queries and keys small and the attention close to uniform. This suggests that calibration of the attention branch contributes to the different gains from SWA. The published LaCT hybrid has no short convolution; adding one in our experiment made training unstable. These observations do not isolate an inherent synergy between LaCT and SWA.

Normalization also reduces long-range NIAH accuracy for Gated DeltaNet and \method{}. One possible explanation is that the models learn to rely more on the local window, which cannot retrieve a needle beyond its receptive field. We leave this as a hypothesis.
\end{revblocks}

\subsection{Centering ablation}
\label{app:centering-ablation}
\cref{tab:centering} isolates the readout parameterization. All three models share the recipe, initialization and data order of
\cref{app:configs} and differ only in how the readout layer $\bW^{(L)}$ is parameterized:
\begin{itemize}
    \item \textbf{Zero-initialized readout.} $\bW^{(L)}_0=\mathbf 0$ and $f_{\bW}(\bk_t)=\Delta\bW^{(L)}\bphi_t$, the standard
    delta-rule state that starts at zero and decays toward zero (the \revminor{``zero-init readout''} row of \cref{tab:main}). By \cref{eq:dk} the hidden
    layers receive no gradient until the fast readout has grown.
    \item \textbf{Learned initial readout, uncentered.} A slow $\bW^{(L)}_0$ is learned and $f_{\bW}(\bk_t)=(\Delta\bW^{(L)}+\bW^{(L)}_0)\bphi_t$:
    the \revminor{full} weights decay toward $\bW^{(L)}_0$ instead of zero, so the hidden layers receive gradient from the first token, but the output
    carries the bias $\bW^{(L)}_0\bphi_t$ before any test-time learning has taken place. This model has exactly the parameters of the
    centered one.
    \item \textbf{Centered readout.} \cref{eq:center}, the \method{} default.
\end{itemize}
Read against the Gated DeltaNet reference, the three rows form a ladder. The zero-initialized loop already improves recall over the
linear recurrence (SWDE $+4$, NIAH at \revminor{4K} $+3$) but costs $3$ points of LAMBADA perplexity. The learned initial readout repairs LAMBADA to
the Gated DeltaNet level and leaves recall essentially unchanged, showing that giving the hidden layers a gradient at the start of a sequence matters.
Centering buys the rest with the same parameters: $0.7$ \revminor{WikiText-103} and $2.1$ LAMBADA perplexity, $4$ points of SWDE and $2.5$ of
TriviaQA, and the largest NIAH gain at \revminor{32K.} The commonsense tasks move within noise across the ladder, and each row is a single seed.
The gain of centering over the uncentered learned readout is therefore not explained by the extra slow readout: it comes from removing the
bias $\bW^{(L)}_0\bphi_t$ so that the readout at the start of a sequence is zero while its Jacobian is not.

\begin{table}[t]
\centering
\setlength{\tabcolsep}{3pt}
\resizebox{\textwidth}{!}{%
\begin{tabular}{@{}l rr rrrrrrr rrrr rrrr rrrr@{}}
\toprule
 & \multicolumn{2}{c}{Perplexity $\downarrow$} & \multicolumn{7}{c}{Zero-shot acc.\ $\uparrow$} & \multicolumn{4}{c}{Recall $\uparrow$} & \multicolumn{4}{c}{NIAH $\uparrow$} & \multicolumn{4}{c}{LongBench $\uparrow$} \\
\cmidrule(lr){2-3} \cmidrule(lr){4-10} \cmidrule(lr){11-14} \cmidrule(lr){15-18} \cmidrule(lr){19-22}
Readout & Wiki & Lamb & PIQA & Hella & ARC-e & Wino & ARC-c & BoolQ & SciQ & SWDE & FDA & TQA & DROP & \revminor{4K} & \revminor{8K} & \revminor{16K} & \revminor{32K} & \revminor{4K} & \revminor{8K} & \revminor{16K} & \revminor{32K} \\
\midrule
Gated DeltaNet & 26.57 & 29.66 & 65.2 & 36.5 & 47.1 & 51.5 & 24.1 & 56.6 & 78.4 & 32.5 & 17.8 & 35.7 & 15.3 & 25.9 & 17.0 & 13.7 & 9.3 & 11.9 & 12.1 & 12.3 & 12.2 \\
\midrule
Zero-init & 26.06 & 32.62 & 64.5 & 36.8 & 48.4 & 50.6 & \textbf{25.3} & 58.2 & 78.9 & 36.6 & 19.6 & 37.0 & 14.3 & 29.0 & \textbf{21.1} & 15.0 & 12.2 & \textbf{12.6} & \textbf{12.6} & \textbf{12.5} & \textbf{12.5} \\
Learned init, uncentered & 26.13 & 29.06 & \textbf{65.8} & 36.7 & 46.1 & \textbf{52.6} & 24.4 & 59.7 & \textbf{80.4} & 37.7 & 19.2 & 37.1 & 13.5 & 32.5 & 19.2 & 14.4 & 14.4 & 11.6 & 11.2 & 11.2 & 11.4 \\
Centered (\methodshort{}) & \textbf{25.46} & \textbf{26.93} & 65.3 & \textbf{37.7} & \textbf{48.6} & 51.2 & 24.1 & \textbf{59.8} & 79.0 & \textbf{41.9} & \textbf{20.3} & \textbf{39.6} & \textbf{14.4} & \textbf{32.6} & 20.7 & \textbf{15.6} & \textbf{16.0} & 12.4 & 12.5 & 12.4 & 12.1 \\
\bottomrule
\end{tabular}}
\caption{Centering ablation at 340M. Three parameterizations of the readout layer, trained with the same recipe, initialization and data order; the gated DeltaNet baseline is shown for reference. Columns as in \cref{tab:main}; best of the three \methodshort{} variants per column in bold.}
\label{tab:centering}
\end{table}

\paragraph{Centering for TTT-MLP.}
We also trained TTT-MLP with the centered readout of \cref{eq:center}. It does not require centering: the results moved by
about two points in either direction (SWDE $33.1\to35.7$, FDA $16.2\to18.2$, perplexity unchanged within noise), so we report the
original recipe.

\begin{revblocks}
\subsection{Prefill--decode mismatch on long sequences}
\label{app:mismatch}
Prompts are processed by the chunkwise prefill with $N$ refinement iterations, whereas generated tokens are processed by the exact
recurrence, which needs no iteration (\cref{sec:method:alg}). The two agree only in the limit of the iteration, so a model trained
at a finite $N$ decodes with a slightly different memory than the one it was trained with. \Cref{fig:convergence}(b,c) compares the
two on 512-token sequences. Here we measure whole documents and the effect on decoding, using the held-out SlimPajama documents of
\cref{fig:perpos-evaln}(a,b) truncated to 4K, 16K or 32K tokens (16 documents at 4K, 8 at 16K and at 32K). Besides the two
\method{} models of \cref{tab:main}, we include two variants of the 1.3B model that differ only in the iteration count during
training: one switches to $N=5$ for the last 11B of its 100B training tokens, and one samples $N$ per layer from a Poisson distribution with parameter 2, resampling zero and capping the result at 5, throughout training. All models are trained at 4K tokens.

\paragraph{Decoding after a prefill.}
We prefill the first part of each document at $N=2$ and process the remaining tokens one at a time with the exact recurrence. Both evaluations use the reference tokens, so the sequence stays fixed while the recurrent computation changes. This teacher-forced evaluation uses the decoding update, but does not measure free-running generation quality. We compare perplexity on the continuation with its perplexity when the whole document is prefilled at $N=2$ (\cref{tab:mismatch}).

The mismatch has a measurable cost for the models trained at a fixed $N=2$, and the penalty grows with continuation length for the submitted 1.3B model. Both training changes substantially reduce it. With randomized $N$, the increase stays below $2\%$ even when the continuation is four times the training length, and remains small throughout that span.

To test exact prompt processing as well, we run the exact recurrence from the first token of each 4K document. The whole-document column in \cref{tab:mismatch} shows the same pattern: the submitted checkpoints incur a perplexity penalty, which the new training recipes reduce.

\paragraph{Agreement over whole documents.}
\Cref{tab:mismatch-agree} gives the top-1 agreement between the prefill of a whole document and the exact recurrence run from its
first token. Agreement is lower than on 512-token sequences and decreases with the document length. Beyond the training length the
prefill and the exact recurrence drift apart gradually, and additional iterations close only part of the gap. For every tested model and length, the $N=2$ prefill has a lower perplexity than the exact recurrence. As a reference for numerical differences between implementations, the chunkwise and token-by-token evaluations of Gated DeltaNet agree on $99.0$--$99.3\%$ of tokens in bf16, despite computing the same function. This reference does not establish how much of \method{}'s gap comes from numerical precision. The exact recurrence is the fixed
point of the iteration, but a model trained at a fixed small $N$ is not trained to operate at that fixed point over long sequences;
randomizing $N$ during training reduces the effect on decoding (\cref{tab:mismatch}) but does not remove the drift at 32K
(\cref{tab:mismatch-agree}).
\end{revblocks}

\begin{table}[h]
\centering
\small
\setlength{\tabcolsep}{4pt}
\begin{tabular}{@{}lrrrr@{}}
\toprule
 & Whole document & \multicolumn{3}{c}{Prefilled + decoded tokens} \\
\cmidrule(l){3-5}
Model & 4K & 256 + 3.8K & 8K + 8K & 16K + 16K \\
\midrule
340M & 2.4 & 2.4 & 3.3 & 4.9 \\
1.3B & 5.1 & 4.6 & 6.4 & 10.4 \\
1.3B, last 11B at $N=5$ & 1.9 & 1.6 & 2.8 & 3.8 \\
1.3B, Poisson $N$ & 1.1 & 0.9 & 1.5 & 1.7 \\
\bottomrule
\end{tabular}
\caption{\revs{Prefill--decode mismatch: increase in perplexity (\%) relative to the $N=2$ prefill. The whole-document column processes and scores all 4K tokens with the exact recurrence. The remaining columns prefill a prompt at $N=2$, then process and score only the continuation with the exact recurrence. Both paths use the reference tokens.}}
\label{tab:mismatch}
\end{table}

\begin{table}[h]
\centering
\small
\begin{tabular}{@{}lrrrrrr@{}}
\toprule
 & \multicolumn{2}{c}{4K} & \multicolumn{2}{c}{16K} & \multicolumn{2}{c}{32K} \\
\cmidrule(lr){2-3}\cmidrule(lr){4-5}\cmidrule(l){6-7}
Model & $N=2$ & $N=10$ & $N=2$ & $N=10$ & $N=2$ & $N=10$ \\
\midrule
1.3B & 89.5 & 94.7 & 87.8 & 93.7 & 86.8 & 92.3 \\
1.3B, last 11B at $N=5$ & 90.7 & 94.0 & 89.0 & 91.9 & 88.5 & 91.2 \\
1.3B, Poisson $N$ & 94.5 & 95.9 & 91.7 & 93.0 & 88.1 & 89.2 \\
\bottomrule
\end{tabular}
\caption{\revs{Top-1 agreement (\%) between the prefill at $N$ iterations and the exact recurrence over whole documents of 4K, 16K
and 32K tokens, for the 1.3B models.}}
\label{tab:mismatch-agree}
\end{table}

\begin{revblocks}
\subsection{Learned decay and learning rates}
\label{app:gates}
The conditions of \cref{thm:convergence,thm:convergence-delta} are stated in terms of $\epsilon=\bar\eta/(1-\bar\alpha)$.
\Cref{tab:gates} reports the decay $\alpha_t$ and the learning rates $\eta^{(l)}_t$ that the trained models produce on held-out
SlimPajama text. In most layers the decay is close to one, so the ratio $\eta_t/(1-\alpha_t)$ has a median of about $4$ for the
readout and $7$ to $9$ for the hidden layer, with layer medians up to $100$. The conditions require $\epsilon$ well below one, so the trained models are far outside them. Their agreement with the exact recurrence is measured in \cref{sec:exp:convergence,app:mismatch}.
\end{revblocks}

\begin{table}[h]
\centering
\small
\begin{tabular}{@{}lrlrl@{}}
\toprule
 & \multicolumn{2}{c}{340M} & \multicolumn{2}{c}{1.3B} \\
\cmidrule(lr){2-3} \cmidrule(l){4-5}
Quantity & median & range over layers & median & range over layers \\
\midrule
Decay $\alpha$ & 0.96 & 0.21--0.997 & 0.96 & 0.39--0.998 \\
Readout learning rate $\eta^{(2)}$ & 0.14 & 0.06--0.33 & 0.11 & 0.04--0.59 \\
Hidden learning rate $\eta^{(1)}$ & 0.35 & 0.20--0.49 & 0.23 & 0.13--0.40 \\
$\eta^{(2)}/(1-\alpha)$ & 3.9 & 0.23--27 & 4.1 & 0.54--28 \\
$\eta^{(1)}/(1-\alpha)$ & 9.0 & 0.58--86 & 7.5 & 0.54--100 \\
\bottomrule
\end{tabular}
\caption{\revs{Learned decay and learning rates of the trained \method{} models on held-out text. Each layer is summarized by its
median over tokens and heads; the table gives the median and the range of these 24 layer medians.}}
\label{tab:gates}
\end{table}

\section{Kernel implementation and efficiency details}
\label{app:kernels}
The \method{} layer is built from the chunkwise Triton kernels of FLA with a chunk size of 64: the gated-linear-attention and
gated-delta-rule forward and backward kernels, the transposed reads of \cref{eq:body-la-T} (the same kernel with the roles of keys
and values exchanged) and \kgrad{}, which is the gated-delta-rule kernel with the final read transposed; the pointwise epilogues
(nonlinearity, Jacobian, projection, damping) are fused into two Triton kernels described in \cref{app:epilogue}. Per iteration and hidden layer the layer stores the
pseudo-labels, the features and the Jacobian, each $T\times d_{\text{head}}$ per head; the readout states at chunk boundaries are recomputed in the backward pass. \revs{The layer and full-model measurements are described separately below.}

\begin{revblocks}
\paragraph{Layer benchmarks.}
For \cref{fig:efficiency}(a,b), we time the forward and backward passes of one token-mixing layer on one MI325X, in bf16 at batch size 8, over 20 iterations after 3 warm-up iterations. TTT-MLP uses its fused Triton kernel at head dimension 64 and block size 16. The larger-head points in (b) use a \texttt{torch.compile}d version of its PyTorch scan; at the supported configuration, this scan is $2.2\times$ slower than the fused kernel in the same forward-and-backward setting. The batch-1 matched-error experiment in \cref{fig:convergence}(a) instead uses whichever implementation is faster at each block size (\cref{app:convergence-sweep}).

\paragraph{Full-model training.}
For \cref{fig:efficiency}(c), all 340M models use the recipe and data loader of \cref{app:configs}: 4096-token sequences, eight sequences per GPU, two accumulation steps and eight MI325X GPUs, in bf16. This gives 0.5M tokens per optimizer step. We average throughput over 50 steps after warm-up and report peak memory alongside it in \cref{tab:throughput}. \method{} stores activations across all refinement iterations, so its peak memory grows with $N$. TTT-MLP uses its supported fused configuration (16 heads of dimension 64, block size 16) and checkpoints segments of its sequential scan. These different checkpointing strategies are part of the reported time--memory trade-off.
\end{revblocks}

\begin{table}[h]
\centering
\small
\begin{tabular}{@{}lrr@{}}
\toprule
Model & tokens/s ($\times10^3$) & peak memory (GiB per GPU) \\
\midrule
Gated DeltaNet & 112.1 & 58 \\
Mamba-2 & 108.1 & 64 \\
Attention & 101.7 & 42 \\
\method{}, $N=1$ & 50.2 & 117 \\
\method{}, $N=2$ & 37.9 & 145 \\
\method{}, $N=3$ & 30.5 & 173 \\
\method{}, $N=5$ & 21.8 & 228 \\
LaCT (SWA 2048) & 34.8 & 94 \\
TTT-MLP & 7.8 & 86 \\
\bottomrule
\end{tabular}
\caption{\revs{Training throughput and peak memory of the full 340M models on eight MI325X GPUs at $T=4096$.}}
\label{tab:throughput}
\end{table}

\begin{revblocks}
\paragraph{Training cost of the two 1.3B variants.}\label{app:variant-cost}
We compare the training logs of the three 1.3B runs, which use the same hardware and batch size. Switching to $N=5$ for the final 11B of 100B tokens increases total training time by about 7\%; randomizing $N$ increases it by about 5\%. The former uses more expensive steps only near the end of training, while the latter averages about 2.3 iterations after zero draws are resampled and draws above 5 are capped. Activation storage also grows with $N$: peak memory is 164 GiB per GPU at $N=2$, 205 GiB with randomized $N$ and 247 GiB at $N=5$.
\end{revblocks}

\begin{figure}[h]
    \centering
    \includegraphics[width=\textwidth]{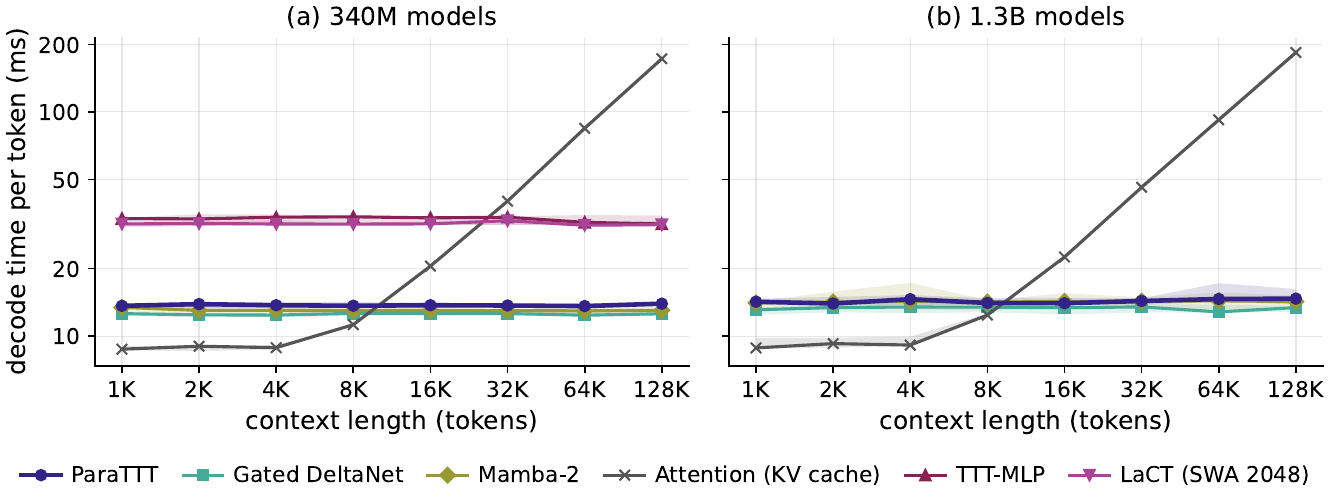}
    \caption{\revs{Decode time per token against context length at batch size 1 on one MI325X, for the trained models of
    \cref{tab:main}. Lines: median over five passes; bands: their range. TTT-MLP and LaCT are trained at 340M only.}}
    \label{fig:decode-latency}
\end{figure}

\begin{revblocks}
\subsection{Decode latency}
\label{app:decode-latency}
\Cref{fig:decode-latency} shows the time to decode one token as a function of the context length, for the trained models of
\cref{tab:main}. Each model processes a prompt of 1K to 128K tokens and then generates one token at a time, at batch size 1 in bf16
on one MI325X. We drop the first 32 generated tokens and take the median time of the next 224 (2048 for LaCT, which covers one of
its chunks). All models of a size run back to back on the same GPU, five times; we report the median of the five passes and their
range.

The recurrent models decode in constant time per token at every context length. \method{} decodes with the exact recurrence,
without iterations, and the single-token step of a layer is implemented as one fused kernel. Its decode time is close to that of
Gated DeltaNet and Mamba-2, about 10\% above Gated DeltaNet, and less than half that of TTT-MLP and LaCT, the two TTT baselines.
The two model sizes are close because, at batch size 1, the time is dominated by the fixed cost of launching the kernels of each
layer. Attention with a key--value cache is the fastest model at short contexts, but its time grows linearly with the context, and
beyond 8K tokens it is slower than \method{}. The decoding state of \method{} is the fast weights of the hidden layer and of the
readout, two $128\times128$ matrices per head against one for Gated DeltaNet, and does not grow with the context.
\end{revblocks}

\subsection{The fused epilogue kernels}
\label{app:epilogue}
With one hidden layer ($L=2$) the only operations of an iteration that are not scans are pointwise in the token: the nonlinearity and
projection that turn the hidden pre-activation into the features (Step 1), and their transposed Jacobians, which are all that Step 3
needs, because the transposed fast weight of \cref{eq:body-la-T} is only required to reach a layer below the first. Both are
implemented as one Triton kernel each, with one program per (sequence, token, head) that holds the $d_{\text{head}}$-vector of its row in
fp32 registers; every reduction is over this vector, so the kernels are memory-bound and read and write each activation once. We write
$s(x)=1/(1+e^{-x})$ for the sigmoid, $\sigma(x)=x\,s(x)$ for SiLU and $n(\bx)=\bx\min(1,1/\|\bx\|)$ for the unit-ball projection.

\paragraph{Forward epilogue.}
The hidden pre-activation of token $t$ is the sum of a slow part and a fast part,
\begin{equation}
    \bz_t \;=\; \underbrace{\bW^{(1)}_0\bk_t+\bb^{(1)}}_{\text{one matmul per forward}}\;+\;\kappa\,\underbrace{\alpha_t\Delta\bW^{(1)}_{t-1}\bk_t}_{\text{shifted scan, \cref{eq:body-la}}},
    \qquad \bh_t=\sigma(\bz_t),\qquad \bphi_t=n(\bh_t),
    \label{eq:epi-fwd}
\end{equation}
where $\kappa$ is the learned per-head gain of \cref{sec:method:practical}. The slow part does not depend on the pseudo-labels, so it is
computed once per layer forward and shared by all iterations; the scan output is read in bf16 straight from the layout the scan
writes, without the copy that separating the two arms of a SwiGLU would need (the value arm is the constant $1$ in our models). The
kernel forms $\bz_t$, $\bh_t$ and $\bphi_t$ in fp32, stores $\bphi_t$ in bf16 for the readout kernels, and saves $\bz_t$ and $\bh_t$
for the backward epilogue; $\|\bh_t\|$ is recomputed there from $\bh_t$.

\paragraph{Backward epilogue.}
\kgrad{} returns $\be_t=\eta^{(L)}_t\,\partial\ell_t/\partial\bphi_t$ \cref{eq:dk}. The backward epilogue turns it into the value the hidden
scan consumes in the next iteration, in five register-resident steps:
\begin{equation}
    \bg_t=\be_t/\eta^{(L)}_t,\qquad
    \bg'_t=\begin{cases}\big(\bg_t-\hat\bh_t\,\hat\bh_t^{\top}\bg_t\big)/\|\bh_t\| & \|\bh_t\|\ge1\\ \bg_t & \|\bh_t\|<1\end{cases},\qquad
    \dzh_t=\sigma'(\bz_t)\odot\bg'_t,
    \label{eq:epi-bwd}
\end{equation}
with $\hat\bh_t=\bh_t/\|\bh_t\|$ and $\sigma'(x)=s(x)\big(1+x(1-s(x))\big)$. The second step is the transposed Jacobian of the projection,
$(\bI-\hat\bh\hat\bh^{\top})/\|\bh\|$ outside the unit ball and the identity inside it, and the third that of SiLU; together they are
the transposed Jacobian $\mathbf{H}_t^{\top}$ of \cref{sec:method:theory}, with the hard projection in place of the soft clip. The value the scan
receives, $\hat\bv_t=-\eta^{(1)}_t\dzh_t$, attaches the hidden learning rate and the sign with which the pseudo-label enters as a value. The damped Picard step
\cref{eq:picard} is fused into the same store, $\bv^{[n]}_t=\bv^{[n-1]}_t+\lambda(\hat\bv_t-\bv^{[n-1]}_t)$, with $\bv^{[0]}=\mathbf 0$
handled without a read, and the result is written directly in the $(B,T,H,d_{\text{head}})$ layout the scan reads, so an iteration
is four kernel launches per layer: the hidden scan, the forward epilogue, \kgrad{} and the backward epilogue. The kernels also
implement a pre-activation RMSNorm, a per-token norm clip of $\dzh_t$ (hard or soft), a skip connection and a second SwiGLU arm, none of
which the reported models use.

\paragraph{Training through the iteration.}
End-to-end training backpropagates through all $N$ iterations, so each epilogue is an autograd function whose backward is itself a
fused kernel: the vector-Jacobian product of \cref{eq:epi-fwd} (through $\sigma'$ and the projection Jacobian) and that of
\cref{eq:epi-bwd}, which differentiates the SiLU Jacobian a second time ($\sigma''$) and the learning rates. For decoding, the hidden scan, the forward epilogue and the \kgrad{} step of one token are
fused into a single kernel per (sequence, head) with the states resident in registers, since at that length the launch latency of the
chunkwise kernels dominates.

\section{Cost of the hidden-layer update and of centering}
\label{app:speed-variants}
\cref{tab:speed-variants} measures what the two \revminor{design choices} of \cref{sec:method:practical} cost, in the two settings of
\cref{fig:efficiency}: one token-mixing layer in isolation and the full 340M model in training.

\paragraph{Readout parameterization.}
The three readouts of \cref{tab:centering} run on the same kernels and differ only in the readout offset and in the extra terms of
\cref{eq:center}. Relative to the zero-initialized readout, the uncentered learned readout adds the offset $\bW^{(L)}_0$ to the final
read of \kgrad{} and costs $12\%$ of the layer time at \revminor{4K} and $9\%$ of training throughput. Centering adds to this one more pass of the
slow MLP over the queries (the features $\bphi_{0,\bq,t}$ of \cref{app:centering}) and the corrected values, which is a further
$1$--$2\%$ of layer time and throughput and $9$\,GiB of activation memory. Centering therefore costs about $11\%$ of training
throughput in total, against the gains of \cref{tab:centering}.

\paragraph{Hidden-layer update.}
The delta-rule body replaces the gated-linear-attention scan of the hidden layer with a gated-delta-rule scan of the same shape, plus
the transposed reads of \cref{eq:dk} that the delta rule's gradient needs (\cref{app:body-delta}); the readout side of the layer is
unchanged. In the reference implementation, which supports both updates, the delta-rule body costs $17\%$ of the layer time at every
length ($35.6$ against $30.4$\,ms at \revminor{4K)} and $1.7$\,GiB more activation memory per layer for the extra chunk states; in training this is $13\%$ of throughput
($32.5$k against $37.4$k tokens/s) and $30$\,GiB of peak memory per GPU. Together with \cref{tab:main} (SWDE $28.4\to41.9$, NIAH at \revminor{4K} $7.2\to32.6$), \revminor{this is the more cost-effective of the two design choices.}

\begin{table}[t]
\centering
\small
\setlength{\tabcolsep}{4pt}
\begin{tabular}{@{}l rrrrr rr@{}}
\toprule
 & \multicolumn{5}{c}{Layer forward+backward (ms) at $T$} & \multicolumn{2}{c}{Training, 8 GPUs} \\
\cmidrule(lr){2-6} \cmidrule(lr){7-8}
Variant & \revminor{1K} & \revminor{2K} & \revminor{4K} & \revminor{8K} & \revminor{16K} & tokens/s ($\times10^3$) & GiB/GPU \\
\midrule
\multicolumn{8}{@{}l}{\textit{Readout parameterization (delta-rule body, optimized kernels)}} \\
Zero-init readout & 7.3 & 13.1 & 26.2 & 52.2 & 104.1 & 42.7 & 131 \\
Learned-init readout, uncentered & 8.3 & 14.8 & 29.4 & 58.8 & 116.7 & 38.7 & 136 \\
Centered readout (\methodshort{}) & 8.5 & 15.1 & 29.8 & 59.8 & 118.8 & 38.0 & 145 \\
\midrule
\multicolumn{8}{@{}l}{\textit{Hidden-layer update (centered readout, reference implementation)}} \\
Linear-attention body & 8.4 & 15.3 & 30.4 & 60.7 & 120.3 & 37.4 & 107 \\
Delta-rule body (\methodshort{}) & 10.0 & 18.0 & 35.6 & 70.9 & 140.3 & 32.5 & 138 \\
\midrule
\multicolumn{8}{@{}l}{\textit{Reference}} \\
Gated DeltaNet & 2.1 & 3.3 & 6.3 & 11.9 & 23.2 & 112.1 & 58 \\
\bottomrule
\end{tabular}
\caption{Cost of the hidden-layer update and of the readout parameterization at 340M, $N=2$. Left: forward+backward time of one token-mixing layer (bf16, batch 8, $d=1024$, 8 heads of dimension 128, one MI325X), the setting of \cref{fig:efficiency}(a). Right: training throughput and peak memory of the full model on one node of eight MI325X GPUs with the recipe of \cref{app:configs}. The readout rows use the optimized kernels of \cref{app:kernels}; the two bodies are timed in the reference implementation, which supports both updates, so the body rows are comparable with each other but slower than the readout rows.}
\label{tab:speed-variants}
\end{table}

\newpage
\section{Convergence versus compute at other sequence lengths}
\label{app:convergence-sweep}
\begin{figure}[h]
    \centering
    \includegraphics[width=\textwidth]{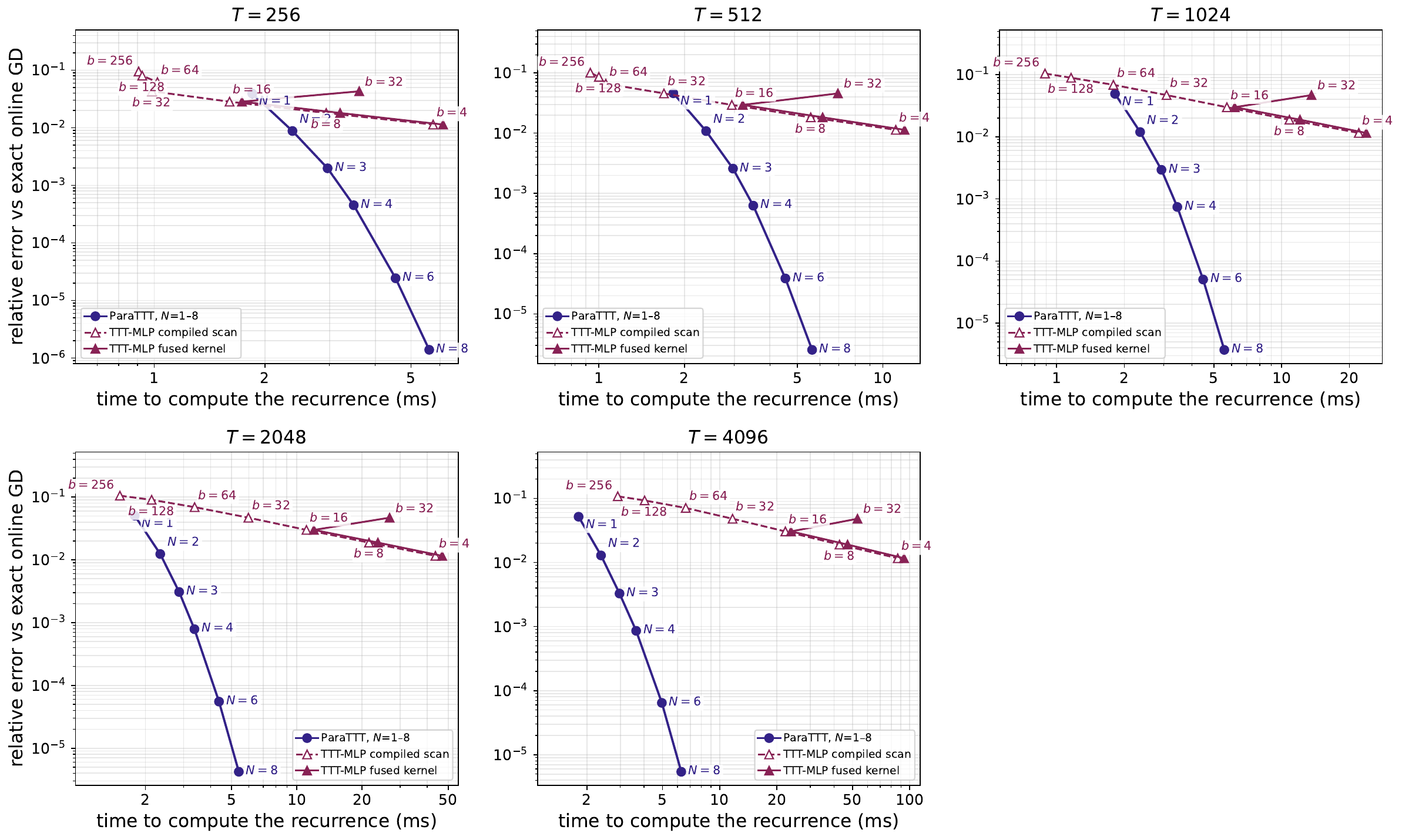}
    \caption{Relative error against exact online GD versus the time to compute the forward recurrence of one 340M layer (batch 1),
    at five sequence lengths; same setting and legend as \cref{fig:convergence}(a).}
    \label{fig:convergence-sweep}
\end{figure}

\Cref{fig:convergence-sweep} shows the comparison of \cref{fig:convergence}(a) at every sequence length from $256$ to $4096$ with the same inner MLP, step
size and decay. The picture is the same at every length: TTT-MLP's error is set by its block size and its time by the number
of sequential blocks $T/b$, whereas \method{}'s error is set by the iteration count and its time, being a fixed number of chunkwise-parallel
scans, is nearly independent of $T$ at batch 1, so the gap widens with the sequence length.

\begin{revblocks}
\Cref{tab:error-time} lists the points of \cref{fig:convergence}(a) at $T=4096$. At its default block size of 16, TTT-MLP has
$2.4\times$ the error of \method{} at $N=2$ and takes $9.3\times$ as long. The $32\times$ of \cref{sec:exp:convergence} compares the
two at equal error ($1.3\%$), which TTT-MLP reaches at $b\approx5$ by interpolation between $b=4$ and $b=8$. TTT-MLP is timed with
the faster of its fused kernel and a compiled version of its scan; the fused kernel pads blocks to 16 tokens, so the compiled scan
is the faster one at every block size in the table. A bf16 variant of the fused kernel that we wrote for $b=16$ takes $19.5$\,ms;
it is not part of the baseline implementation and is not used in the comparison.
\end{revblocks}

\begin{table}[h]
\centering
\small
\begin{tabular}{@{}lrrr@{}}
\toprule
 & relative error & \multicolumn{2}{c}{time (ms)} \\
\cmidrule(l){3-4}
Configuration & vs.\ exact GD (\%) & compiled scan & fused kernel \\
\midrule
\method{}, $N=1$ & 5.2 & \multicolumn{2}{c}{1.8} \\
\method{}, $N=2$ & 1.3 & \multicolumn{2}{c}{2.4} \\
\method{}, $N=3$ & 0.33 & \multicolumn{2}{c}{3.0} \\
\method{}, $N=4$ & 0.09 & \multicolumn{2}{c}{3.6} \\
\midrule
TTT-MLP, $b=4$ & 1.2 & 86.6 & 93.4 \\
TTT-MLP, $b=8$ & 1.9 & 42.7 & 47.1 \\
TTT-MLP, $b=16$ & 3.1 & 22.2 & 23.7 \\
TTT-MLP, $b=32$ & 4.8 & 11.7 & 53.1 \\
\bottomrule
\end{tabular}
\caption{\revs{Error against exact online gradient descent and time to compute the recurrence at $T=4096$, for the setting of
\cref{fig:convergence}(a): one random inner MLP, $d=64$, eight heads, step size $0.2$ and decay $0.98$.}}
\label{tab:error-time}
\end{table}

\end{document}